\documentclass{article}
\pdfoutput=1

\usepackage{arxiv}
\usepackage[utf8]{inputenc} % allow utf-8 input
\usepackage[T1]{fontenc}    % use 8-bit T1 fonts
\usepackage{hyperref}       % hyperlinks, to remove boxes \usepackage[hidelinks]{hyperref}
\usepackage{url}            % simple URL typesetting
\usepackage{booktabs}       % professional-quality tables
\usepackage{amsfonts}       % blackboard math symbols
\usepackage{nicefrac}       % compact symbols for 1/2, etc.
\usepackage{microtype}      % microtypography
\usepackage{lipsum}		% Can be removed after putting your text content
\usepackage{graphicx} % use [draft] to remove figures while drafting
\usepackage{natbib}
\usepackage{doi}
\usepackage{caption} % to change the font size of the caption
\usepackage{xcolor} % to change the color of the caption
\usepackage{float}

\definecolor{lightgray}{gray}{0.3}
\title{Emergence and Function of Abstract Representations in Self-Supervised Transformers}

\author{Quentin RV.~Ferry\thanks{Correspondence to \texttt{qferry@mit.edu}, alternatively \texttt{qferry.ai@gmail.com}},\hspace{2mm}Joshua~Ching, \hspace{1mm}Takashi~Kawai\\
	Picower Institute for Learning and Memory\\
	Massachusetts Institute of Technology\\
	Cambridge, MA
}

\date{December 8, 2023}

\renewcommand{\shorttitle}{Abstraction in Self-Supervised Transformer}

\hypersetup{
pdftitle={Emergence and Function of Abstract Representations in Self-Supervised Transformers},
pdfsubject={Mechanistic Interpretability, Self-Supervised Learning, Transformer, Abstraction, Compositionality},
pdfauthor={Quentin RV.~Ferry, Joshua~Ching, Takashi~Kawai},
pdfkeywords={Mechanistic Interpretability, Self-Supervised Learning, Transformer, Abstraction, Compositionality},
}

\begin{document}
\maketitle

%/////////////////////////////////////////////////////////////////////////////////////////
% ABSTRACT
%/////////////////////////////////////////////////////////////////////////////////////////

\begin{abstract}
	Driven by the need to predict how our actions might affect the environment, our brains have evolved the ability to abstract from raw experiences a mental model of the world that succinctly captures the hidden blueprint of our reality. Central to human intelligence, this abstract world model notably allows us to effortlessly handle novel situations by generalizing prior knowledge, a trait deep learning systems have historically struggled to replicate. However, the recent shift from label-based (supervised) to more natural predictive objectives (self-supervised), combined with expressive transformer-based architectures, have yielded powerful foundation models that appear to learn versatile representations that generalize to support a wide range of downstream tasks. This promising development raises the intriguing possibility of such models developing in silico abstract world models. We test this hypothesis by studying the inner workings of small-scale transformers trained to reconstruct partially masked visual scenes generated from a simple blueprint. We show that the network develops intermediate abstract representations, or abstractions, that encode all semantic features of the dataset. These abstractions manifest as low-dimensional manifolds where the embeddings of semantically related tokens transiently converge, thus allowing for the generalization of downstream computations. Using precise manipulation experiments, we demonstrate that abstractions play a central role in the network’s decision-making process. Our research also suggests that these abstractions are compositionally structured, exhibiting features like contextual independence and part-whole relationships that mirror the compositional nature of the dataset. Finally, we introduce a Language-Enhanced Architecture (LEA) designed to encourage the network to talk about its computations. We find that LEA develops a specialized language centered around the aforementioned abstractions, allowing us to steer the network's decision-making more readily.
\end{abstract}

% keywords can be removed

%/////////////////////////////////////////////////////////////////////////////////////////
% TABLE OF CONTENTS
%/////////////////////////////////////////////////////////////////////////////////////////

% \tableofcontents

%/////////////////////////////////////////////////////////////////////////////////////////
% INTRODUCTION
%/////////////////////////////////////////////////////////////////////////////////////////

\section{Introduction}
\label{sec:intro}

Despite achieving impressive fits, Deep Learning systems trained via supervised learning notoriously struggle to adapt to scenarios not covered in their training data \cite{thompson2020computational, ye2022ood}. In contrast, humans excel at navigating novel situations, such as getting a much-needed cup of coffee while visiting an unfamiliar conference center, by rapidly generalizing previous experiences. This gift for generalization can be attributed to two key factors: (i) Our brain's capacity to construct, through repeated experience, an \textit{abstract world model}, which succinctly describes our environment and specifies how it might change as a consequence of our actions; (ii) The ability to map incoming raw sensory data onto this model, enabling real-time adaptation and decision-making \cite{gyorgy2019brain,grossberg2021conscious,friston2021world,behrens2018cognitive}. Abstract world models essentially serve as cognitive renditions of an underlying blueprint of reality. Remarkably, this blueprint is never directly given to us. Instead, our brains have evolved to abstract this information from raw sensory inputs, as a product of strong prediction objectives (e.g., foresee the outcome of our actions, fill in incomplete sensory data, etc.)\cite{friston2010free,grossberg2013adaptive}. This form of learning stands in stark contrast with the aforementioned supervised learning paradigm where AI systems are spoon-fed the blueprint via labeled data.   

However, the field of Artificial Intelligence is now shifting its emphasis from narrow, specialized models trained with supervised learning to broader, more general models nurtured through self-supervised learning (SSL)\cite{shwartz2023compress,lecunpath,rani2023self,gui2023survey}. Intriguingly, this learning paradigm relies on predictive objectives similar to the ones our brain faces: models are not provided with explicit labels but instead are trained to predict or reconstruct their inputs. This new strategy, applied to expressive transformer-based architectures \cite{vaswani2017attention}, has yielded a family of powerful \textit{foundation models} that seem to develop latent representations that can generalize to support a large array of downstream tasks\cite{devlin2018bert,radford2021learning,caron2021emerging,baevski2022data2vec}. The emergence of such versatile representations suggests the possibility of these models learning an \emph{in-silico} equivalent to biological abstract world models.

In light of these observations, we set out to empirically test the hypothesis that deep learning systems, when trained using self-supervised objectives, construct internal representations that capture the latent blueprint used to generate their inputs. Drawing upon a careful analysis of small-scale transformers trained on simple tasks, we uncover evidence to support this notion. We found that these systems evolve a collection of abstract representations, or \textit{abstractions}, that together form an abstract world model. We show that these abstractions act as low-dimensional, linearly separable attractors that serve as pivot points for downstream computational generalization. Using gain-of-function manipulation experiments, we further validate the causal role that these abstract representations play in the network's decision-making process. We also provide qualitative evidence supporting the fact that abstractions are organized in a computationally advantageous manner: factorized at the representational level \cite{bengio2013representation,higgins2016beta,kulkarni2015deep} and organized in part-whole hierarchies \cite{szabo2012case,lake2015human,feldman1997structure,biederman1987recognition,bienenstock1996compositionality,hinton2023represent}. Finally, we introduce a novel network architecture featuring a language bottleneck, designed to forces the SSL-trained transformer to talk about its internal computations. We find that such systems evolve a language centered around abstractions, creating a quasi-one-to-one mapping between words and semantic features of the inputs’ blueprint. We show that uncovering this mapping is relatively straightforward, providing us with a simple way to steer the network's behavior. Altogether, our work not only substantiates the claim that SSL-trained transformers can form an abstract world model but also uncovers the computational and organizational properties of these models, offering new avenues for more interpretable artificial intelligence systems.

%/////////////////////////////////////////////////////////////////////////////////////////
% METHODS
%/////////////////////////////////////////////////////////////////////////////////////////

\section{Methods}
\label{sec:methods}

Several studies, that trained self-supervised vision transformers\cite{dosovitskiy2020image} to reconstruct partially masked images, have reported results suggesting that these models learn to segment objects \cite{caron2021emerging,oquab2023dinov2}: Despite being only trained to perform predictions at the level of pixels and never being given any label information (e.g. this image contains a dog), such systems appear to learn something about the various objects present in the images. From a prediction standpoint, learning to discern objects is certainly advantageous as knowing the structure of a given object helps guess missing parts from partial views. Additionally, the detection of certain objects might also help the system establish a context from which to better infer information about other parts of the image (e.g. a masked piano in the middle of an orchestra).

In this study, we sought to test whether self-supervised transformers evolve computations that assign similar representations to perceptually distinct visual tokens sharing the same semantic meaning (e.g., being part of the same object). In particular, we wished to show that such representations exist and understand how they play into the inference process of the studied networks. Additionally, we wondered if the latent representations would capture the compositional nature of more complex objects (e.g. a face is composed of a mouth, two eyes, etc.). To facilitate interpretability, we have drastically simplified the problem to gain full control over the type and complexity of the inputs, and training curriculum. We replaced complex natural images with boards of $n \times n$ "visual" tokens, featuring a uniform background onto which one or multiple synthetic objects are overlaid (Figure \ref{fig:methods_dataset}). We refer to such datasets as “hierarchical object datasets” (HOD).

\subsection{Dataset and training objectives}
\label{subsec:dataset}

\begin{figure}[ht]
	\centering
	\includegraphics[width=0.5\textwidth]{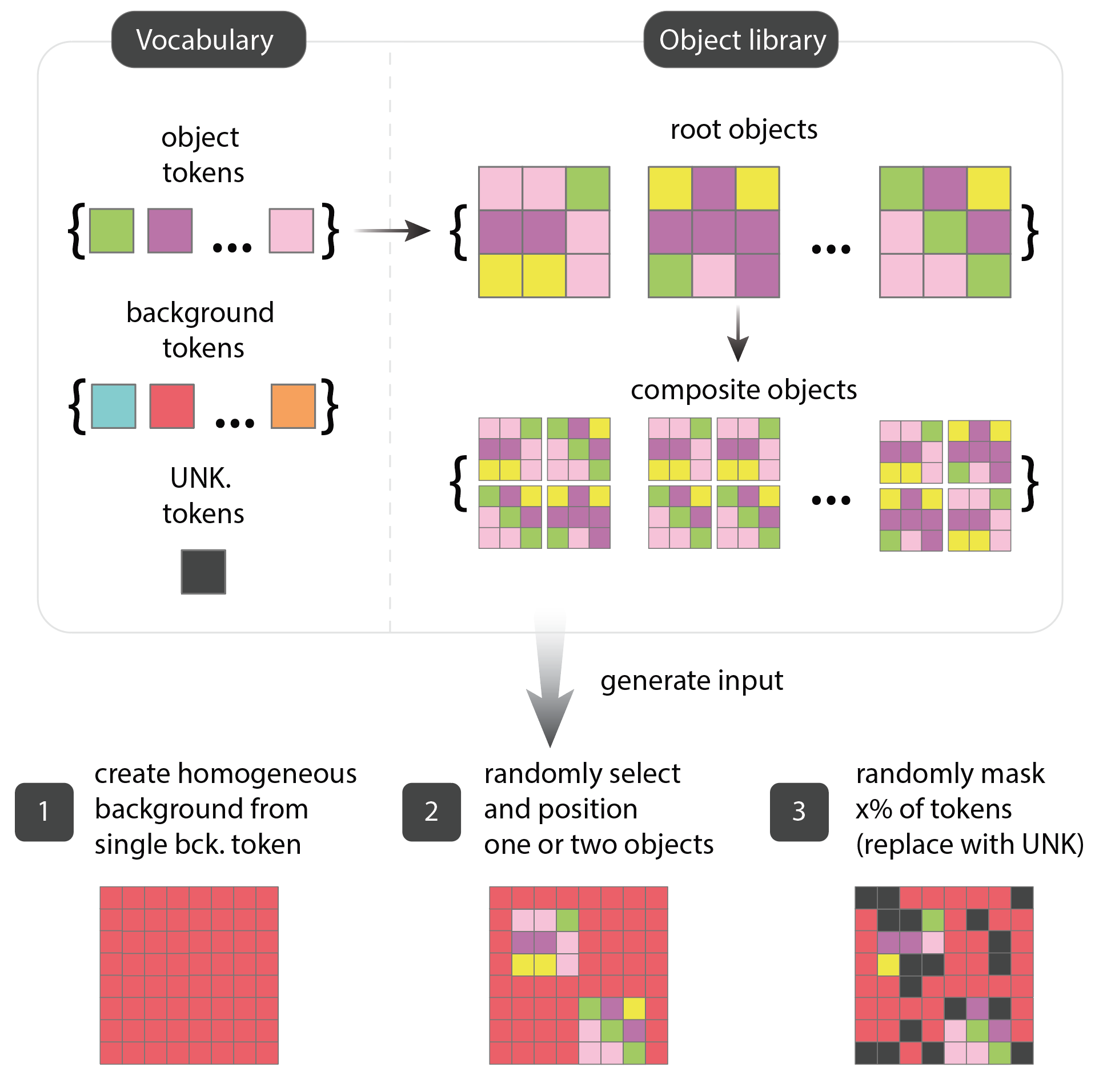}
	\caption{Schematic of the Hierarchical Object Dataset (HOD). \textbf{Top}: Dataset blueprint: Each dataset has a specific blueprint determined before training. The blueprint contains a set of root and composite objects. \textbf{Bottom}: Instance generation process. The blueprint is used to generate a large number of boards that serve as inputs for the neural network.}
	\label{fig:methods_dataset}
\end{figure}

HOD features a vocabulary of $n_b$ background tokens (identified by integers 1 through $n_b$), $n_o$ object tokens ($n_b+1$ through $n_b+n_o$), and a single unknown token (UNK., $n_b+n_o+1$) (Figure \ref{fig:methods_dataset}, top). From this vocabulary, a fixed set of $N_{root}$ root objects are created by randomly sampling $m_{root}$ object tokens with replacement (i.e., a given token can appear in different objects) and arranging them on a specific relational grid $\mathbf{g}_{root}$. For example, an object could be created by positioning object tokens 3, 3, 2, 5 at coordinates (1,1), (1,3), (2,2), and (3,2) of a $3 \times 3$ grid, similar to the way eyes, nose, mouth come together to form a face. In certain cases, we create different instances of the same object (\textit{fuzzy} objects), by allowing certain tokens within an object prototype to take one of $k$ possible token identities (IDs). Going back to the example, tokens at positions (1,1) and (1,3) could either be 1 or 3 to indicate that eyes are close or open. Finally, to test for representational compositionality, we generate a second set of $N_{comp}$ composite objects created by assembling a randomly sampled set of $m_{comp}$ root objects and arranging them according to $\textbf{g}_{comp}$. Unless otherwise stated, we used the following parameter values for the datasets in this study: $n_b=10$, $n_o=10$, $N_{root}=10$, $m_{root} = 9$, $\textbf{g}_{root} = (3,3)$, $N_{comp}=5$, $m_{comp} = 4$, $\textbf{g}_{comp} = (2,2)$, and $n=\{8,10\}$ (see Sup. Table \ref{tab:parameters_HOD}).

Using a given HOB instance (i.e., a specific set of root and composite objects), a large number of masked inputs can be created using the following procedure (Figure \ref{fig:methods_dataset}, bottom): (i) Sample a random background token to tile the entire board; (ii) Randomly select and position one or multiple objects on the board; (iii) Mask a certain percentage of all token (i.e., replace the original token by the UNK. token). When indicated in the text, we also perform patch masking, where larger patches of the board are masked at once.

The masked boards are fed to a Transformer (see section \ref{subsec:architectures}) tasked with reconstructing the original board, i.e., inferring the true token IDs hidden behind UNK. tokens. We perform batch training using Adam optimizer\cite{kingma2014adam} to optimize a cross-entropy loss\cite{mao2023cross} over token IDs. As in Bert\cite{devlin2018bert}, inferences are based on the entire board, meaning that we do not use the causal attention masks employed in training autoregressive models like GPT\cite{brown2020language}.

%-----------------------------------------------------------------------------------------

\subsection{Model architectures}
\label{subsec:architectures}

Here we detail the transformer architectures used throughout the paper, making the distinction between the vanilla architecture used for the majority of the study, and the language-enhanced architecture featured in the last result section.

\begin{figure}[ht]
	\centering
	\includegraphics[width=0.65\textwidth]{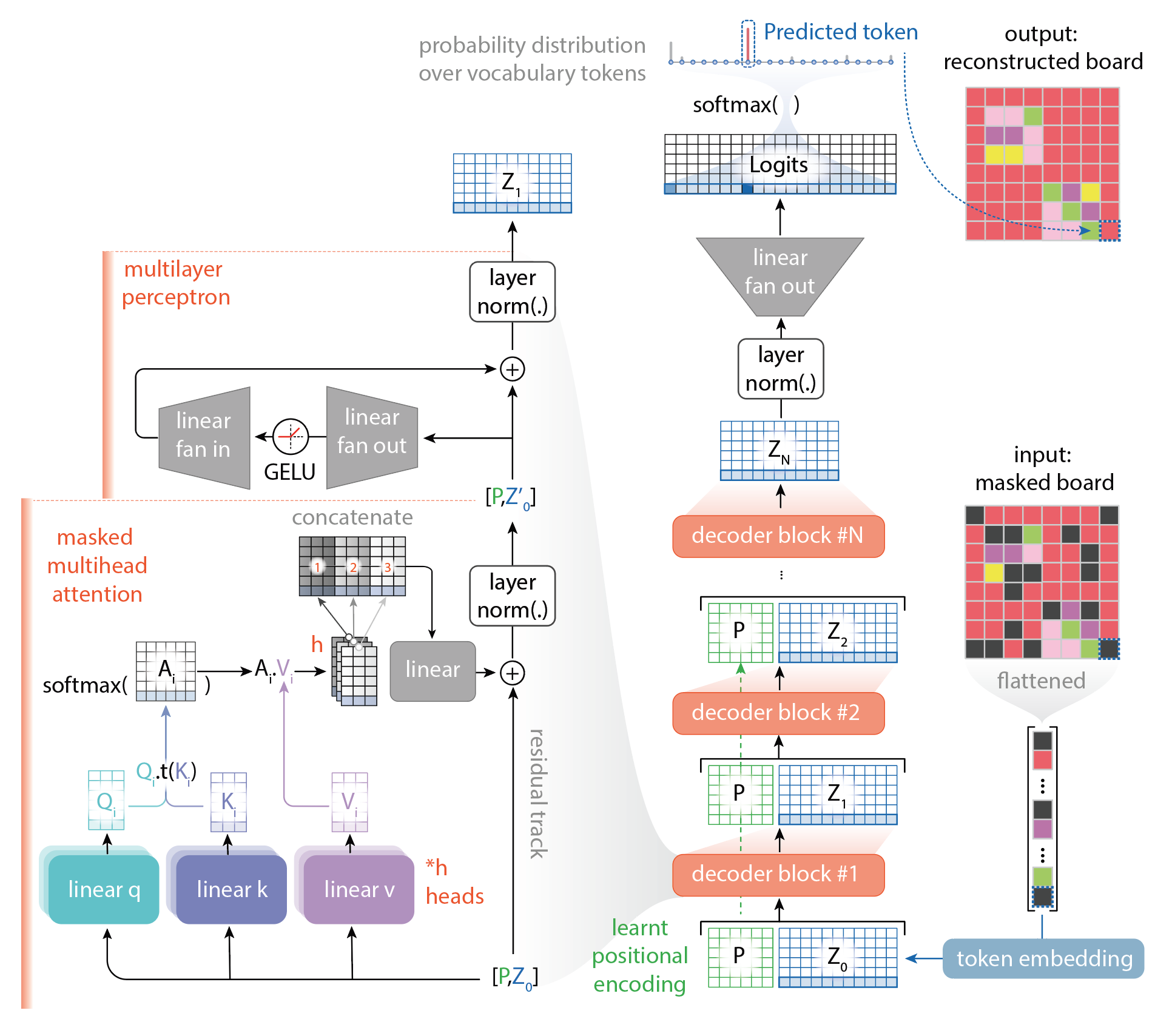}
	\caption{Schematic of the factorized vanilla transformer architecture. A variant of the encoder transformer architecture from Vaswani et al. 2017\cite{vaswani2017attention}.}
	\label{fig:methods_architecture_vanilla}
\end{figure}

\textbf{Vanilla architecture}\footnote{Used in result sections \ref{sec:parameter_tuning},\ref{sec:abstraction_existence},\ref{sec:abstraction_manipulation},\ref{sec:compositionality}.}: We are training a transformer encoder\cite{vaswani2017attention} obtained by stacking several self-attention + multi-layer perceptron (MLP) blocks (see Figure \ref{fig:methods_architecture_vanilla}). However, we have modified the original architecture to separate positional encodings and token embeddings throughout all computational stages: Instead of adding the learned positional encoding $\mathbf{P}$ and token embeddings together and feeding the resulting matrix as input to the first block, $\mathbf{P}$ is provided as an independent immutable matrix throughout all stages of processing. $\mathbf{P}$ is concatenated to the current latent embeddings $\mathbf{Z}$ before every self-attention and MLP subblock (Figure \ref{fig:methods_architecture_vanilla}). While transformers can easily handle representations that mix several streams of information (e.g. position and token ID), we reasoned that using factorized $\mathbf{Z} \times \mathbf{P}$ representations would relieve the need for the network to encode positional information in the residual stream, therefore reducing potential confounds and facilitating both interpretation and manipulation of the network’s computations.

Each $n \times n$ board in the batch gets reshaped into $\mathbf{Z}^0$, an $n^2 \times d_e$ matrix where each token $i$ is given a $d_e$-dimensional latent representation $\mathbf{z}^0_i$, according to a vocabulary of learned token embeddings. $\mathbf{Z}^0$ is then passed through a series of $k_b$ transformer blocks to yield $\mathbf{Z}^{k_b}$. At each block, the input $\mathbf{Z}^i$ is layer normalized\cite{ba2016layer} and concatenated with the matrix of learned positional encodings $\mathbf{P}$ before being passed in succession through attention and MLP subblocks to yield $\mathbf{Z}^{i+1}$ (Figure \ref{fig:methods_architecture_vanilla}): $\mathbf{Y} = \mathbf{Z}^i + attn([ln(\mathbf{Z}^i), \mathbf{P}])$, $\mathbf{Z}^{i+1} = \mathbf{Y} + mlp([ln(\mathbf{Y}), \mathbf{P}])$ where $attn$, $mlp$, and $ln$ stands for self-attention subblock, MLP subblock, and layer norm, respectively. The final $\mathbf{Z}^{k_b}$, obtained after passing through all transformer blocks, is then layer normalized passed through a linear prediction head $h$ to produce logits over the token vocabulary (degree of belief for a particular token at a particular board position) for each board position: $logit_i = h(ln(\mathbf{Z}^{k_b}_i))$. We score the quality of the reconstruction by comparing the logits to the ground truth board using a cross-entropy loss\cite{mao2023cross}. Unless otherwise stated, we use the following parameter values for the networks used in this study: $n=\{8,10\}$, $d_e=64$, $d_p=32$, $k_b=3$, number of attention heads $k_h=2$ (see Sup. Table \ref{tab:parameters_vanilla}).

\begin{figure}[ht]
	\centering
	\includegraphics[width=0.55\textwidth]{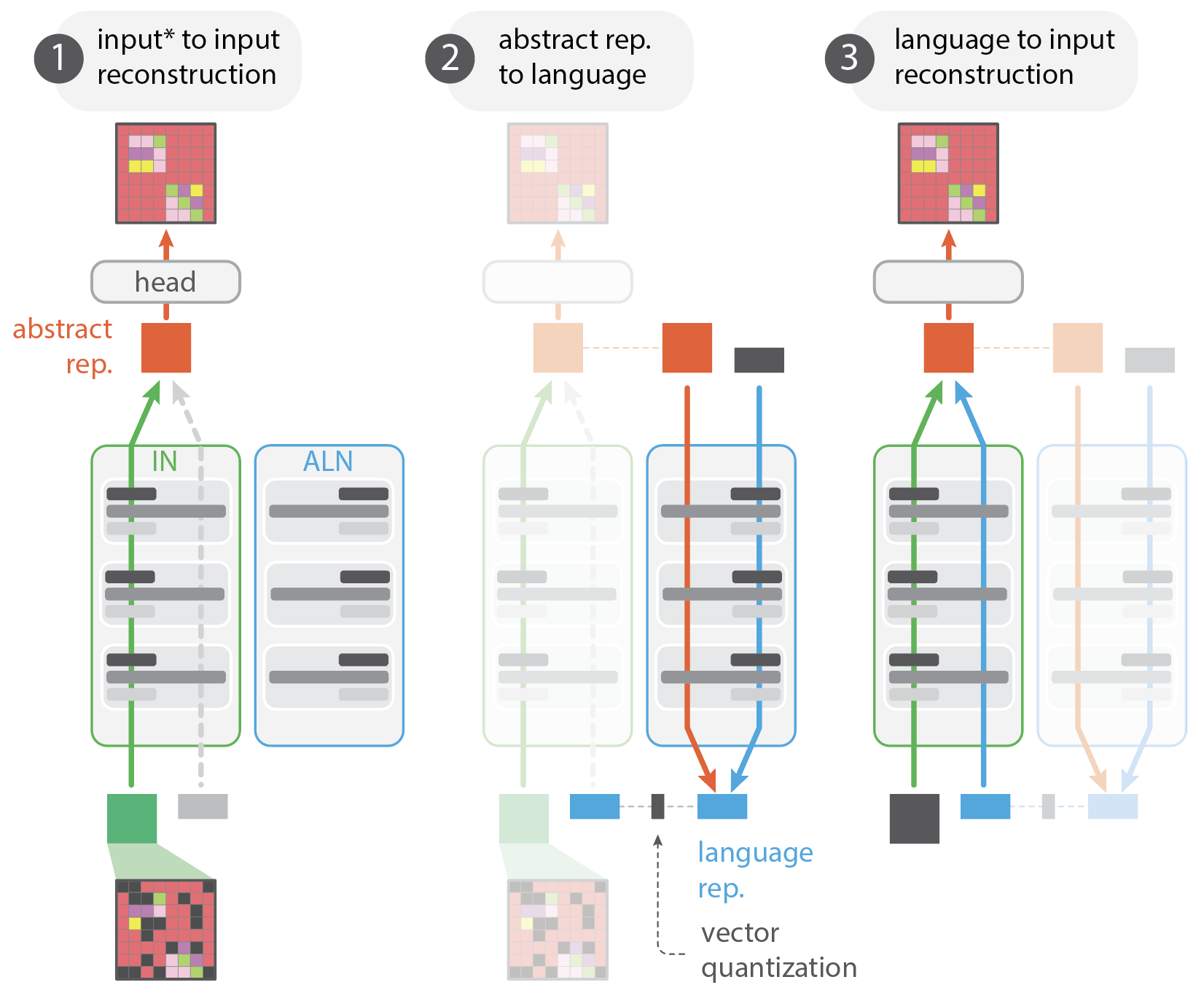}
	\caption{Schematic of the language-enhanced architecture (LEA). The figure shows three passes. \textbf{Pass 1}: a masked board input is reconstructed by the inference network (IN); \textbf{Pass 2}: The last abstract representations $\mathbf{Z}^{k_b}$ of IN are used by the auxiliary language network (ALN) to produce a representation that is vector-quantized to generate the sentence $\mathbf{S}$. \textbf{Pass 3}: Starting from a fully masked board, IN uses $\mathbf{S}$ to reconstruct the original unmasked input.}
	\label{fig:methods_architecture_language}
\end{figure}

\textbf{Language-enhanced architecture}\footnote{Used in result section \ref{sec:language_bottleneck}.}: In section \ref{sec:abstraction_existence}, we show that the vanilla architecture detailed above develops abstract representations. While these representations can be observed by examining the intermediate activations of the network, they are not part of the network’s output and therefore remain hidden from human interpreters. To remedy this limitation, we introduce a language-enhanced architecture (LEA, Figure \ref{fig:methods_architecture_language}), designed to extract the learned abstractions in a language-like form. LEA features two transformer decoder networks\cite{vaswani2017attention} (Figure \ref{fig:methods_architecture_language}): (i) An inference network (IN), whose role is to reconstruct board from either masked input or language; (ii) An auxiliary language network (ALN), which generates a language-like representation of IN's inner-workings. 

As in the vanilla architecture described above, IN takes in masked inputs and outputs the corresponding unmasked boards. It converts $\mathbf{Z}^0$ into $\mathbf{Z}^{k_b}$ before producing logits. However, its transformer blocks were modified to include a cross-attention subblock which allows IN to also generate $\mathbf{Z}^{k_b}$ by interpreting ALN's output. ALN uses self and cross-attention to construct a language-like description $\mathbf{S}$ of the board (or IN inner-workings) from $\mathbf{Z}^{k_b}$. Therefore, both networks feature a residual stream onto which computations are performed and an immutable external source of information. In the case of IN, the external source of information is $\mathbf{S}$, the language representation, and in the case of ALN, it is $\mathbf{Z}^{k_b}$, the last hidden representation produced by IN. In general terms, the computations of a single decoder block can be summarized as follows: The block takes in the current latent representation $\mathbf{Z}^i$ and additional external information in the form of a $\mathbf{W}$ matrix. $\mathbf{Z}^i$ is first passed through a layer norm, concatenated with the corresponding positional encoding $\mathbf{P}_Z$ before being processed by a self-attention subblock: $\mathbf{Y}^i = \mathbf{Z}^i + attn_{self}([ln(\mathbf{Z}^i), \mathbf{P}_Z])$. In the self-attention subblock, keys, queries, and values are all generated from $\mathbf{Z}^i$. The output $\mathbf{Y}^i$ is then processed by a cross-attention subblock: $\mathbf{X}^i = \mathbf{Y}^i + attn_{cross}([ln(\mathbf{Y}^i),\mathbf{P}_Z], [\mathbf{W}, \mathbf{P}_W])$. In the cross-attention subblock, keys and values are generated from the external information source $\mathbf{W}$, while $\mathbf{Y}^i$ is used to generate the queries. Finally, $\mathbf{X}^i$ is passed through an MLP to yield $\mathbf{Z}^{i+1}$: $\mathbf{Z}^{i+1} = \mathbf{X}^i + mlp(ln([\mathbf{X}^i), \mathbf{P}_Z])$.

A complete forward pass through LEA consists of the following three steps (Figure \ref{fig:methods_architecture_language}): (i, \textbf{pass 1}) IN reconstructs the ground truth unmasked board from masked board input: Each $n \times n$ board in the batch is converted to a $n^2 \times d_e$ matrix $\mathbf{Z}^0$ with the initial learned token embeddings. $\mathbf{Z}^0$ is passed through the inference network to yield $\mathbf{Z}^{k_b}$ (output of the last decoder block), and then logits $L_{b}$. IN’s $k_b$ decoder blocks also require $\mathbf{S}$, an external information matrix that encodes a sentence. During pass 1, we set $\mathbf{S}$ to $\mathbf{0}$, an empty matrix with $\mathbf{S}$'s dimensions. (ii, \textbf{pass 2}) ALN constructs a linguistic representation from $\mathbf{Z}^{k_b}$: Each of ALN's $k_b$ decoder blocks takes in $\mathbf{Q}^i$, the output of the previous block, and $\mathbf{Z}^{k_b}$, as the external source of information, to yield $\mathbf{Q}^{i+1}$. We initialize $\mathbf{Q}^0 = \mathbf{Q}_{init}$ where $\mathbf{Q}_{init}$ is a learned language primer matrix. The resulting $\mathbf{Q}^{k_b}$ is then vector quantized (VQ, our implementation uses exponential moving average\cite{van2017neural}) where each $\mathbf{Q}^{k_b}_i$ is matched to one of $n_q$ codebook vectors to yield $\mathbf{S}$. The process also gives us a $l \times V$ one-hot encoding matrix, where $l$ is the number of “words” in a sentence and $V$ is the size of the VQ codebook. (iii, \textbf{pass 3}) IN is tasked with reconstructing the board from $\mathbf{S}$ alone: Same as (i) but $\mathbf{Z}^0$ is initialized with UNK. token embeddings and the output $\mathbf{S}$ from pass 2 is used as the external source of information. IN produces a new $\mathbf{Z}^{k_b}$ and logits $L_{s}$. Altogether, the network learns two positional encoding $\mathbf{P}_Z$ and $\mathbf{P}_Q$, a set of token embeddings, the language primer matrix $\mathbf{Q}_{init}$, and a codebook of $n_q$ vectors.

To train LEA, we use a custom loss function that features the following terms: (i) cross-entropy loss for reconstruction from masked board input (pass 1, $\mathbf{L}_{b}$); (ii) cross-entropy loss for reconstruction from linguistic representation (pass 3, $\mathbf{L}_{s}$); (iii) mean squared error for the vector quantization error (commitment cost + usage cost)\cite{van2017neural}; (iv) sparsity loss on the linguistic representation to encourage the use of fewer words. All terms are combined in a weighted sum with coefficients that are hyperparameters of the system (see Sup. Table \ref{tab:parameters_LEA} for all parameter values).

%/////////////////////////////////////////////////////////////////////////////////////////
% RESULTS 1 - Model performance & hyperparameter tuning
%/////////////////////////////////////////////////////////////////////////////////////////

\section{Results}
\label{sec:results}

\subsection{Model performance and hyperparameter tuning}
\label{sec:parameter_tuning}

\begin{figure}[ht]
	\centering
	\includegraphics[width=0.9\textwidth]{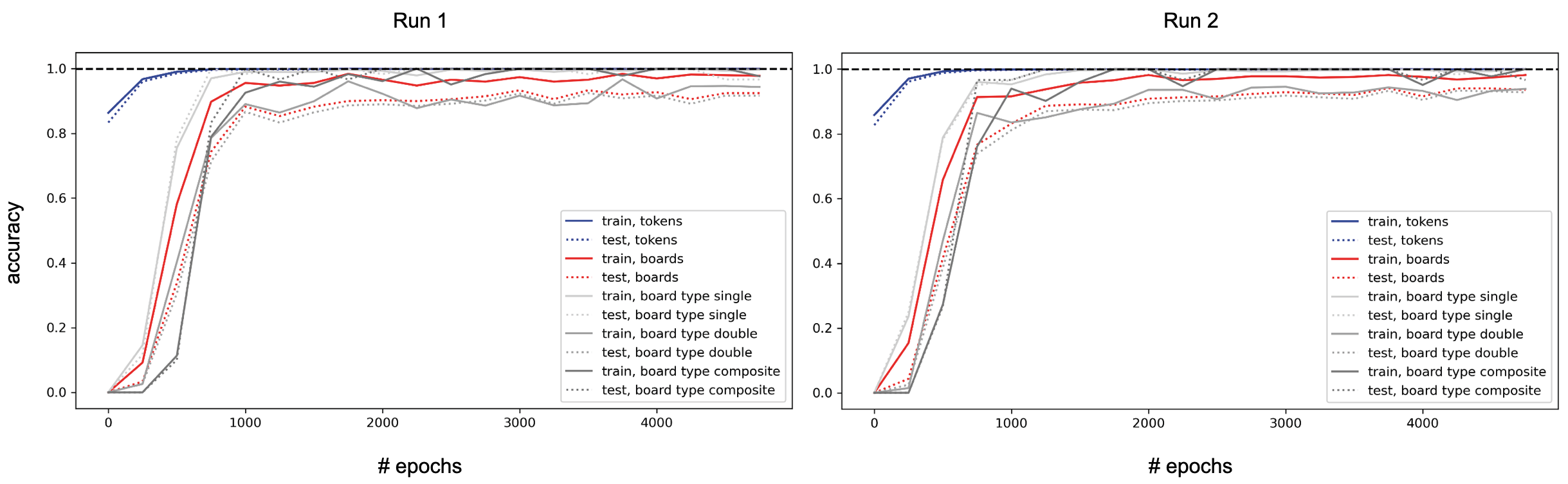}
	\caption{Reconstruction accuracies over training and test set across training epochs. \textbf{Left} and \textbf{right} panels show results from two independent runs (different network initialization and dataset instance). At various epochs, networks were tasked with reconstructing three types of masked boards featuring a single root object (single), two root objects (double), or one composite object (composite), respectively. Two types of accuracies are reported: (i) token accuracy (blue), which measures how many tokens are correctly reconstructed across all boards in a set; (ii) board accuracy (red), which measures the fraction of all masked boards that are fully correctly reconstructed. Board accuracies for various types of boards are shown as different shades of gray.}
	\label{fig:parameter_tuning_performance}
\end{figure}

We first sought to identify hyperparameters for the vanilla architecture that would offer a good compromise between model size (smaller networks are arguably easier to analyze and less prone to overfitting) and model performance. To that end, we conducted a hyperparameter search that included varying the number of transformer blocks, the number of attention heads, and the dimension of token embeddings (Sup. Figure \ref{supfig:hyperparameter_search}). Each model was trained on an instance of HOD with identical parameters (cardinality of token vocabulary, number and complexity of objects) and evaluated on a test set of held-out boards. Ultimately, we settled on an intermediate configuration featuring 3 transformer blocks, 2 attention heads, and 64-dimensional embeddings (see Sup. Table \ref{tab:parameters_vanilla}). We found that despite their relatively small size, networks with these hyperparameters were able to consistently reconstruct masked boards that were not part of their training data (Figure \ref{fig:parameter_tuning_performance}). This generalization to a held-out test set suggested that these networks might have learned something about the compositional nature of the board generation process rather than solving the reconstruction problem via simple rote memorization. We used this combination of hyperparameters for all subsequent experiments.

%/////////////////////////////////////////////////////////////////////////////////////////
% RESULTS 2 - Existent of abstract representations
%/////////////////////////////////////////////////////////////////////////////////////////

\subsection{Self-supervised transformers evolve abstract representations}
\label{sec:abstraction_existence}

\subsubsection{Introduction to the concept of abstraction} 

We hypothesize that the building blocks of abstract world models are \textit{abstractions}, short for abstract representations, that represent in symbolic form elements of the latent blueprint of an agent's environment. Inside biological and artificial neural networks, abstractions manifest as shared representations elicited in response to inputs that share a semantic feature but are otherwise distinct\cite{reed2016taxonomic}. As such, abstractions could constitute attractors (convergence in representations), which we believe might play a key role in the generalization of downstream computations. 

After showing that transformers can successfully master the board reconstruction task (section \ref{sec:parameter_tuning}), we set out to understand the computations that have developed through training to support the decision-making of such models. In particular, we asked whether these models evolve abstractions that encode elements from the HOD latent blueprint (this section), and if so, what is their role in the overall computation of the system (section \ref{sec:abstraction_manipulation}).

The computational pipeline of the studied transformer can be summarized as follows: (i) Masked boards are flattened and each token (background, object, or UNK.) is replaced by its corresponding learned token embedding; (ii) These embeddings are serially edited by each of the transformer blocks; (iii) The output embeddings are subsequently fed to the prediction head which produces probability distributions over a token vocabulary (see section \ref{subsec:architectures} for details). Therefore, understanding these systems primarily revolves around tracking and making sense of the alterations in latent token embeddings throughout the different computational stages of the network. According to the aforementioned hypothesis, we expect the embedding for certain tokens that are perceptually different (e.g.,  object tokens 1 and 2) yet semantically related (e.g., they are part of the same object) to share similar representations at specific computational stages. In our dataset, three examples of shared semantic features that could potentially be represented by abstractions are (i) the concept of background (vs. foreground, i.e., object), (ii) object membership (i.e., tokens are part of object A as opposed to B), and (iii) the relative position of a token within a parent object. We believe that these abstractions could function as labels allowing the system to generalize specific downstream computations to all tokens bearing them.

When looking for abstractions, one might expect a collection of point attractors in representational space. However, a deep learning network does not need such stringent convergence to attach the same meaning to various representations. Instead, the meaning could be attached to a cluster of representations which together form a lower dimensional manifold in representational space. For example, an abstraction could be defined by a specific activation range over a small set of units that can be read out by downstream layers. Therefore, we propose studying abstractions at two levels of granularity: (i) At the level of the entire embedding $z$: If the support of the abstraction is dominant, i.e., it involves most of the embedding, we might be able to see clustering amongst token embeddings based on semantic features. (ii) Representational subspace $\bar{z}$: If the support is small, clustering at the level of the entire embedding might not be visible. However, focusing on the meaningful part of $z$ while removing stronger sources of variation between embeddings might reveal the convergence. 

\subsubsection{Splitting token embeddings for analysis}

Given that our goal is to identify points of representational convergence, we first needed to understand what drove the majority of the variance amongst token embeddings $\{z_{i}\}_i$. To that end, we subjected a trained network to a large set of masked boards while recording activations throughout its layers. We pulled all token embeddings and performed a 2d principal component analysis (PCA, \cite{jolliffe2016principal}) of $\{z_{i}\}_i$ for a total of 12 computations stages. Specifically, we looked at four stages per block, including (i) the attention update \texttt{attn\_update}, (ii) the input of the MLP subblock \texttt{mlp\_in}, (iii) the MLP update \texttt{mlp\_update}, and (iv) the output of the block \texttt{z\_attn\_mlp} (see Sup. Figure \ref{supfig:computational_stages} for details on stages)\footnote{We prefix these stage names with \texttt{b0}, \texttt{b1}, or \texttt{b2} to indicate that they are part of transformer block 1, 2, and 3, respectively.}. We found that, at most stages, $\{z_{i}\}_i$ formed three distinct clusters comprised of masked background tokens, unmasked background tokens, and masked + unmasked object tokens, respectively (i.e., foreground, see Sup. Figure \ref{fig:pca_bck_for_mask_unmask}). The first two principal components (PCs) explained on average ~38\% of the variance and seemed to code for background vs. object tokens, and masked vs. unmasked, respectively. These findings, which were reproducible across runs, prompted us to consider the four token groups independently when looking for abstractions. However, we note that when performing the same PCA analysis on the background and object token separately, we could appreciate that embedding for masked and unmasked tokens progressively converged to finally overlap at the last computation stage.

When looking for abstractions, one might make a case for discarding unmasked tokens on the basis that their initial embeddings already contain the token ID information needed for prediction. The network could notably use the residual stream to propagate these initial representations unchanged to the prediction head, such that there would be no reason for them to evolve to encode shared semantics. While this argument has merit, we will see that this is not the case. Instead, we provide evidence showing that the network actively modifies the representations of unmasked tokens to support the predictions of masked ones (see section \ref{subsec:object_membership_abstraction}), making them prime candidates in the search for abstractions.

\subsubsection{Backgorund abstraction}

When solving the board reconstruction task, it would be advantageous for the network to differentiate between tokens that belong to the background and foreground (i.e., object), respectively. Therefore, we hypothesized that trained networks might label all background tokens with the same background abstraction, allowing downstream blocks to treat them as a whole and process them similarly. To test that hypothesis, we followed and compared the representations given to distinct background tokens throughout the network.

\begin{figure}[ht]
	\centering
	\includegraphics[width=0.7\textwidth]{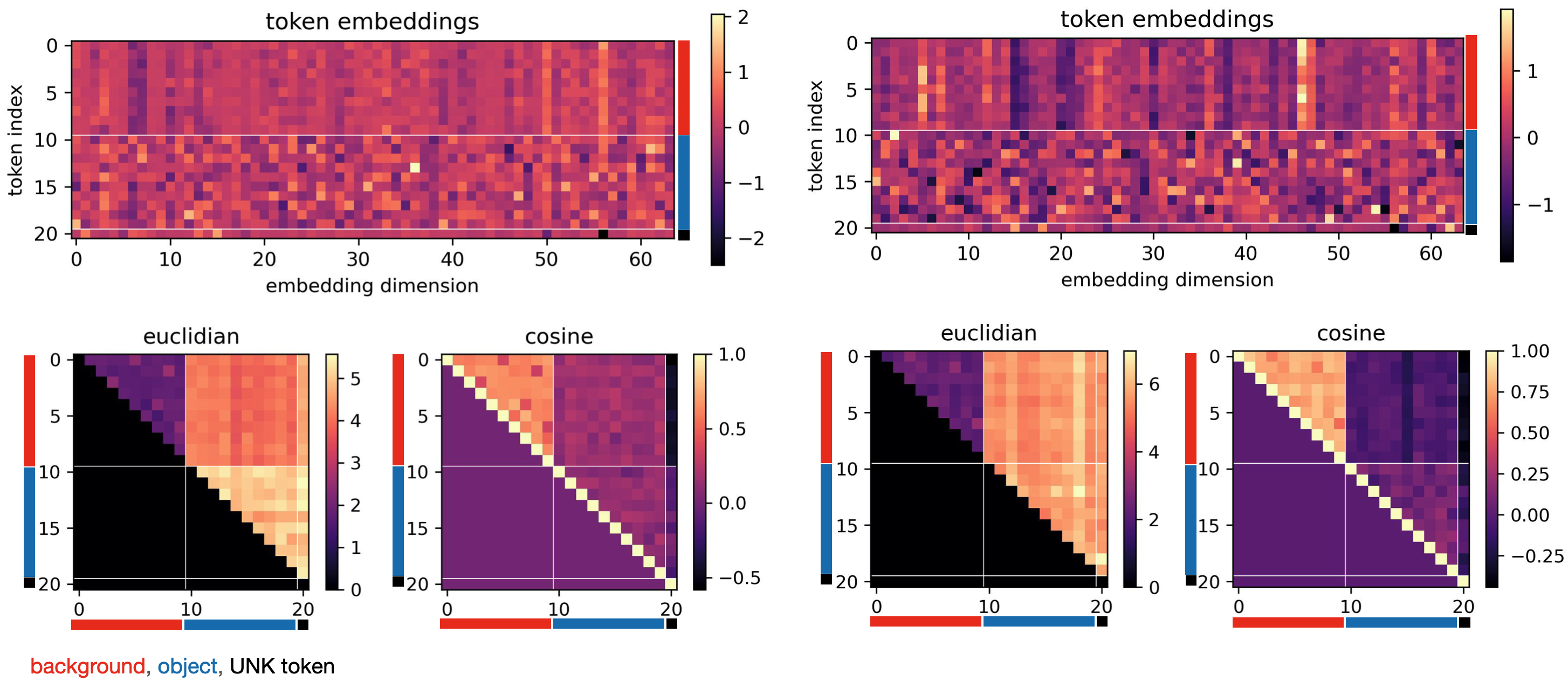}
	\caption{Similarity between initial token embeddings: $Z^0$ for two different runs (\textbf{left}, and \textbf{right}). \textbf{Top}: $Z^0$, each row is a token embedding. Token indices are as follows [0-9] background tokens, [10-19] object tokens, and 20 UNK. token (see blue and red color bars). \textbf{Bottom}: Matrix of pairwise Euclidean distances between embeddings (\textbf{left}) and matrix of cosine similarity between embeddings (\textbf{right}).}
	\label{fig:bck_abs_embeddings}
\end{figure}

Starting with the initial token embeddings $Z^0$, we found that embeddings for background tokens were consistently more similar to each other compared to embeddings for background vs. object and object vs. object tokens (Figure \ref{fig:bck_abs_embeddings}). This representational similarity suggests that, at the level of the initial token embeddings, the network has already sorted tokens into background and foreground.

\begin{figure}[ht]
	\centering
	\includegraphics[width=0.7\textwidth]{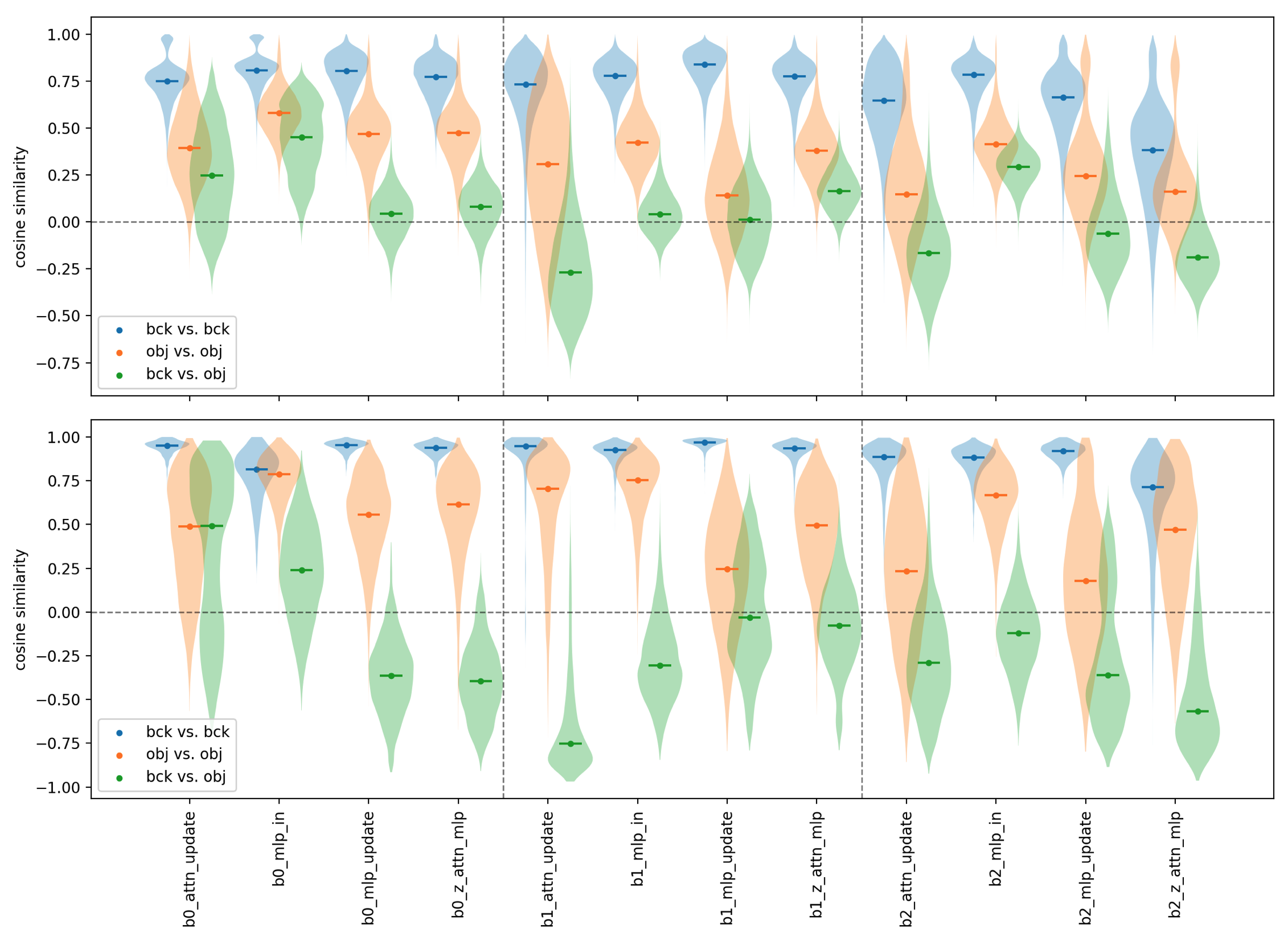}
	\caption{Pairwise cosine similarities across computational stages: 1000 random masked boards featuring each a single root object are fed to the network and representations for unmasked tokens (background and object) are collected at various computational stages. Representations are then sorted into background (bck) and object (obj) groups. Violin plots represent the distribution of pairwise cosine similarities between and within groups. Lines show the median for each violin. Comparisons are performed either at the level of the entire representation (\textbf{Top}) or on a rationally selected 10d representational subspace (\textbf{Bottom}).}
	\label{fig:bck_abs_violin}
\end{figure}

We then extended this analysis to 12 computational stages (Sup. Figure \ref{supfig:computational_stages}): The network was tasked with reconstructing a random collection of masked boards, and the representations of unmasked tokens were collected for each stage shown in Figure \ref{fig:bck_abs_violin}. We found that representations for pairs of background tokens remained significantly closer to each other, as measured by cosine similarity, compared to other pair types (background vs. object, object vs. object, Figure \ref{fig:bck_abs_violin}, top). The convergence ended at the last stage to facilitate the prediction of the token IDs.

Our ability to see a convergence in representation when comparing entire vector embeddings was fortunate. As mentioned before, abstractions are more likely to materialize as low-dimensional manifolds within the representational space. To test and illustrate this point, we attempted to better delineate the abstraction within the original 64-dimensional space and repeat the experiment on these lower-dimensional representations (Figure \ref{fig:bck_abs_violin}, bottom). Specifically, we identified at each computational stage the top 10 most informative units to distinguish between background and object token representations (selection based on mutual information). In comparison with results obtained when looking at the full representations, we found that background-to-background similarity measures significantly increased while object-to-object representation diverged across stages (Figure \ref{fig:bck_abs_violin}, bottom). Altogether, these results provide evidence for the existence of an attractor, where distinct tokens (i.e., different token IDs) converge in representation on the basis that they share the same semantic feature.  

\subsubsection{Object membership abstraction}
\label{subsec:object_membership_abstraction}

\begin{figure}[ht]
	\centering
	\includegraphics[width=0.75\textwidth]{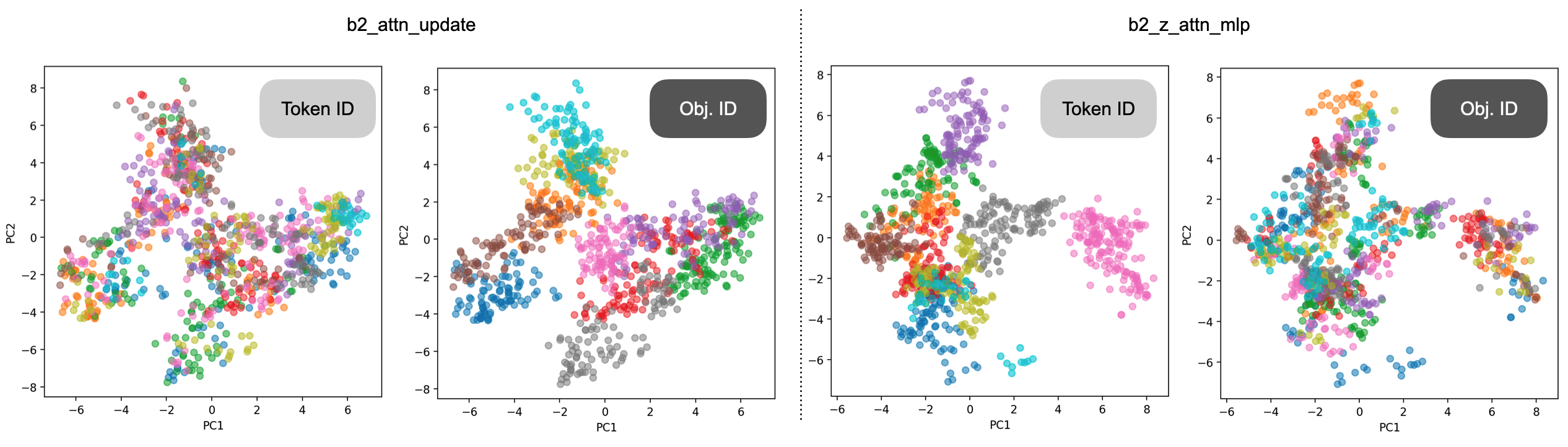}
	\caption{2d PCA projection of token embeddings at two computational stages (\textnormal{\texttt{b2\_attn\_update}}, \textbf{left}, \textnormal{\texttt{b2\_z\_attn\_mlp}}, \textbf{right}) color-coded by either token ID or object membership (Obj. ID): 1000 random masked boards featuring each a single root object were fed to the network and representations for unmasked object tokens were collected at various computational stages. PCA was performed for each stage after normalizing of embeddings.}
	\label{fig:obj_abs_pca_joint}
\end{figure}

In our dataset, object tokens are composed together to create specific root objects (see section\ref{subsec:dataset}). Accordingly, we asked whether distinct tokens that are part of the same object would transiently converge in representation to what could be considered an \textit{object membership} abstraction. In contrast with the background abstraction, we found no specific clustering amongst the initial embeddings of object tokens at $Z^0$ (Figure \ref{fig:bck_abs_embeddings}). This is easily explained by the fact that object tokens can participate in more than one object prototype, making it impossible for the network to encode object membership a priori. Next, we used 2d PCA projections to visualize the representations of unmasked object tokens at various stages of the computation as a trained network processed a collection of masked boards. Color coding the token representations based on token ID or object membership revealed some interesting patterns (Figure \ref{fig:obj_abs_pca_joint}): Representations coming out of the last transformer block (\texttt{b2\_z\_attn\_mlp}) heavily clustered based on token ID, most likely to facilitate inference at the prediction head. More interestingly, we were able to see the clustering of token representations by object membership at the last attention subblock (\texttt{b2\_attn\_update}). Extending the analysis from 2d to the full representational space (64d), further confirmed that tokens that are part of the same object converge in representation at key computational stages (Sup. Figure \ref{supfig:obj_abs_rsm_obj}). Across multiple runs, convergence was the strongest at the attention layers, aligning with their role in integrating contextural information to determine object identity\footnote{The analysis reported above provides clear evidence of the fact that the network edits the representations of unmasked tokens to support the prediction of masked ones. Again, this is non-trivial as the network could use its residual stream to simply propagate to the decision head the token ID already present in the initial embeddings.}.

\begin{figure}[ht]
	\centering
	\includegraphics[width=0.8\textwidth]{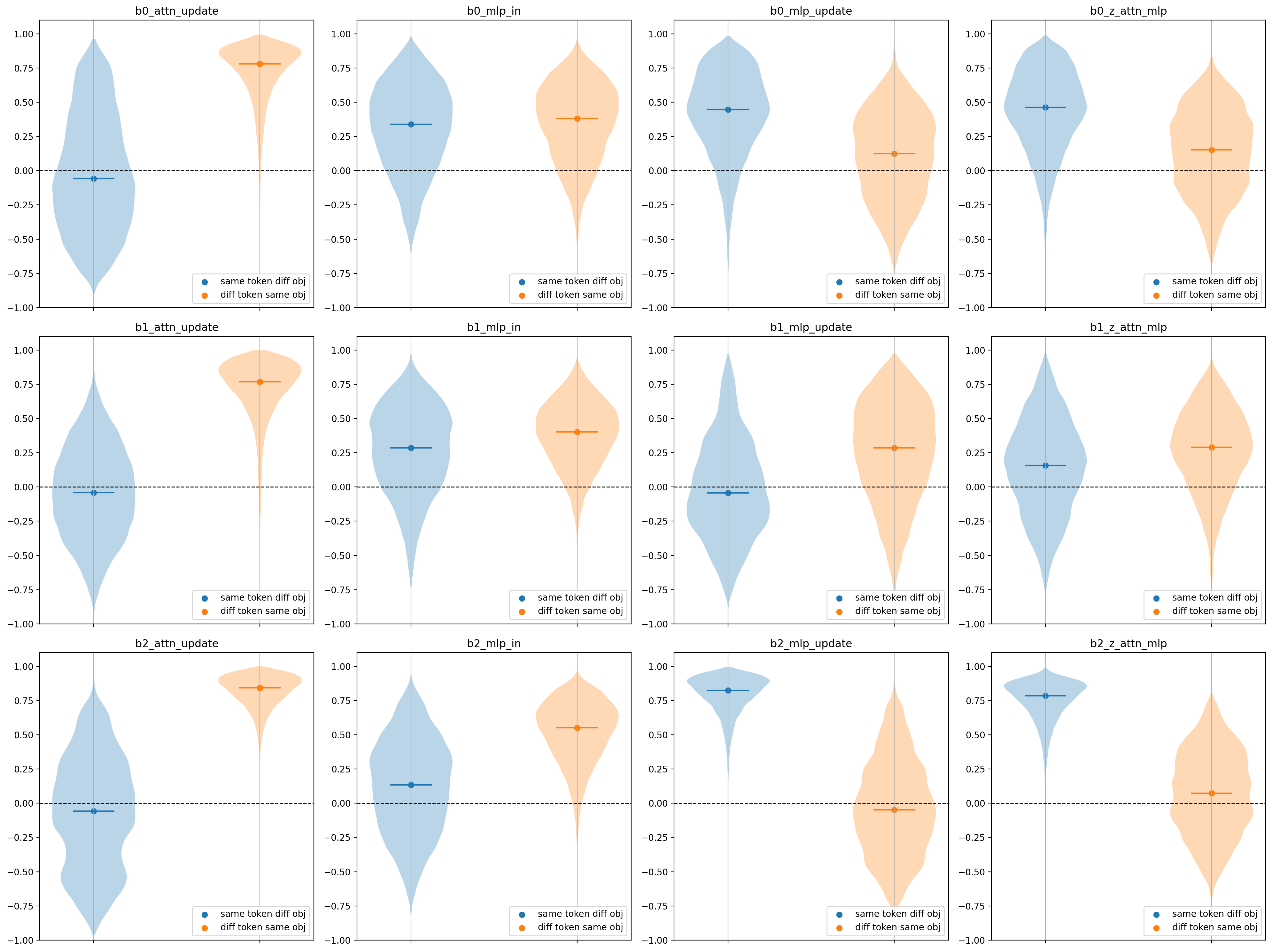}
	\caption{Distribution of pairwise cosine similarities between object tokens sharing token ID but participating in different objects (same token diff obj, blue) and tokens that are different but part of the same objects (diff token same obj, orange). Representations are restrained to the top 10 most informative units for object membership as measured by mutual information.}
	\label{fig:obj_abs_violin_stringent}
\end{figure}

We have constructed our datasets such that the same object token would appear in distinct root objects (see section \ref{subsec:dataset}). This allows us to compare the representational similarity of tokens that share the same token ID but are part of different objects with those of tokens sharing object membership but having different token IDs (Figure \ref{fig:obj_abs_violin_stringent}). We found that at computation stages where the convergence is the strongest, tokens that started with the same embedding (share token ID) but participated in different objects diverged in representations, while the representations of tokens that started with different embeddings but shared object membership converged. This finding was found to be consistent across runs (Sup. Figure \ref{supfig:obj_abs_cohenD_linearprobe}, top).

Our findings paint a picture where object membership abstractions appear in the attention layers of the network, materializing as a partial convergence in representation. Contrasting with that result, we also found that object membership information can be read out from unmasked token embedding with near-perfect accuracy from the first attention update onward (Sup. Figure \ref{supfig:obj_abs_cohenD_linearprobe}, bottom). Together, these results suggest there might exist a qualitative difference between representations found at various stages. In particular, we hypothesize that only a subset of these representations that can be probed might be actively participating in shaping the network’s computations. (see section \ref{sec:abstraction_manipulation}).

\textbf{Learning pressure}: In our dataset, root objects appear either independently from each other or together as part of a larger composite object (see section \ref{subsec:dataset}). Therefore, the object abstractions discovered above could emerge as a result of two possible “learning pressures”: (i) The object membership abstraction emerges from the fact that specific tokens form a consistent pattern (root object); (ii) The abstraction emerges because together, the tokens of a root object form a unit that helps predict other constituent roots within a composite object. To help tease out which learning pressure is responsible for the emergence of object membership abstractions, we repeated the analysis conducted in this section with networks trained on impoverished datasets that do not feature composite objects. We found that in the absence of composites, clear object abstractions still emerge, suggesting that the organization of tokens into consistent patterns (root object) is sufficient for the network to develop corresponding abstractions (data not shown).

\subsubsection{From instances to class prototypes}

\begin{figure}[ht]
	\centering
	\includegraphics[width=0.7\textwidth]{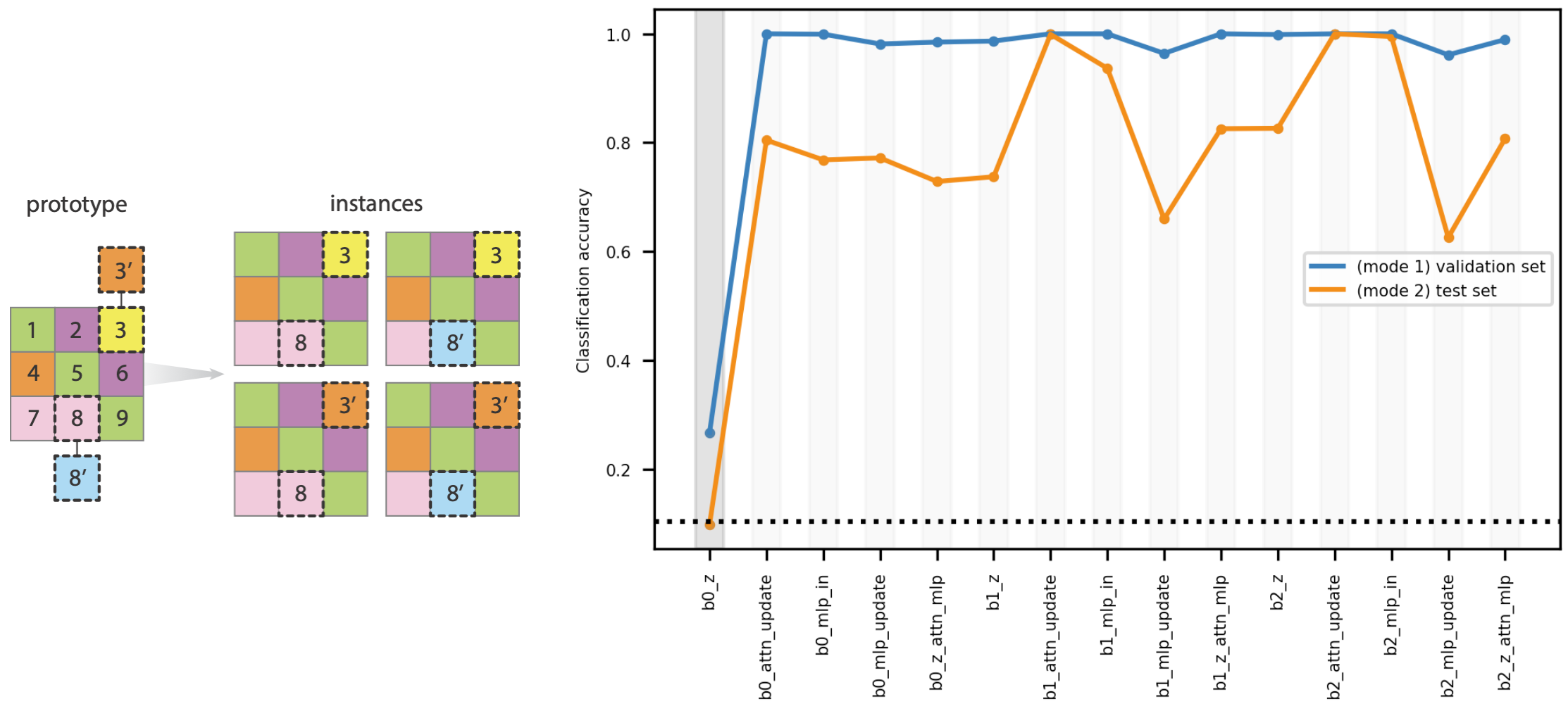}
	\caption{\textbf{Left}: Schematic describing “fuzzy” objects and their instantiation process. \textbf{Right}: Linear probe for object membership generalizes across object instances. 1000 random boards, generated from a dataset featuring 10 different fuzzy objects (two instances per object, i.e., modes), were fed to the network and activations were recorded. Unmasked object tokens were separated based on the instance they participated in to form a training (mode 1) and test set (mode 2). We then trained a linear multi-class classifier to predict object classes on the training set and evaluated accuracies on validation (blue) and held-out test (orange) sets.}
	\label{fig:obj_abs_fuzzy}
\end{figure}

We have so far focused exclusively on abstractions for objects defined as deterministic arrangements of specific tokens. However, this deterministic framework does not capture the variability of object categories encountered in nature. Real-life objects are better conceptualized as instances belonging to a particular object class (e.g., different individual cats belonging to the general category of 'cat'). To test whether a transformer subjected to a collection of class instances would form an abstraction to talk about the class itself, we extended our dataset to feature object classes defined as collections of slightly different instances. Specifically, we introduced “fuzzy” objects by allowing certain tokens within an object prototype to take one of $k$ possible token identities (modes, Figure \ref{fig:obj_abs_fuzzy}, left). When the fuzzy object is selected to be displayed on the board, an instance of that object is generated by randomly selecting amongst modes. 

After training a network on such a dataset, we tested whether a linear probe trained to classify object membership using only one set of instances could generalize to a held-out set with the remaining instances. We found that the linear probes generalized perfectly at attention updates (\texttt{b1\_attn\_update}, \texttt{b2\_attn\_update} in Figure \ref{fig:obj_abs_fuzzy}, right). In line with our results showing that representational convergence based on object membership is the strongest at attention subblocks, this finding suggests that the network develops a similar object class abstraction which is shared across instances. Altogether, these results show that object membership abstraction extends to object categories.

\subsubsection{The case of masked tokens}

\begin{figure}[ht]
	\centering
	\includegraphics[width=0.65\textwidth]{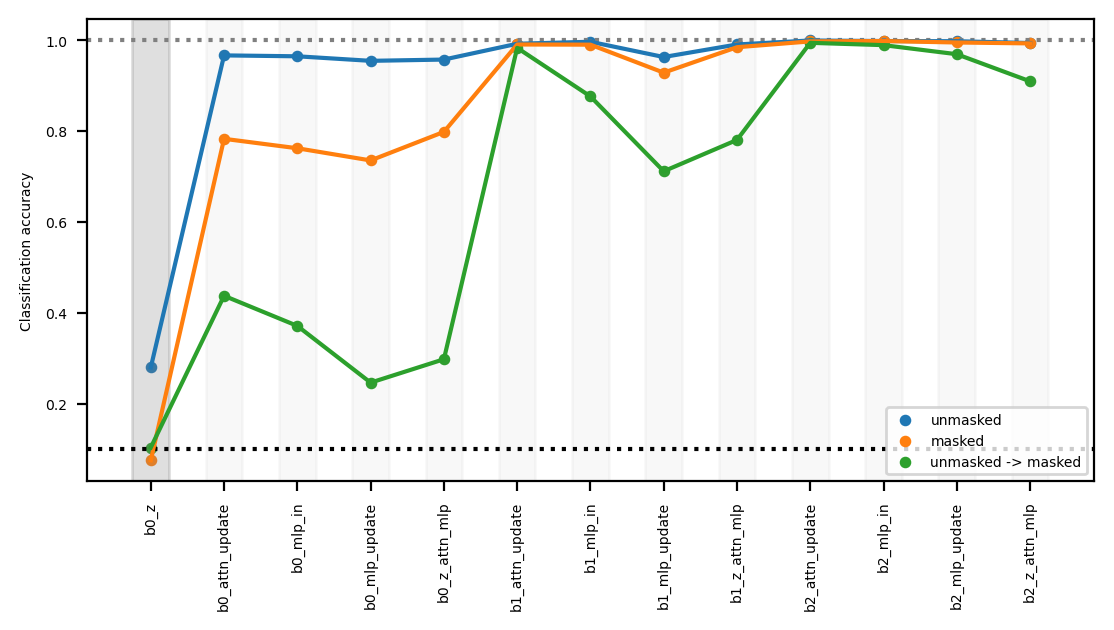}
	\caption{Linear probe for object membership generalizes from unmasked to masked token embeddings. We generated a collection of 1000 random boards, each with a single root object and 25\% masking. We collected token embeddings across 12 computational stages separating representations for unmasked and masked tokens. We then trained linear probes to read out object membership from unmasked tokens and evaluated them on a held-out validation set of unmasked tokens (blue). We did the same for masked tokens (orange). Finally, we report the classification accuracies of the unmasked probe on masked token embeddings (green).}
	\label{fig:obj_abs_maskedTokens}
\end{figure}

Our analysis has so far focused on unmasked tokens, as we endeavored to show that the network updates their embeddings to encode abstractions. However, the main function of the network is to modify the representations of masked tokens to accurately predict their true identity. Accordingly, we repeated our analysis with a focus on masked tokens. Specifically, we wanted to find out if the representations of masked tokens also included object abstractions, and if so, whether they were the same as the ones we found when analyzing the embeddings of unmasked tokens. To answer these questions, we employed the same probing generalization methodology that we used earlier. At each computational stage, we trained a linear classifier to identify object membership based on unmasked token embeddings and later tested generalization on the representations of masked tokens (Figure \ref{fig:obj_abs_maskedTokens}). We observed that the probes got better at generalizing to masked token embeddings as we progressed through the network layers. In particular, we found perfect generalization accuracies at the attention layers where we previously saw representational convergence (Figure \ref{fig:obj_abs_violin_stringent}). This finding not only suggests that masked tokens also feature object membership abstractions but that the network uses the same abstractions for masked and unmasked tokens.

%/////////////////////////////////////////////////////////////////////////////////////////
% RESULTS 3 - Abstractions are key to the network's inference process
%/////////////////////////////////////////////////////////////////////////////////////////

\subsection{Abstractions are key to the network's inference process}
\label{sec:abstraction_manipulation}

In section \ref{sec:abstraction_existence}, we have seen that linear probes trained on token embedding can be used to identify representations whose expression correlates with specific semantic features of the dataset (e.g., object membership abstractions, Sup. Figure \ref{supfig:obj_abs_cohenD_linearprobe}). Based on this result, it would be tempting to conclude that the network uses these representations to “talk” about the said semantic features and alter its decision-making based on their presence or not. However, our ability to read out that information merely suggests that the trained network has learned to generate and use abstractions. It does not constitute proof that these abstract representations are causal, i.e., the network actively uses them to support its inferences\cite{geiger2021causal}. Instead, the representations delineated by our probes could very well be byproducts that play no role in the overall computation, and as such carry no meaning for the network. In this section, we seek to demonstrate that a subset of the probed abstractions are both causal (i.e., necessary for inference) and meaningful (i.e., encode the semantic feature of interest) to the network. To robustly identify whether a representation is causal as opposed to merely correlative, manipulation experiments are essential\cite{geiger2021causal}. These involve perturbing the representation and observing the subsequent change, if any, in the network's behavior. In addition, if one wishes to speculate about the meaning of a specific representation for the network, one should demonstrate that rationally crafted gain-of-function manipulations produce predictable changes in the network's output.

\subsubsection{Methodology}

\begin{figure}[ht]
	\centering
	\includegraphics[width=0.65\textwidth]{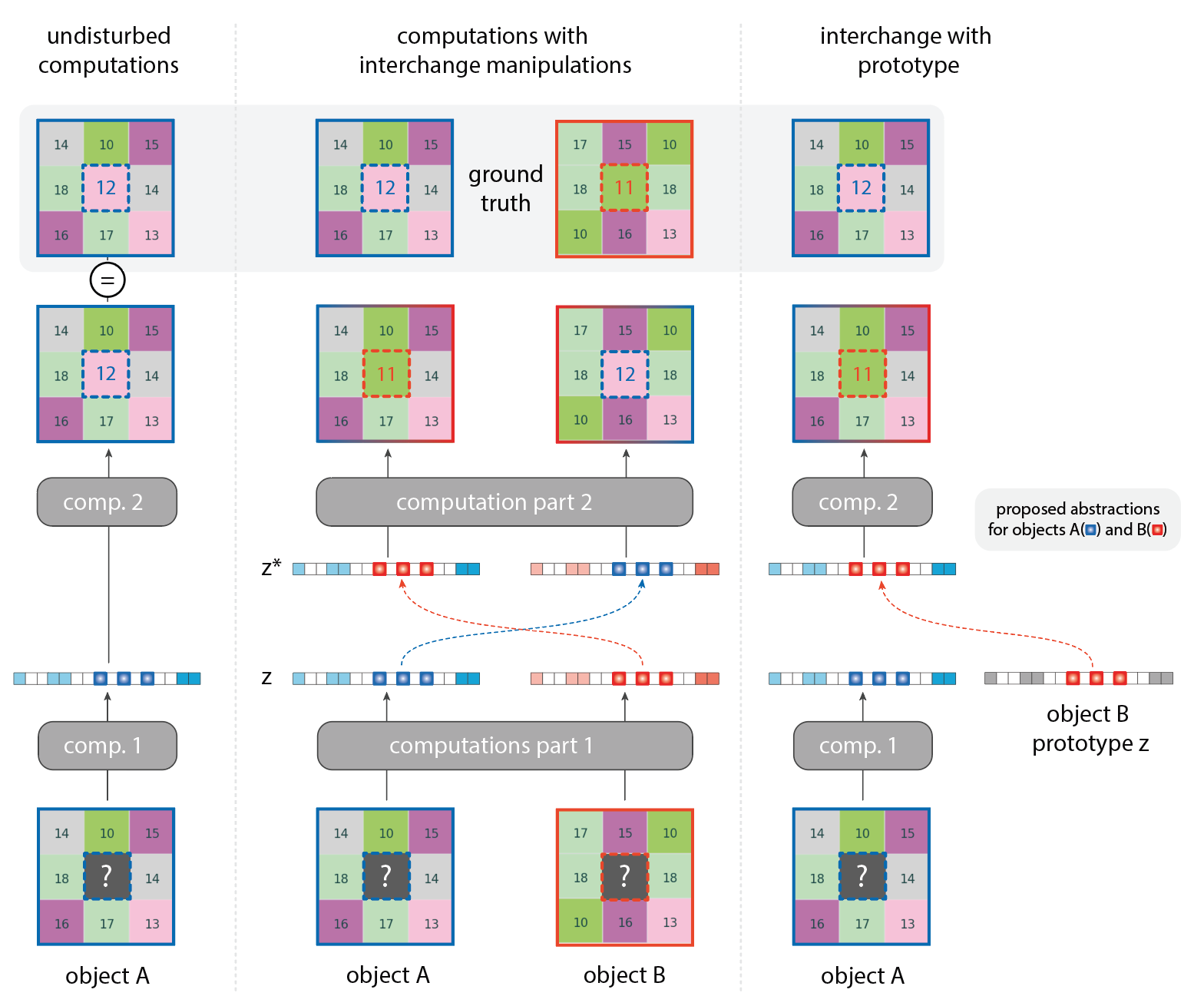}
	\caption{Schematic of the gain-of-function experiment used to demonstrate that abstractions are causal and meaningful to the network. \textbf{Left}: Unaltered processing of a board with a single masked token at the center of a root object produces the correct output. The first part of the computation (comp. 1) produces the intermediate embedding $z$ for the masked token, which is then processed by the remainder of the network (comp. 2) to generate the correct guess. \textbf{Center}: Two boards are processed in parallel. After comp. 1, the embeddings for the masked tokens are edited to swap segments that code for the “object A” and “object B” abstractions, respectively. The output of the network is predictably wrong and consistent with the network being tricked into seeing the wrong object. \textbf{Right}: Similar manipulation experiment the swapped object membership abstraction is taken from a prototype embedding for object B, thus removing the confound of swapping token ID information between embeddings.}
	\label{fig:manip_methodology}
\end{figure}

To test whether candidate abstractions are causal and meaningful, we conducted \textit{interchange intervention experiments}\cite{geiger2023finding} where representations coding for distinct objects (A and B) are swapped while we record the effect on the network’s predictions. To illustrate the methodology, let us consider the case where the network is asked to reconstruct a board for which the central token of object A is masked (Figure \ref{fig:manip_methodology}, left). We have previously shown that, as processing progresses, an abstraction correlating with the semantic feature “object A” appears in the embedding $z$ of the masked token. If this abstraction is causal, then removing that representation should significantly impair the network's ability to correctly infer the true token ID. Additionally, if the abstraction truly signifies "object A" to the network, it should be possible to alter the network's output predictably by replacing the abstraction for object A with the abstraction for a different object B (Figure \ref{fig:manip_methodology}, center). To perform such a gain-of-function experiment, we first delineate candidate representations that we believe encode object membership abstractions (objects A and B) at a specific computational stage. We then ask the network to independently reconstruct the following two boards: Board 1 features a single randomly positioned root object A on a random background, and a single token of the object is masked; Board 2 follows an identical layout but object A is replaced by object B. The position of the masked token is identical between the two boards. During network inference, we record the activations corresponding to the masked token embeddings $\mathbf{z}_A$ and $\mathbf{z}_B$. Finally, we repeat the forward pass while manipulating the activations: computations are halted at the stage of interest, and the abstractions coding for objects A and B on $\mathbf{z}_A$ and $\mathbf{z}_B$ are swapped. We then resume processing on these modified embeddings and examine the logits produced by the network for both masked tokens. We conclude that the abstractions for objects A and B are causal and carry the proposed meaning if the network’s predictions for the manipulated masked tokens are switched, i.e., $P(t=t_B|\mathbf{z}_A^*) \approx 1$ and $P(t=t_A|\mathbf{z}_B^*)\approx 1$, where $t_A$ and $t_B$ are the ground truth token IDs behind masked tokens in A and B, and $\mathbf{z}_A^*$ and $\mathbf{z}_B^*$ are the corresponding edited embeddings. Our approach is targeted to a single computational stage at a time, hence allowing us to attribute the effect to specific layers. Within each block, we focus on the stages shown in Sup. Figure \ref{supfig:computational_stages}.

\subsubsection{Unit-based approach} 
\label{subsec:unit_based_approach}

\begin{figure}[ht]
	\centering
	\includegraphics[width=0.75\textwidth]{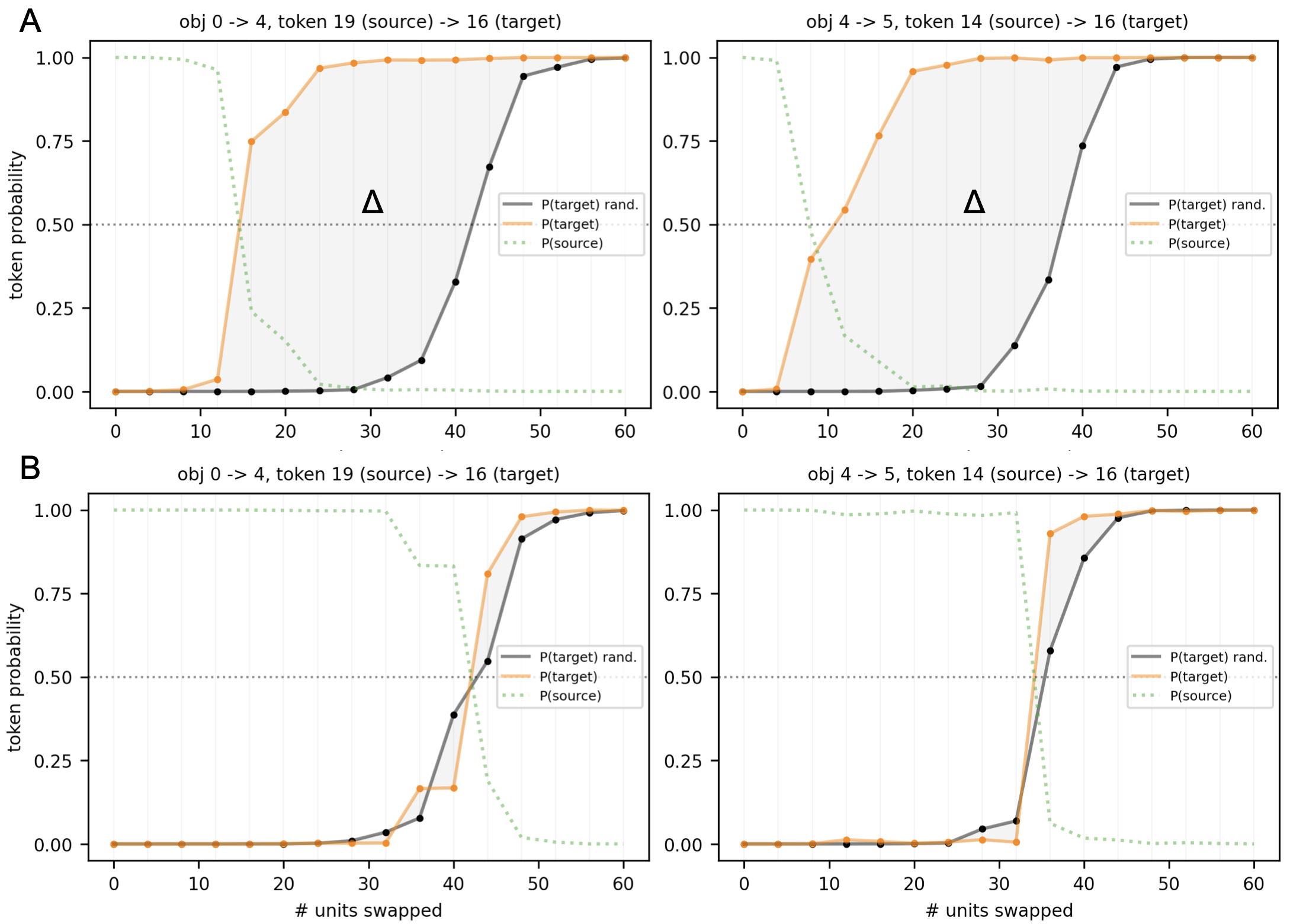}
	\caption{Effect of unit swapping manipulations on network's prediction. (\textbf{A}) Results for two different manipulation experiments both conducted at stage \texttt{b2\_mlp\_in}. The black curve shows mean $P(t=t_B|\mathbf{z}_A^*)$ when swapping activations over a random set of $n$ units. Orange and dashed green curves show $P(t=t_B|\mathbf{z}_A^*)$ and $P(t=t_A|\mathbf{z}_A^*)$ when the manipulation is carried out over a rationally selected set of $n$ units (targeted at object membership abstractions). Shaded regions show $\Delta$. (\textbf{B}) Same as (\textbf{a}) but units are ranked based on their importance in encoding token ID information.}
	\label{fig:single_manip_joint}
\end{figure}

We started with a unit-based approach to swapping object membership abstractions, where we interchange the activation values of $\mathbf{z}_A$ and $\mathbf{z}_B$ over a specific subset of units ($s$). To delineate the subset of units that best encodes the abstractions, we train a linear probe to distinguish between tokens belonging to object A and object B and use the absolute value of the decision boundary vector to rank the units from most to least relevant\footnote{Alternatively, we perform ranking based on mutual information.}. We can then compare the effect on the network’s predictions (i.e., $P(t=t_B|\mathbf{z}_A^*)$) of swapping activations over the top $n$ ranking units (orange curve in Figure \ref{fig:single_manip_joint}) with the average effect obtained when considering random sets of equal cardinality (black curve in Figure \ref{fig:single_manip_joint}). The latter serves as an empirical baseline to which we can compare the effect of swapping a rationally selected set of units: If the true causal abstraction is encoded over a subset $s^*_k$ of units, then swapping activations over $s^*_k$, as opposed to a random $s_k$ should lead to a greater increase in $P(t=t_B|\mathbf{z}_A^*)$. To obtain an overall value reflecting the importance of the abstraction manipulated, we compute $\Delta$, the signed area between the orange and black curves. The more positive the value the more evidence we have to claim that the abstractions for objects A and B are causal and carry the hypothesized meaning for the network. 

Figure \ref{fig:single_manip_joint}a shows the effect of two manipulations that were both conducted right after the third attention layer (\texttt{b2\_mlp\_in}). In both cases, we found that swapping the activations of the top 10-20 units was sufficient to trick the network into producing the wrong output. In comparison, it took on average 40 units to obtain the same effect when selecting random sets of units. Interestingly, we found that when repeating the experiment while ranking the units based on their importance encoding token ID information rather than object membership, the overall manipulation effect was much weaker (Figure \ref{fig:single_manip_joint}b). This result adds credibility to the idea that the change in output is indeed driven by the manipulation of object membership abstractions rather than token ID information. In particular, it suggests that, at this point in the network’s computations, object membership is more relevant than token ID information to infer the hidden identity of the masked token.

\begin{figure}[ht]
	\centering
	\includegraphics[width=1.\textwidth]{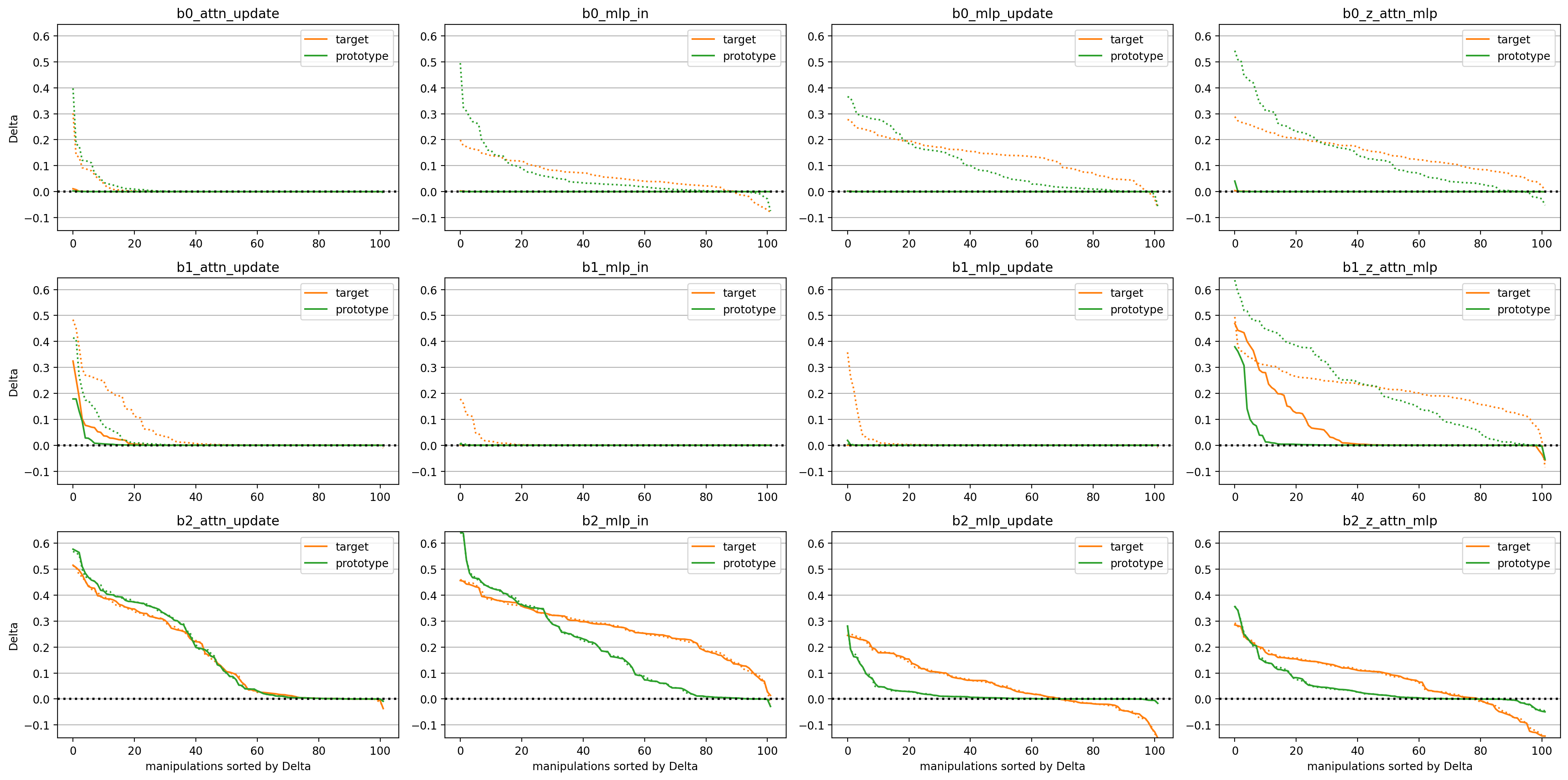}
	\caption{Effect of manipulating object membership abstractions across cases and computational stages. Aggregated manipulations results for each of 12 computational stages (3 transformer blocks, rows, 4 computational stages, columns). Solid and dotted lines show single token edits and bulk object edits, respectively. Orange and green colors indicate swapping with a target embedding ($\mathbf{z}_B$) or object prototype embedding ($z_{\bar{B}}$), respectively.}
	\label{fig:many_manip_singlevsbulk}
\end{figure}

In Figure \ref{fig:single_manip_joint} we detailed the effect of two individual manipulations, performed at a specific computational stage, editing the embedding of one particular token within the object, and specifically swapping object membership A to B. To delineate causal abstraction network-wide, we extended the analysis within a computational stage, looking at more instances of manipulation, as well as across computational stages (Figure \ref{fig:many_manip_singlevsbulk}). In particular, we looked at the distribution of $\Delta$ values (solid orange lines) across computation stages. We found the strongest effect in the \texttt{attn\_update} and \texttt{mlp\_in} stages of the last block. This result was in line with findings reported in section \ref{subsec:object_membership_abstraction} showing maximum convergence in representations in the last attention layer (Figure \ref{fig:obj_abs_violin_stringent}). Interestingly, we found a lesser effect overall when manipulating embeddings right before the decision head (\texttt{b2\_z\_attn\_mlp}). This finding suggests that the network transiently uses the object membership abstractions to support its inferences, after which the information loses computational relevance\footnote{Similar results were obtained across different runs (Sup. Figure \ref{supfig:many_manip_severalRuns}).}. 

Because we are editing $\mathbf{z}_A$ by borrowing activations directly from the target embedding $\mathbf{z}_B$, the manipulation effects observed could simply be due to us swapping token ID information between embeddings. This is notably more likely to happen if the network encodes token ID and object membership on the same set of units (superposition\cite{elhage2022superposition}). To rule out that possibility and show that the observed effects are specific to a change in object membership abstraction, we conducted a more stringent version of the interchange intervention experiment where the object B abstraction was taken from a prototype embedding $\mathbf{z}_{\bar{B}}$ rather than the target token embedding $\mathbf{z}_B$ (Figure \ref{fig:manip_methodology}, right). $\mathbf{z}_{\bar{B}}$ was obtained by averaging the embedding of a large collection of object B tokens, thus removing any token ID information. While the change caused a slight reduction of the manipulation effects, the conclusion remained the same (Figure \ref{fig:many_manip_singlevsbulk}, dotted orange lines). Additionally, we showed that directly swapping token ID information between $\mathbf{z}_A$ and $\mathbf{z}_B$, rather than object membership, had little effect around the last attention layer (\texttt{b2\_attn\_update}) and only became critical after the last MLP (Sup. Figure \ref{supfig:many_manip_objvtokenID}). 

Figure \ref{fig:many_manip_singlevsbulk} shows that the causal object abstractions are concentrated on the last attention subblock. However, our results from the section \ref{subsec:object_membership_abstraction}, indicated the presence of object membership abstractions (strong convergence in representation) at earlier computational stages too. While these abstractions might solely be correlative, a more exciting interpretation for this discrepancy is the \textit{self-healing} hypothesis, which refers to the network’s ability to correct corrupted representations\cite{mcgrath2023hydra}. In our particular case, because we only edit a single token within the object, the network can potentially use the intact part of that object to overwrite our edits at subsequent attention layers. To test that eventuality, we extended our manipulations to simultaneously edit all tokens in the object (Figure \ref{fig:many_manip_singlevsbulk} and Sup. Figure \ref{supfig:many_manip_severalRuns}, bulk edits). As expected, we found that when no additional attention layers were present downstream, bulk editing did not improve the effect. However, we observed a significant gain when bulk editing was performed at earlier stages. These results confirm the existence of meaningful causal abstractions early in the computations and show that the studied transformer has evolved computational redundancies that allow for self-healing.

\subsubsection{Unit-agnostic approach}
\label{subsec:unit_agnostic_approach}

Our manipulations so far consisted of swapping activations along a specific set of units. However, when performing its computations the network does not need to “think” in terms of units. Instead, it is far more likely to operate in a unit-agnostic fashion, reasoning at the level of the entire representational space. Notably, the network could encode a particular semantic feature (e.g., object membership) as a low-dimensional manifold, which, if not lined up with the base formed by the physical units of the network, would appear encoded over a far larger number of dimensions. In that case, transforming the embeddings from the physical base $\mathcal{B}$ defined by the units to a more appropriate base $\mathcal{B}'$ could, in theory, give us similar manipulation effects while changing a lower number of features.

Assuming that our manipulation aims at swapping object membership abstractions (object A → B), we propose to create one such optimal base $\mathcal{B}'$ using the following iterative process: (i) Collect the d-dimensional token embeddings from objects A and B to create a training set $\mathbf{Z}^{i=d}$ (matrix of embeddings) for the binary classification $z \rightarrow \{A, B\}$. (ii) Train a binary classifier over $\mathbf{Z}^i$ and use the weight vector $\mathbf{w}$ as a basis vector for $\mathcal{B}'$. (iii) Project $\mathbf{Z}^i$ into the null space of  $\mathbf{w}$ to get the (i-1)-dimensional embeddings $\mathbf{Z}^{i-1}$. (iv) Repeat steps ii and iii until a full base has been constructed. 

\begin{figure}[ht]
	\centering
	\includegraphics[width=0.85\textwidth]{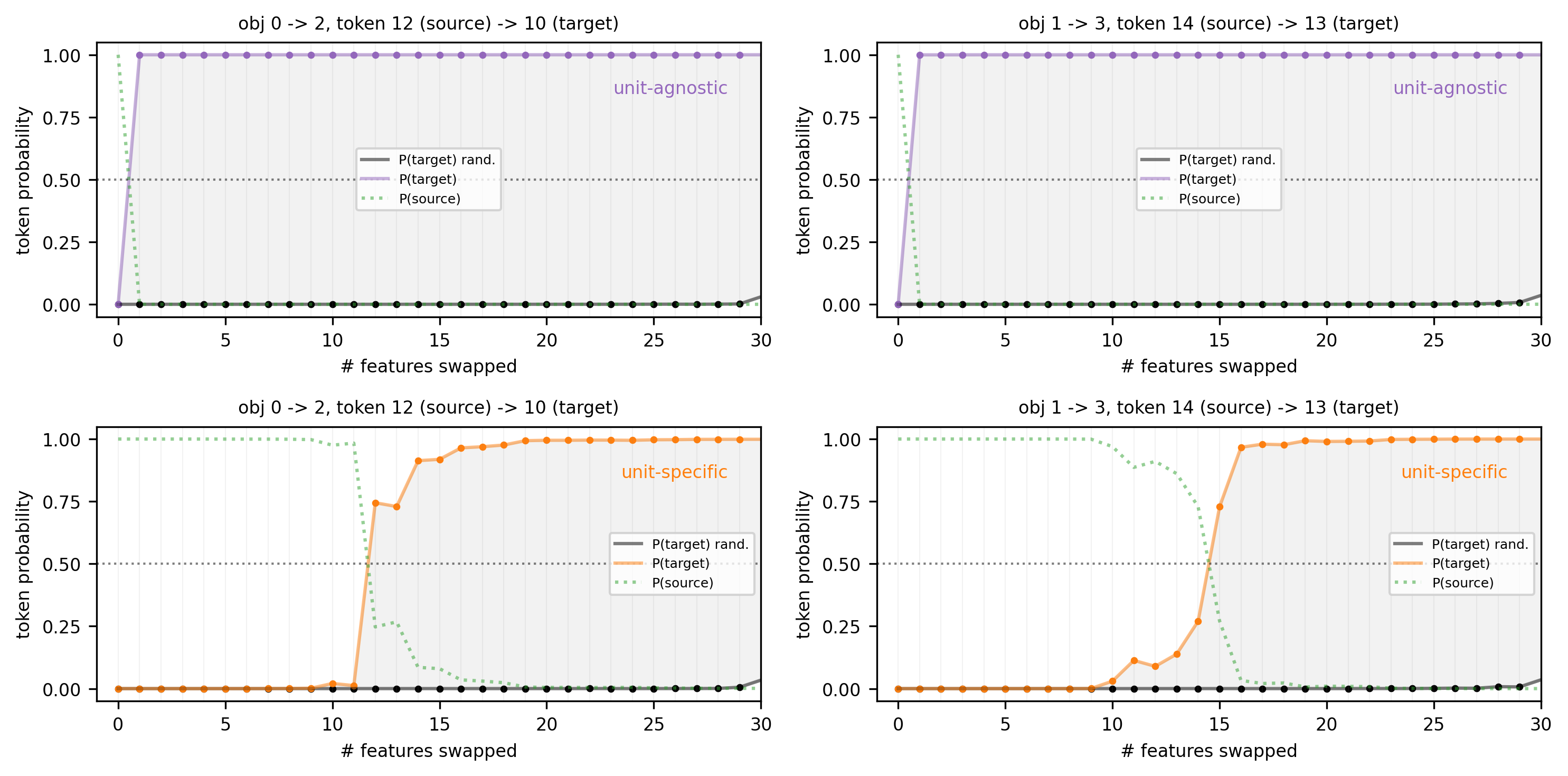}
	\caption{Comparison of unit-based vs unit-agnostic manipulation methods. \textbf{Left} and \textbf{right} show two different manipulation cases conducted at stage \textnormal{\texttt{b2\_mlp\_in}}. \textbf{Top}: effect of manipulating 0-30 features in the rotated $\mathcal{B}'$ base (unit-agnostic method). \textbf{Bottom}: effect of manipulating 0-30 features in the original base (unit-specific method). In both cases, we followed the strategy presented in Figure \ref{fig:manip_methodology}, and the ranking of the units was made based on their estimated role in coding object membership abstractions. The black curve shows the average $P(t=t_B|\mathbf{z}_A^*)$ when swapping activations over a random set of $n$ dimensions. Purple/orange and dashed green curves show $P(t=t_B|\mathbf{z}_A^*)$ and $P(t=t_A|\mathbf{z}_A^*)$ when the manipulation is carried out over a rationally selected set of $n$ dimensions. Shaded region shows $\Delta$.}
	\label{fig:unit_agnostic_manip}
\end{figure}

Repeating our manipulation experiment in $\mathcal{B}'$, we found that fewer features needed swapping for the network to be tricked into predicting the target rather than the ground truth source token (Figure \ref{fig:unit_agnostic_manip}). In fact, in the majority of the cases when the network could be tricked, swapping a single feature often gave the desired effect ($P(t=t_B|z^*) \geq 0.5$). In contrast, we find that an average of 18 units was required to get a similar effect while performing unit-based manipulations (Sup. Figure \ref{supfig:unit_agnostic_bars}). Altogether, these results support the notion that the network encodes object membership abstractions as low-dimensional manifolds, which do not necessarily align with the physical units of the network. Additionally, our ability to edit abstractions with as little as one feature swap suggests that unit-agnostic methods could allow for more precise manipulations, thereby reducing the risk of unintended off-target effects. Substantiating that claim, we notably show in section \ref{subsec:compositional_organization} that unit-agnostic 1d edits can be used to specifically alter one of two abstractions encoded on the same token embedding. 

%/////////////////////////////////////////////////////////////////////////////////////////
% RESULTS 4 - Compositional nature of abstractions
%/////////////////////////////////////////////////////////////////////////////////////////

\subsection{Compositional organization of abstractions}
\label{sec:compositionality}

After showing that abstractions exist (section \ref{sec:abstraction_existence}) and play a critical role in the network's computations (section \ref{sec:abstraction_manipulation}), we studied the nature of their interactions. In particular, we sought to answer the following two questions: (i) Is there representational independence\footnote{Also referred to as factorization\cite{bengio2013representation} or contextual independence\cite{murty2022characterizing, dziri2023faith}} between abstractions encoding independent semantic features? (ii) If an object is constructed by the composition of smaller objects, is this part-whole hierarchy reflected at the level of abstractions, i.e., is the abstraction of the whole derived from the abstractions of the parts or is it constructed as a stand-alone separate entity? For both questions, we conducted qualitative case studies that we report in sections \ref{subsec:representational_independence} and \ref{subsec:compositional_organization}, respectively. Representational independence and compositional organization are key properties of human language and reasoning, and as such, are regarded as essential ingredients for building AI systems that are both more interpretable and possess greater generalization abilities\cite{bengio2013representation, murty2022characterizing, szabo2012case, lake2017building}.

\subsubsection{Representational independence}
\label{subsec:representational_independence}

To study representational independence, we looked for two semantically independent features, whose abstractions could, in theory, be encoded on the same token embedding. We ultimately chose to focus on representations of \textit{object membership} (studied in section \ref{sec:abstraction_manipulation}), which indicate what object a token belongs to, and representations of \textit{relative spatial orientation} (RSO), which give the position of that token within the object. We reasoned that such RSO abstractions could be valuable to the network as knowing both object membership and RSO for a given token is sufficient to infer token ID.

Probing for RSO along computational stages, we found that this information could be linearly decoded from masked token embeddings from stage \texttt{b1\_mlp\_in} onward except \texttt{b2\_attn\_update} (Sup. Figure \ref{supfig:rep_ind_rso_probeAndManip}, top). We then refined our delineation by determining which of these representations were causal. We conducted manipulation experiments where RSO abstractions were edited and found strong effects at \texttt{b1\_z\_attn\_mlp} and \texttt{b2\_mlp\_in} (Sup. Figure \ref{supfig:rep_ind_rso_probeAndManip}, bottom). Based on these curves and results from our manipulations of object membership abstractions (Figure \ref{fig:many_manip_singlevsbulk}), we chose to test for independence at \texttt{b2\_mlp\_in}, where both object and RSO abstractions were found to be causal.

\begin{figure}[ht]
	\centering
	\includegraphics[width=0.5\textwidth]{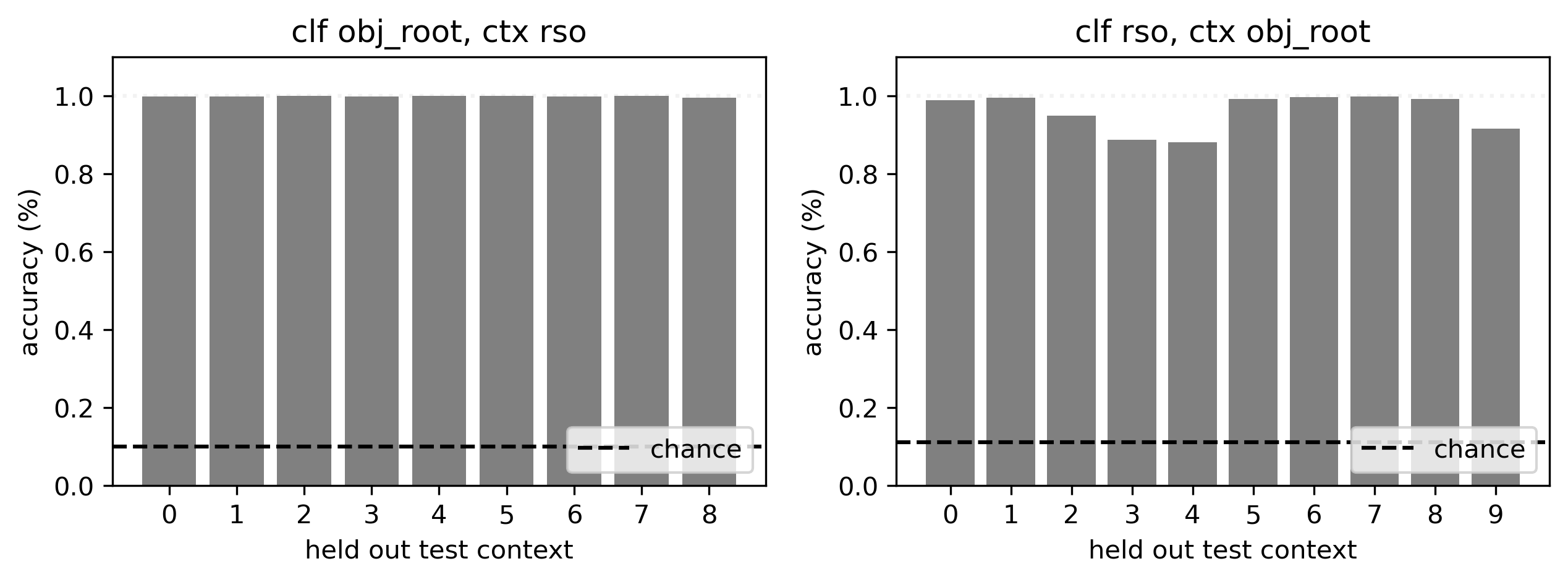}
	\caption{Representational factorization of abstractions encoding independent semantic features. Bar graph showing the generalization accuracy of linear probes trained to classify variable $A$ as contextual variable $B$ changes. \textbf{Left}: $A$ = object membership and $B$ = RSO. \textbf{Right}: $A$ = RSO and $B$ = object membership.}
	\label{fig:rso_generalization}
\end{figure}

To assess whether the network encodes the two abstractions independently, we performed a probe generalization experiment that measures how consistent the representations for object membership are when RSO changes, and vice versa. Working with a collection of embeddings $\{\mathbf{z}_i\}_i$, each expressing information about their specific object membership (semantic feature $A$, with classes $\{A_i\}_{i=1,k_A}$) and RSO (semantic feature $B$, with classes $\{B_i\}_{i=1,k_B}$), we trained a linear classifier for A ($\mathbf{z} \rightarrow \{A_i\}$) on a subset of tokens excluding those belonging to $B_i$ ($\{\mathbf{z} \mid B_\mathbf{z} \in \{B_j\}, j \neq i\}$) and then tested how well this probe generalizes to the held-out subset $\{\mathbf{z} \mid B_\mathbf{z} = B_i\}$. We repeated the process for each $B_i$ to get a picture of how consistent the $A_i$ representations were to a change in contextual variable $B$. Figure \ref{fig:rso_generalization} shows the results of this experiment when the classifier predicts object membership (left), and RSO (right), respectively. We found that probes for object membership generalized perfectly across RSOs. Looking at the reverse experiment, namely generalizing RSO classification across tokens with distinct object memberships, we found generalization accuracies to be only marginally weaker (> 80\%). While this analysis is far from being comprehensive, it shows that transformers can represent independent semantic features in a factorized manner. Supporting the idea that representational independence helps with generalization, we have notably reported in section \ref{sec:parameter_tuning} that our models successfully reconstruct masked boards that were not part of their training set. 

\subsubsection{Abstractions organize in a part-whole hierarchy}
\label{subsec:compositional_organization}

\begin{figure}[ht]
	\centering
	\includegraphics[width=0.7\textwidth]{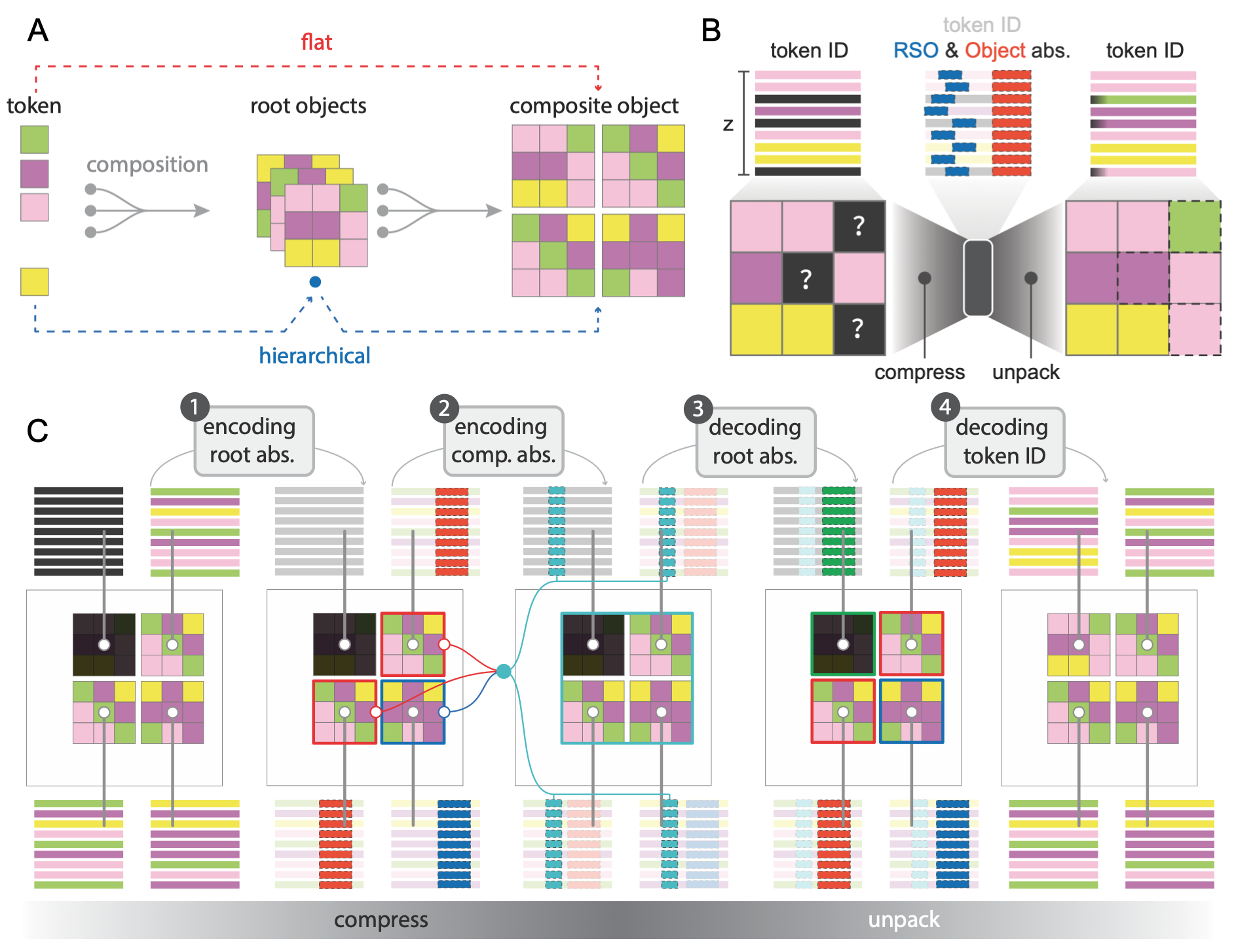}
	\caption{\textbf{A}: schematic of the two schemes for encoding abstractions in the network: hierarchical representations vs. flat representations. \textbf{B}: Schematic showing the hypothesized autoencoder-like computational strategy evolved by the network to reconstruct a root object with sporadic token masking. The figure shows the hypothetical evolution of the object token embeddings through the reconstruction process. Rectangles with dotted black contours represent specific abstractions within the embeddings. \textbf{C}: Same as B, in the case of a composite object with a fully masked constituent object.}
	\label{fig:comp_hypotheses_AE}
\end{figure}

HOD was explicitly constructed to test whether part-whole hierarchies in the dataset blueprint, where composite objects are constructed through the arrangement of smaller root objects (Figure \ref{fig:methods_dataset}), would transpire in the realm of abstractions. In particular, we sought to substantiate one of the two following hypotheses (Figure \ref{fig:comp_hypotheses_AE}a): (i) The \textit{hierarchical representation hypothesis}, whereby "level 2" abstractions for composite objects, if they exist, are constructed by integrating "level 1" root object abstractions; (ii) The \textit{flat representation hypothesis}, where the network fails to build on top of level 1 abstractions and constructs level 2 abstractions from token information directly. In general, the hierarchical representations scheme is regarded as desirable as it is more information efficient than the flat representation scheme\cite{lake2017building}.

\textbf{Conditions for the emergence of a composite object abstraction}: Having previously shown that abstractions for root objects arise from the masking of constituent tokens (section \ref{subsec:object_membership_abstraction}), we anticipated that composite object abstractions would also emerge despite the lack of even larger structures. However, we also reasoned that sporadic masking of the tokens might be insufficient for composite abstractions to develop as networks could in theory infer missing tokens from partially masked constituent objects. To test this hypothesis, we compared the internal representations of networks trained with sporadic masking versus networks trained with additional patch masking, designed to randomly mask entire constituent objects one at a time (Sup. Figure \ref{supfig:comp_patch_sporadic_masking}). We found that networks trained without patch masking failed to develop linearly separable composite object representations or leverage composite information to reconstruct fully masked constituents. In contrast, when patch masking was used, linear probes could more accurately predict composite ID. In particular, the signal was the clearest when probing the embeddings of tokens that were part of an entirely masked constituent object (Sup. Figure \ref{supfig:comp_patch_sporadic_masking}, bottom right panel). We conclude from these observations that when networks have learned to generate level 2 object abstractions, they only do so when local information from constituent objects is insufficient to infer missing token IDs. Accordingly, we conducted all subsequent analyses in that specific condition.

\textbf{Autoencoder interpretation of abstractions}: Based on results from previous sections, we draw Figure \ref{fig:comp_hypotheses_AE}b, which illustrates our current understanding of how abstractions play into the network's inference process: (Phase 1, compression) Starting from a partially masked root object, whose token embeddings primarily encode token ID, the network modifies these embeddings to encode RSO and object membership abstractions. (Phase 2, decompression) This abstract information is then unpacked to produce token ID for both masked and unmasked tokens. We see the two phases of this inference process as analogous to the encoding and decoding networks used by autoencoders architecture\cite{bank2023autoencoders}, with abstractions playing a similar role to latent representations found at the bottleneck of these architectures. Extending that understanding to composite abstractions, and assuming the hierarchical representation hypothesis, gives Figure \ref{fig:comp_hypotheses_AE}c: First, each unmasked object token on the board expresses a level 1 abstraction that represents the constituent object they are part of (see step 1). Level 1 abstractions are then integrated to produce a level 2 abstraction shared across both mask and unmasked tokens (see step 2). Finally, level 2 abstractions, along with composite RSO, are decompressed into level 1 abstractions, which are then combined with root RSO to infer token identity (see steps 3 and 4). In contrast to that picture, processing according to the flat representation hypothesis would resemble \ref{fig:comp_hypotheses_AE}b, with level 2 abstractions replacing level 1 and being constructed directly from token ID information.

\textbf{Delineating causal composite abstractions}: We first asked whether level 2 abstractions could be linearly read out from token embeddings, and if so, at which computational stages. We recorded the network’s activations as it attempted to reconstruct boards featuring a composite object with one of its four constituents masked. We divided tokens between masked and unmasked and trained linear probes to read out level 1 and level 2 abstractions (Sup. Figure \ref{supfig:comp_level2_delineation}, top). Our findings were as follows: (i) In the case of unmasked tokens, level 1 abstractions were perfectly read out from the first attention layer onward; (ii) In the case of masked tokens, the same information could only be read out at a later computational stage; (iii) Probes for level 2 abstractions poorly performed across all computational stages for unmasked tokens; (iv) However, level 2 abstractions appear to be linearly separable from the second attention layer onward in masked tokens. These four points were consistently observed across several network initializations and dataset instances (data not shown). Altogether, we found that unmasked object tokens first broadcast their level 1 abstractions, followed later on by masked object tokens changing their embeddings to express a level 2 abstraction. These preliminary findings fit the hierarchical representation hypothesis.

Next, we ran a series of manipulation experiments designed to find causal representations of level 2 abstractions. Working from a collection of boards of the kind shown in Figure \ref{fig:comp_hypotheses_AE}c (single composite object randomly positioned with one constituent root object fully masked), we tested one computational stage at a time, modifying the embeddings of masked tokens to replace activations coding for level 2 abstraction $A$ with prototype activations for a different composite $B$. We considered the manipulation successful when the network reconstructed the masked tokens as if it were perceiving $B$ rather than $A$, For example, if composite $A$ has constituent roots (a, b, c, d) and $B$ = (e,f,g,h), editing the level 2 abstraction $A$ → $B$ in the tokens of “a” would lead the network to reconstruct them as tokens of “e”. As done previously, we assessed the effect of our manipulations using the $\Delta$ method that compares our rationally designed interventions against random ones (Sup. Figure \ref{supfig:comp_level2_delineation}, bottom). Our results indicate that the representations flagged by the linear probes in the second attention layer of the network (\texttt{b1\_attn\_update}) were indeed causality encoding level 2 abstractions. Interestingly, while linear probes also read out level 2 abstractions in downstream stages, we find little to no effect manipulating those, suggesting that they were only transiently meaningful to the network.

\begin{figure}[ht]
	\centering
	\includegraphics[width=0.9\textwidth]{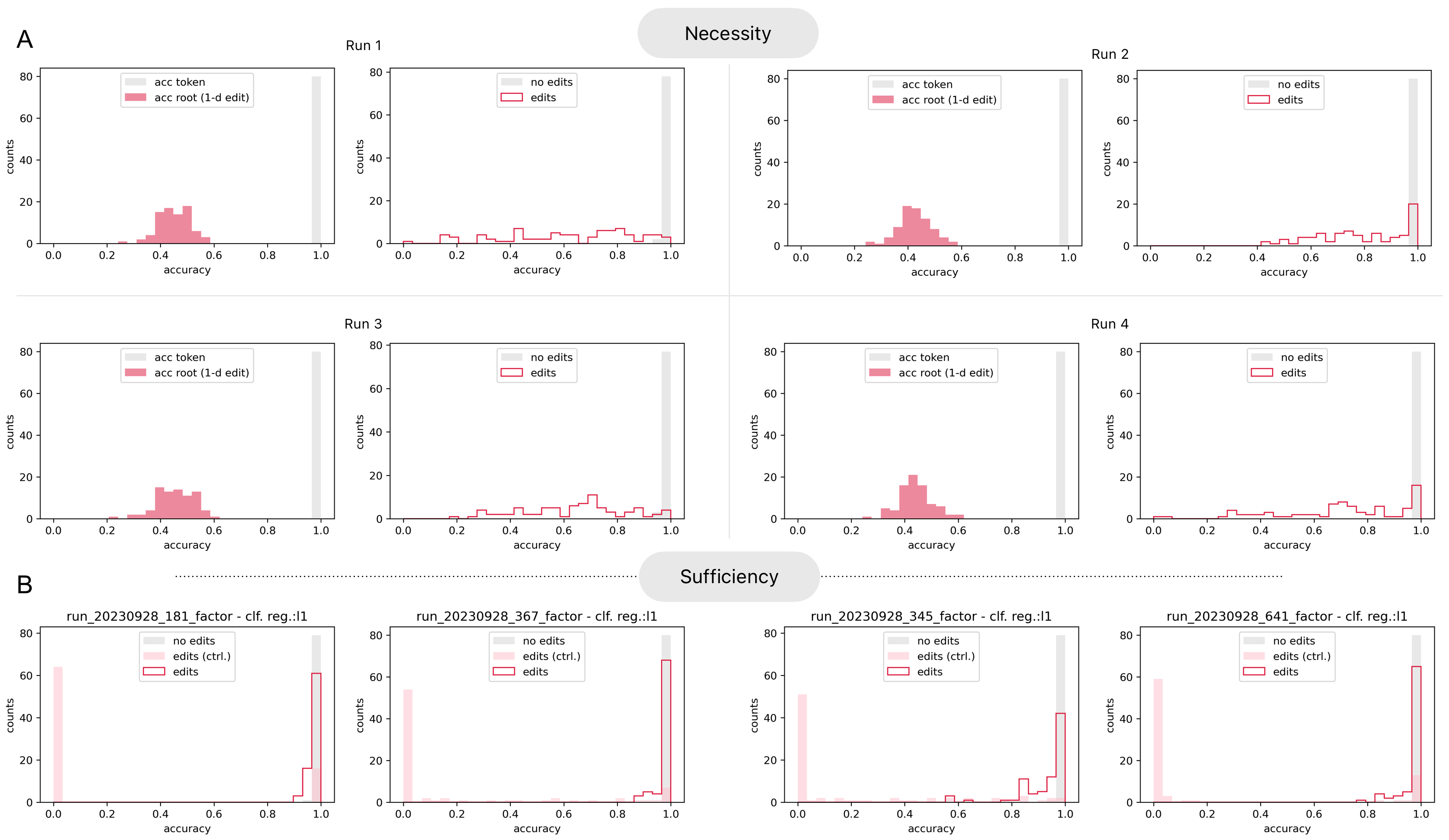}
	\caption{Level 1 object abstractions are sufficient \& necessary to generate level 2 object abstractions. (\textbf{A}): Necessity analysis performed on four different runs (four quadrants). For each run, \textbf{left} panel shows the distribution of prediction accuracies for linear probes trained on token ID (acc token, gray) and level 1 object abstraction (acc root, red) after performing one-dimensional edits designed to remove level 1 object information. \textbf{Right} panel shows the accuracies of linear probes trained to read out level 2 composite abstraction from embeddings at a downstream computational stage when edits have been performed (red) or not (gray). (\textbf{B}): Sufficiency analysis was performed on four different runs. We trained a linear probe to predict level 2 abstractions at stage $i$ (gray bars, no edits, show test accuracies). We then repeated the forward pass while replacing the embeddings of unmasked tokens at stage $i-1$ with a prototype embedding for the object they are part of, and collected the modified embeddings at stage $i$. Finally, we tested the generalization of the probe on these perturbed embeddings (red histogram, edits). Negative control was obtained by using the wrong prototypes (edits ctrl.).}
	\label{fig:comp_encoding}
\end{figure}

\textbf{Level 1 abstractions are necessary and sufficient to generate level 2 abstractions}: We asked whether the causal composite abstraction (level 2) observed in the second attention layer of the network (stage $i$) were constructed from root abstractions (level 1) that emerged upstream in the computation (stage $i-1$), as opposed to level 0 token information. We performed a manipulation experiment at stage $i-1$ designed to meticulously alter level 1 abstractions from unmasked object token while preserving their token ID information. In particular, we used the unit-agnostic editing method introduced in the section \ref{subsec:unit_agnostic_approach} to swap the level 1 abstractions of unmasked constituent objects to a different one. We confirmed that our 1d edits successfully altered the abstractions while leaving token IDs intact by evaluating the post-editing accuracies of linear probes trained to read out level 1 abstraction and token ID information on the unedited embeddings (see Figure \ref{fig:comp_encoding}a, left panel). We then looked at the effect that removing level 1 abstractions from stage $i-1$ embeddings had on the emergence of the level 2 abstraction in the second attention layer of the network (stage $i$). We trained a linear probe to read out composite abstraction for token embedding collected in the absence of edits and used it to read out the same information from embedding produced after upstream edits had been carried out(see Figure \ref{fig:comp_encoding}a, right panel). We found that following the edits, the network failed to produce level 2 abstractions as suggested by low probe accuracies. These results were replicated for several runs and consistently showed that level 1 abstractions were necessary for the network to generate level 2 abstractions (Figure \ref{fig:comp_encoding}a).

To test for sufficiency, i.e., level 1 abstractions alone are responsible for the emergence of level 2 abstractions, we sought to remove token ID information from all unmasked object tokens at stage $i-1$. We accomplished this by replacing each unmasked token embedding with the average embedding for the parent object (object prototype), thus averaging out token ID and positional information. We then looked for the presence of the level 2 abstraction at stage $i$ by using a linear probe (Figure \ref{fig:comp_encoding}b). Consistent across several runs, we found that the edited tokens are sufficient to produce the level 2 abstraction suggesting that the network indeed uses the level 1 abstractions to make the derivation. As a negative control, we also show that performing the edits with the wrong object prototype prevents the network from forming the correct level 2 abstraction. Altogether, these results support the encoding phase of the hierarchical representation hypothesis depicted in Figure \ref{fig:comp_hypotheses_AE}c.

\textbf{Level 1 abstractions for masked constituent are unpacked from level 2 abstractions}: In the decoding phase of the hierarchical representation hypothesis, a level 2 abstraction is unpacked to produce level 1 abstractions, which in turn inform token ID predictions (Figure \ref{fig:comp_hypotheses_AE}c). To test this idea, we sought to measure how essential the level 2 abstraction found in masked tokens is to the downstream construction of level 1 abstractions. The final step that bridges level 1 abstraction to token ID has been covered in length in section \ref{sec:abstraction_manipulation}.

\begin{figure}[ht]
	\centering
	\includegraphics[width=0.8\textwidth]{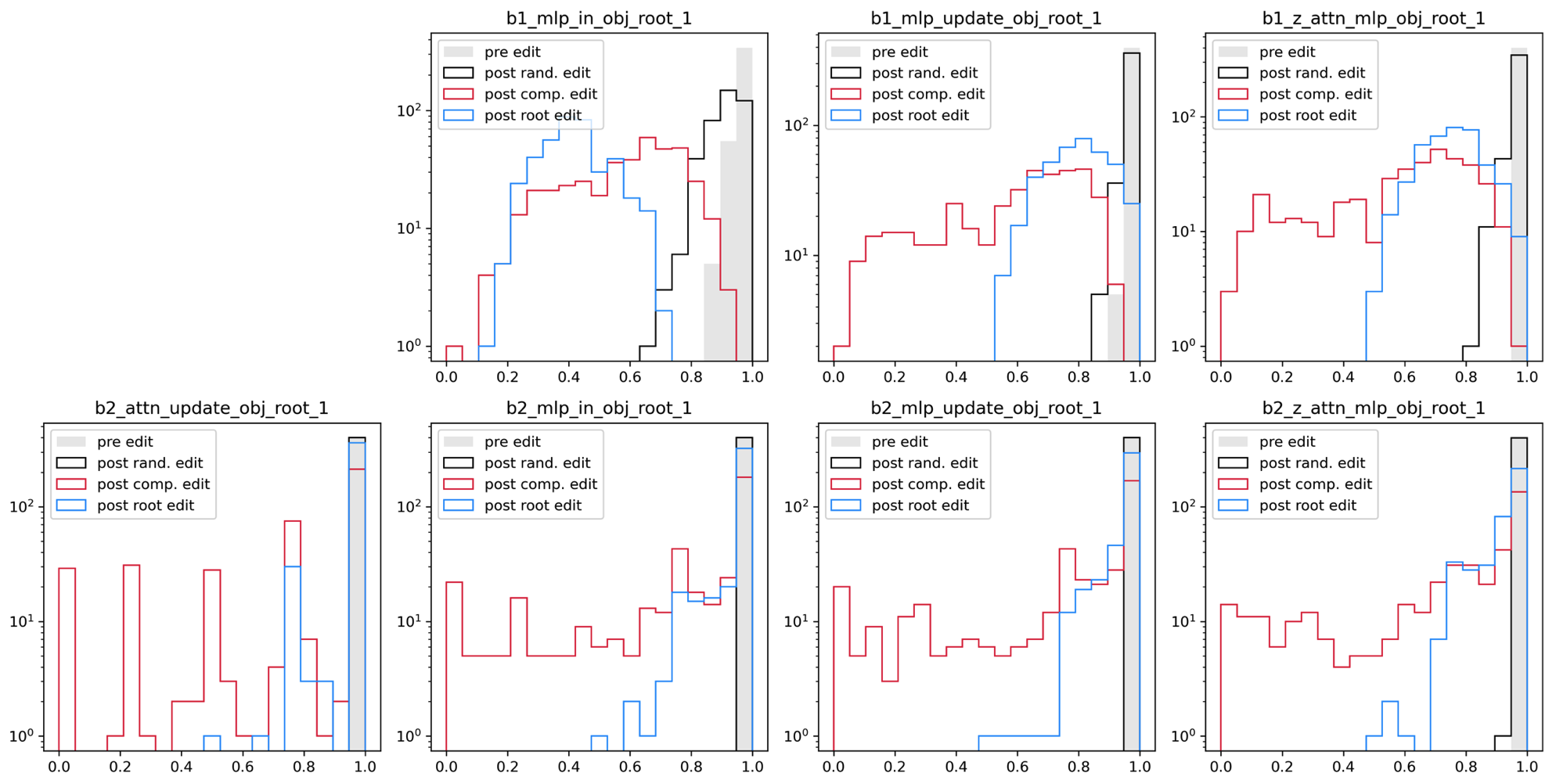}
	\caption{Effect of 1d edits against level 1 and level 2 object abstractions on downstream level 1 object abstractions. We edited the embeddings of patched masked tokens in \textnormal{\texttt{b1\_attn\_update}} using 1d unit-agnostic methods targeted at level 1 (blue) or level 2 (red) object abstractions. As a negative control, we performed an edit along a randomly chosen dimension of the representational space (black). We then used linear probes trained to classify level 1 object membership to measure the effect of our edits on the level 1 abstraction of patched masked tokens in downstream computational stages (one plot per stage). Accuracies are reported as histograms. The gray bars indicate the accuracy of the probe on unaltered embeddings (no edits).}
	\label{fig:comp_decoding}
\end{figure}

Following the same strategy used in the necessity study presented above, we performed 1d edits designed to remove either level 1 or level 2 abstraction from the embeddings of masked tokens at the second attention layer of the network. We then assessed how these modifications affected the level 1 representations in downstream computational steps (Figure \ref{fig:comp_decoding}). In general, we found that it is easier to edit level 1 abstractions while leaving level 2 abstractions intact than the opposite (Sup. Figure \ref{supfig:comp_decoding_edits}). Looking at the downstream effect of our edits, we found that, in 2 out of the 4 runs tested, editing level 2 abstraction had more effect on downstream level 1 abstraction than directly editing the same abstraction at the second attention layer (Figure \ref{fig:comp_decoding}). In the other half of the runs, we found the effects to be comparable (data not shown). In all cases, our edits drastically reduced the network’s ability to produce the same level 1 abstractions observed in the unperturbed computations. In contrast, performing 1d edits along random vectors causes little to no impairment (ctrl. black curves). Altogether, these results give credence to the decoding part of the hierarchical representation hypothesis. However, we were not able to conduct a  sufficiency analysis similar to the one used for the encoding phase since unpacking abstractions requires additional token-specific information such as RSO (section \ref{subsec:representational_independence}). Pulling together the results for the encoding and decoding phases, these findings make a compelling case in favor of the hierarchical representation scheme where low-level abstractions are composed together to form higher-level ones, hence mirroring the part-whole hierarchy of the dataset.

%/////////////////////////////////////////////////////////////////////////////////////////
% RESULTS 5 - Extracting abstraction through language-like information bottleneck
%/////////////////////////////////////////////////////////////////////////////////////////

\subsection{Extracting abstractions via a language-like information bottleneck}
\label{sec:language_bottleneck}

So far, we have seen that despite taking in and outputting raw token information, self-supervised transformers develop intermediate abstract representations that capture key semantic features of their inputs. In addition to being an interesting phenomenon for paralleling an important aspect of human cognition, we believe that gaining access to these abstractions could be extremely valuable to understand the inner workings of black-box deep learning systems and steer the network's decision-making to better align with user goals. While we have demonstrated how a combination of linear probes and gain-of-function experiments can be used to delineate the learned abstractions, the process remains challenging and inexact. In an attempt to simplify this process, we extended the vanilla transformer architecture studied above (IN, inference network) with an auxiliary language network (ALN) designed to encourage the overall system to “talk” about its computations (LEA architecture, Figure \ref{fig:methods_architecture_language} and section \ref{subsec:architectures}). 

LEA processes information as follows: After a forward pass through IN that reconstructs $x$ from a masked board input $\tilde{x}$, the latent representations $\mathbf{Z}^{k_b}$ produced by IN are fed to ALN\footnote{Here ${k_b}$ refers to the number of transformer blocks in IN.}. ALN compresses $\mathbf{Z}^{k_b}$ into a discrete language-like representation $\mathbf{s}$ (sentence) from which IN is then tasked to reconstruct the ground truth board $x$. Similar to an English sentence, $\mathbf{s}$ is a $l$-dimensional vector of integers that each correspond to a specific word in a vocabulary of size $V$: ALN generates as output a $l \times m$ matrix of embeddings, which is then fed to a vector quantizer that matches each of the $l$ vectors to a learned codebook to get the individual words out. In this framework, IN and ALN are trained concomitantly and the system evolves its own language by learning to efficiently talk about its inputs. We designed this system with the hope that the language created by LEA would offer a clearer indexing of the abstractions as well as a simpler way to intervene on them.

\subsubsection{LEA evolves to talk about the abstractions it has learned}
\label{subsec:lea_language}

\begin{figure}[ht]
	\centering
	\includegraphics[width=0.8\textwidth]{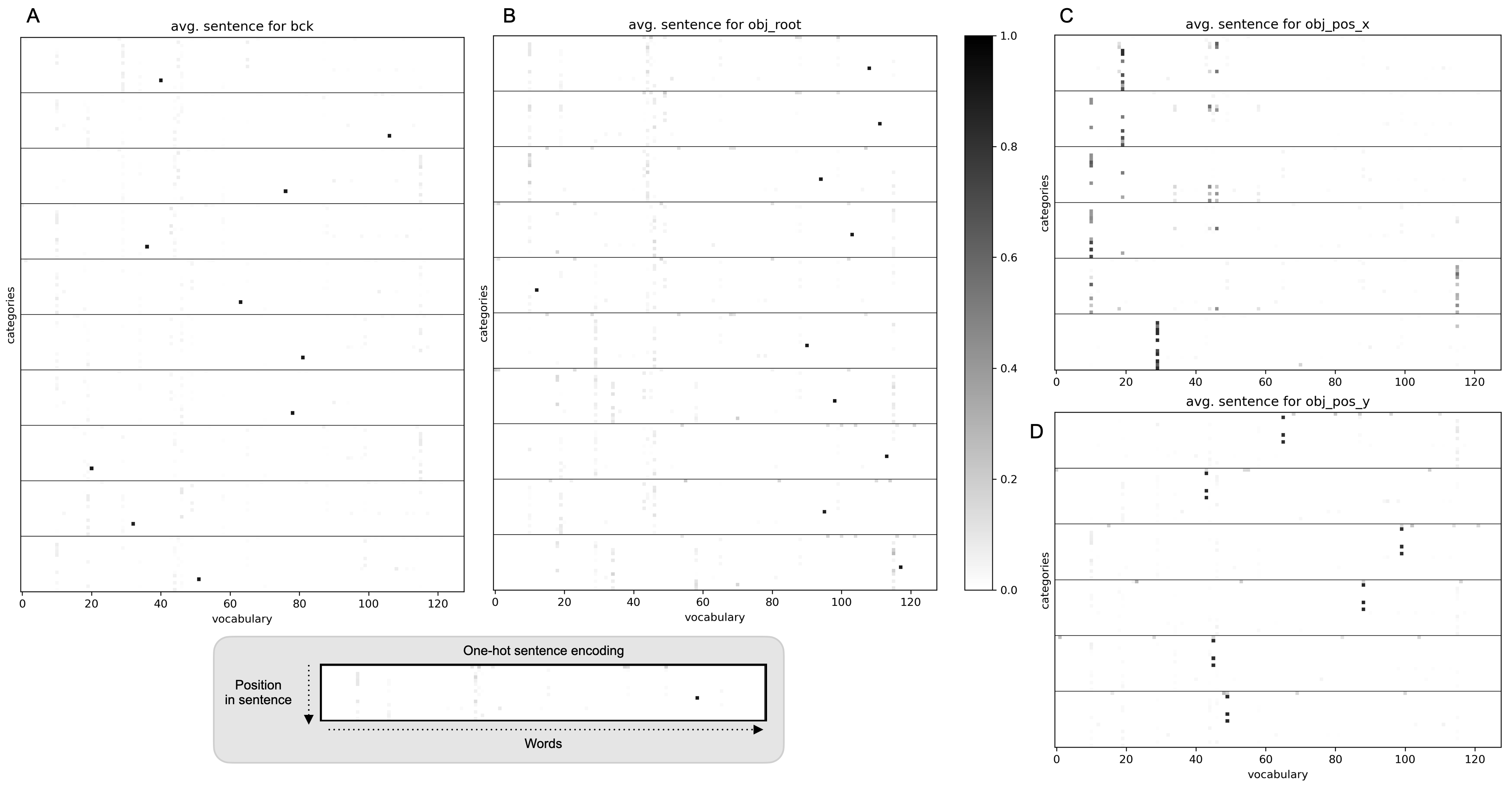}
	\caption{Linguistic patterns associated with specific semantic features of the inputs. Each plot shows the corrected average sentence $\mathbf{S}_{\bar{n}}-\mathbf{S}_{\bar{N}}$ for a specific feature. We looked at correlates for specific backgrounds (\textbf{A}, sentences for 10 different backgrounds are vertically stacked), specific objects (\textbf{A}, 10 different objects), position of the object in the board along the x (\textbf{C}, 6 positions) and y (\textbf{D}, 6 positions) coordinates.}
	\label{fig:language_rosetta}
\end{figure}

We first set out to test whether LEA would produce a human-interpretable language where individual words can be matched with specific semantic features (e.g. objects in the board). We trained LEA on a collection of masked boards featuring one or two randomly positioned root objects (section \ref{subsec:dataset}). Once the system could successfully reconstruct the masked boards from its training set (both from $\tilde{x}$ and $\mathbf{s}$), we confirmed that it generalized over a held-out set of boards (data not shown). LEA's ability to reconstruct novel boards, particularly from $\mathbf{s}$, was encouraging and suggested that the system might have evolved a compositional language (see section \ref{subsec:lea_compositionality}). To analyze LEA’s language, we generated a collection of $N$ random boards $\{x_i\}_i$ each featuring a single object, passed them through the network, and collected the corresponding sentences $\{\mathbf{s}_i\}_i$. To facilitate downstream analysis, we converted each word into its one-hot representation and stored the information in a 3-dimensional tensor $\mathbf{S}$ of shape $N \times l \times V$, where $N$, $l$, $V$ correspond to the number of boards, sentence length, and vocabulary size, respectively. Taking the average of $\mathbf{S}$ over the first dimension gives us the overall frequency $\mathbf{S}_{\bar{N}, i, j}$ of observing a specific word $j$ at a specific place $i$ in the sentence. By comparing this baseline $\mathbf{S}_{\bar{N}}$ with $\mathbf{S}_{\bar{n}}$ obtained by only considering a subset of $n$ boards with a specific semantic feature, we revealed linguistic patterns associated with specific features (Figure \ref{fig:language_rosetta}). Notably, we attempted to find linguistic patterns in $\mathbf{S}$ indicating the presence of a specific background, a specific object, as well as the specific location of that object on the board. We found that, in the majority of the cases, the sentences describing boards sharing a specific feature, all used the same word. In fact, for backgrounds and objects, these sentences not only used the same word but also positioned it similarly within the sentence. Altogether, it appears that LEA has evolved a vocabulary geared towards talking about the abstractions studied in the previous sections. 

\subsubsection{Compositional nature of LEA’s language}
\label{subsec:lea_compositionality}

Human languages are compositionally structured, meaning complex expressions can be constructed from simpler parts, with the meaning of the whole determined by the meanings of the parts and the rules used to combine them\cite{mitchell2010composition}. This allows for an almost infinite variety of expressions from a finite set of elements and rules, enabling humans to generate and understand novel sentences they have never encountered before. One key feature of compositionality that we explored in \ref{subsec:representational_independence} is contextual independence, where words typically have meanings that are independent of specific contexts, allowing them to be flexibly recombined in countless new situations. Given that the boards in our dataset are created through a compositional process that involves putting together a background and one or several objects, we wondered whether LEA’s language would also reflect this compositional nature. In particular, we reasoned that it would be advantageous for the system to evolve a language that implements contextual independence. For example, the way LEA talks about an object and its position should not change based on the type of background it is on. 

\begin{figure}[ht]
	\centering
	\includegraphics[width=0.7\textwidth]{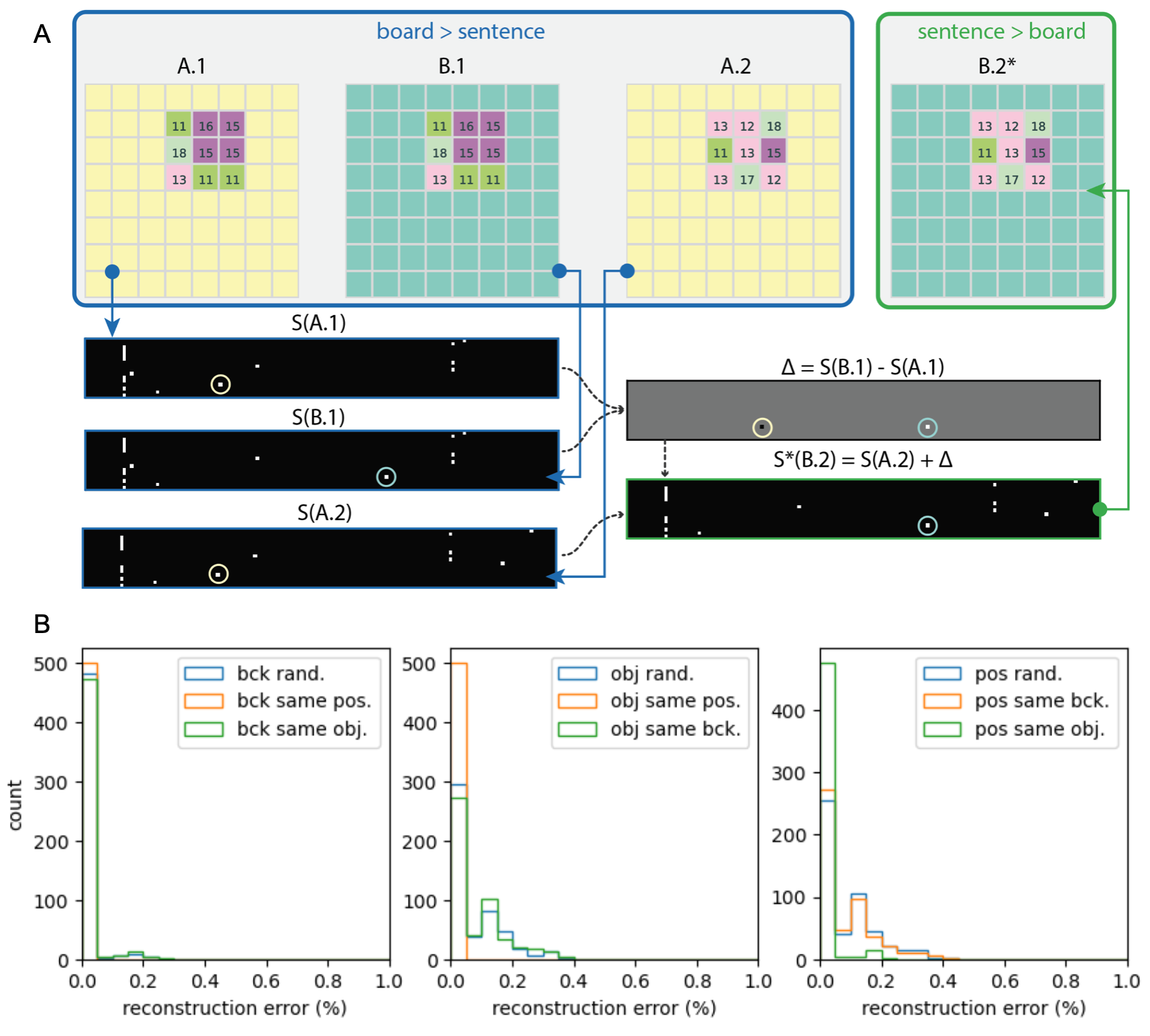}
	\caption{(\textbf{A}):Example of contextual independence, swapping backgrounds. Boards A.1 and B.1, featuring background $bck_1$ and $bck_2$, are processed by IN+ALN (blue arrows) to generate the corresponding sentences $\mathbf{s}$(A.1) and $\mathbf{s}$(B.1). Sentences are shown as $l \times V$ one-hot encodings. The $\Delta$ between the two sentences is calculated and represents the changes that need to be applied to a sentence to switch the background of the board it describes from $bck_1$ to $bck_2$. We then obtain the sentence $\mathbf{s}$(A.2) for board A.2 that matches the background of A.1 but features a different object. We then produce a new synthetic sentence $\mathbf{s}^*$(B.2) by adding $\Delta$ to $\mathbf{s}$(A.2). Finally, we test whether the new sentence is meaningful by asking IN to interpret it. \textbf{B}: Aggregated results for 500 random edits. We measured how meaningful synthetic sentences were to the network by looking at the reconstruction error. Edits were split into three groups depending on the feature that was being edited in the sentence: Edits for background, object, and object position are shown in the left, center, and right histograms, respectively. For each experiment, $\Delta$ is evaluated in one context and applied to a different context. Within each group, we further make a distinction between edits depending on how different these two contexts were. For example, when editing the background (left histogram), we could require the two contexts to share the same object (green), the same object position (orange), or no requirements (blue).}
	\label{fig:comp_context_independence}
\end{figure}

Figure \ref{fig:language_rosetta}, already provided us with preliminary evidence supporting the idea that LEA’s language implements contextual independence: $\mathbf{S}_{\bar{n},i,j}$ values close to 1 indicate that a given word is always present when a specific feature appears in the board, that is, regardless of other elements of the board. However, in the case of human language, contextual independence goes a step further: It says that replacing a part that is independent of its context would yield another meaningful sentence. For example, swapping "cat" for "dog" in the sentence "the cat is eating food" produces a sentence that makes complete sense. Therefore, we sought to check that rational edits of the sentences produced by LEA would also produce sentences that are meaningful to the network. Let us imagine that LEA generates the sentence $\mathbf{s}_{b_i,o_j,p_k}$ to describe a board featuring object $j$ at position $k$ on background $i$. In the case of contextual independence, the changes to the sentence that are required to talk about a different background ($\mathbf{s}_{b_1,o_3,p_2}$ → $\mathbf{s}_{b_2,o_3,p_2}$) should be the same regardless of the position and type of object: $\Delta_{b_{12}} = \mathbf{s}_{\mathbf{b_2},o_3,p_2} - \mathbf{s}_{\mathbf{b_1},o_3,p_2} = \mathbf{s}_{\mathbf{b_2},o_1,p_4} -\mathbf{s}_{\mathbf{b_1},o_1,p_4}$. As a corollary, we should be able to take $\Delta_{b_{12}}$ evaluated in a given context and apply it in a different context to produce the same result. Figure \ref{fig:comp_context_independence}a and Sup. Figure \ref{supfig:comp_context_independence_obj} provide two successful examples of that type of manipulation: In the first case we changed LEA’s original sentence $s_{\mathbf{b_1},o_3,p_2}$ to $s_{\mathbf{b_2},o_3,p_2}$ using $\Delta_{b_{12}}$ evaluated in a different context and showed that the resulting sentence is meaningful to the network (i.e., IN produces the right board from the synthetic $\mathbf{s}^*$). In the second example, we successfully edited sentences to change the object appearing on the board. Attempting the same experiment over 500 random cases (swapping background, object, or object positions), we found that the $\Delta$ evaluated in one context can in general be used to generate meaningful synthetic sentences in different contexts (low reconstruction error from synthetic $s^*$, Figure \ref{fig:comp_context_independence}b). These results provide convincing evidence supporting the fact that LEA's language implements contextual independence.

\subsubsection{Steering LEA's output using its language}
\label{subsec:lea_hallucination}

\begin{figure}[ht]
	\centering
	\includegraphics[width=0.7\textwidth]{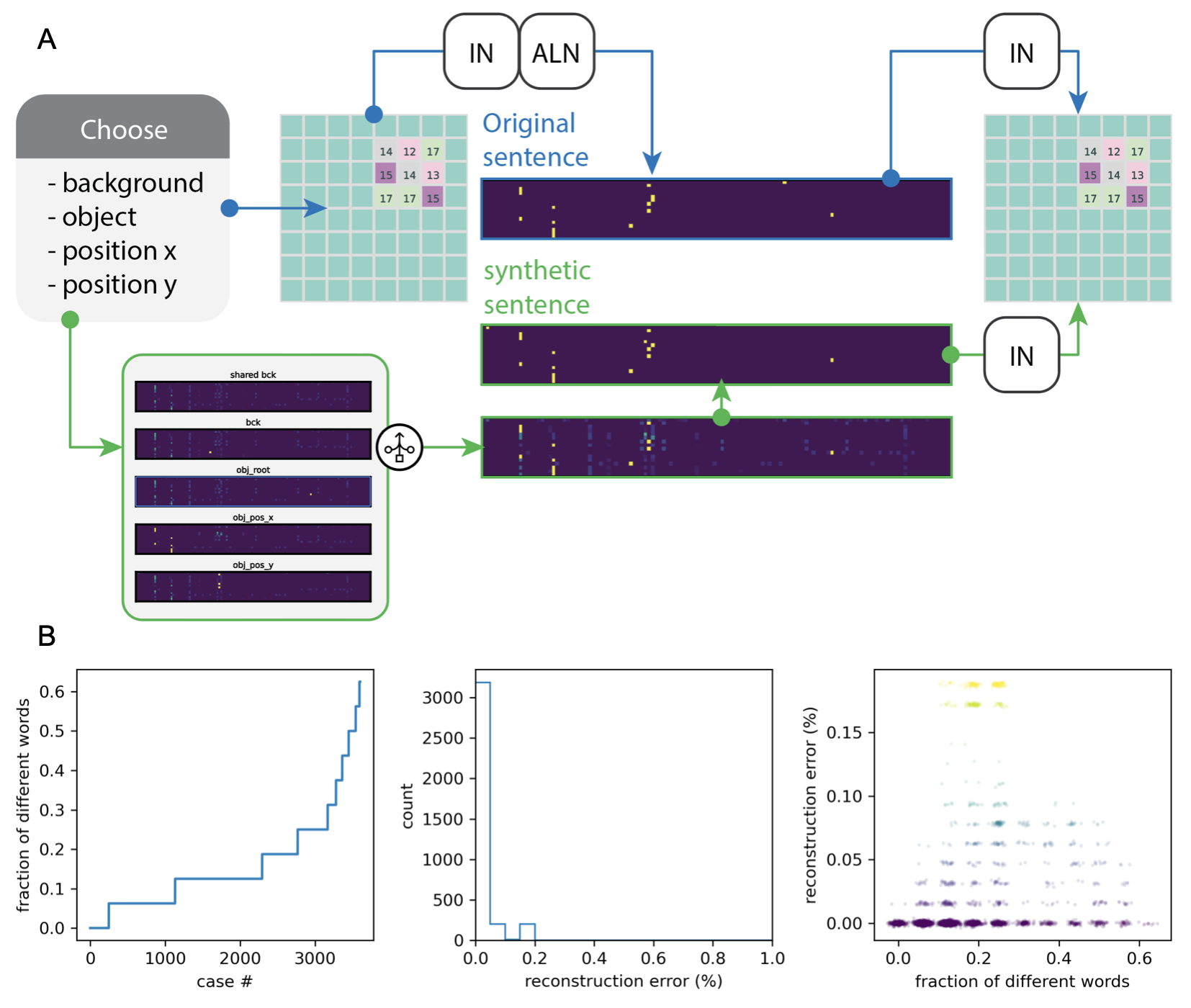}
	\caption{(\textbf{A}): Talking LEA into visualizing specific boards. The user defines a target board by specifying a background, object, and object position. Original processing pipeline (blue): The chosen board is passed through IN and ALN to produce a sentence $\mathbf{s}$, which is then used by IN to reconstruct the board. Alternatively (green), the board can be reconstructed from a synthetic sentence $\mathbf{s}^*$ created by aggregating average sentences $\mathbf{S}_{\bar{n}}$ for each feature. (\textbf{B}): Rationally designed synthetic sentences successfully interpreted by IN. We assessed the reconstruction error for ~4000 different boards. \textbf{Left}: Fraction of words that differed between the ground truth $\mathbf{s}$ and synthetic $\mathbf{s}^*$ sentences for each case (sorted for clarity). \textbf{Center}: Distribution of reconstruction error from $\mathbf{s}^*$ across cases. \textbf{Right}: Reconstruction error as a function of the difference between $\mathbf{s}$ and $\mathbf{s}^*$.}
	\label{fig:synthetic_sentences}
\end{figure}

In Figure \ref{fig:language_rosetta}, we showed a correlation between the use of specific linguistic patterns in LEA’s language and the presence of specific semantic features in the input they describe. This raised the interesting prospect of using this equivalence table to directly "talk" to the network, essentially steering its internal computations. However, as discussed in section \ref{sec:abstraction_manipulation}, correlation does not imply causality. To demonstrate that the linguistic patterns identified in Figure \ref{fig:language_rosetta} carry the same meaning to the network, we tasked IN with reconstructing entire boards from synthetic sentences created by composing these patterns (Figure \ref{fig:synthetic_sentences}a): We began with choosing a target board for LEA to visualize by selecting a specific background, object, and board position for that object. We then create a corresponding synthetic sentence $\mathbf{s}^*$ by aggregating the average sentence $\mathbf{S}_{\bar{n}}$ for each feature. Finally, we passed $\mathbf{s}^*$ to IN for reconstruction and evaluated the success of this process by comparing the network-generated board with our pre-defined target board. Additionally, we also compared our synthetic sentence $\mathbf{s}^*$ with the actual sentence $\mathbf{s}$ that would have been generated by IN+ALN for the same board.

When using this approach to generate over 4000 different boards, we found that the majority of synthetic sentences were successfully interpreted by the network, which produced the corresponding target boards (Figure \ref{fig:synthetic_sentences}b, center panel). That is despite an average difference between the synthetic and true sentences of about 15\% (left panel). We found no clear correlation between this difference and the reconstruction error (right panel). These results demonstrate that the words singled out in Figure \ref{fig:language_rosetta}, are meaningful to the network and their meaning matches our interpretation.

Overall, our results show that LEA evolves a compositional language geared towards talking about the abstract features of its inputs. In addition, we show that the one-to-one relationship that exists between words in LEA’s language and abstractions makes it easy for a user to steer its computations.

%/////////////////////////////////////////////////////////////////////////////////////////
% RELATED WORK
%/////////////////////////////////////////////////////////////////////////////////////////

\section{Related Work}
\label{sec:related_work}

This study joins a rapidly increasing body of work concerned with reverse engineering the inner workings of deep learning systems, a field commonly referred to as \textit{mechanistic interpretability} (see \cite{rauker2023toward} for review). Directly related to the abstractions concept discussed in our paper, Albawi et al. 2017 \cite{albawi2017understanding} revealed the emergence of a hierarchy of increasingly abstract feature detectors in convolutional neural networks.  This pioneering work was later greatly extended by Chris Olah and colleagues at OpenAI\cite{cammarata2020thread:}, who created a comprehensive catalog of the features learned by large vision models\cite{olah2020zoom}. Notably, their work reveals the existence of neurons specifically tuned to a wide range of concepts; from low-level texture neurons to multimodal units that respond to highly abstract concepts such as gender and religion \cite{goh2021multimodal}. 

While a lot of unknowns remain, the field has also made significant contributions toward understanding transformer-based models. Amongst others, these include establishing a solid mathematical framework to talk about the transformer’s computations \cite{elhage2021mathematical}, the discovery of how induction heads and function vectors support in-context learning \cite{olsson2022context, todd2023function, hendel2023context}, as well as the ability to locate and edits facts in large language models (LLM) \cite{meng2022locating, meng2022mass}. Of particular interest to our work, several recent studies provided evidence showing that transformers can form abstract world models, including representation of concepts such as space and time \cite{li2022emergent, gurnee2023language}. Notably, Li et al. 2022 showed that an LLM solely trained over sequences of Othello moves developed a representation of the board that it updated on a turn-by-turn basis \cite{li2022emergent, hazineh2023linear, nanda2023emergent}. 

We dedicated section \ref{sec:compositionality} to the study of compositionality, with a focus on contextual independence and the hierarchical organization of abstractions. These two properties have received a great deal of attention in the subfield of \textit{representation learning} as they could drastically improve the interpretability and the generalization ability of deep learning systems \cite{higgins2016beta, lake2015human, lake2017building}. Focusing on transformers, studies investigating LLM’s ability to handle compositional language \cite{murty2022characterizing} and tasks \cite{dziri2023faith}, have reported that these systems often fail to capture the compositional nature of their inputs. Notably, and in contrast with our findings, Murty et al. 2022 found little evidence for compositionality in transformers trained with self-supervised learning \cite{murty2022characterizing}. 

Another important aspect of our work has to do with the methods we used to delineate abstractions and demonstrate their causal role in the network’s computations (see sections \ref{sec:abstraction_existence} and \ref{sec:abstraction_manipulation}). Probing classifiers \cite{belinkov2022probing}, also known as decoders, have been extensively used on both biological and artificial neural networks as a means to identify representations whose expression correlates with variables of interest (e.g., \cite{nanda2023emergent, tigges2023linear} uses linear probes to explain a transformer network, \cite{tang2023semantic, benchetrit2023brain} use more complex decoders to interpret brain signals). While probes let us speculate on the meaning of certain representations, interventions are required to confirm that they hold the same meaning for the neural network studied (i.e., causal \cite{geiger2021causal}). Gain-of-function experiments, which involve rationally manipulating neural representations to predictably steer a system’s output, have been successfully used to establish a clear connection between neural representations and their meanings or functions \cite{meng2022locating, liu2012optogenetic}. Notably, our manipulations were inspired by the interchange interventions of Geiger et al. 2022\cite{geiger2022inducing} and our edits were based on vector arithmetic as used by Nanda et al. 2023 \cite{nanda2023emergent}. Finally, our unit-agnostic edits borrow ideas from Iterative Nullspace Projection \cite{ravfogel2020null, elazar2021amnesic}.

%/////////////////////////////////////////////////////////////////////////////////////////
% DISCUSSION
%/////////////////////////////////////////////////////////////////////////////////////////

\section{Discussion}
\label{sec:discussion}

Driven by the need to anticipate the results of our actions, our brains have evolved the ability to form an abstract world model of our environment by gathering and processing our experiences. This model contains a set of abstractions, each of which encodes a different aspect of the hidden blueprint of our reality. In this study, we ask whether Transformer models, trained using a supervised learning objective that mimics the learning pressure faced by biological brains, are also capable of developing abstractions that capture the latent blueprint used to construct their inputs. To answer this question, we analyzed the workings of a small transformer tasked with reconstructing partially masked boards of "visual" tokens.

Our findings revealed that, despite taking in and outputting information at the token level, the transformer develops intermediate representations coding for higher-level concepts (section \ref{sec:abstraction_existence}). We showed that the embeddings of perceptually distinct tokens that share a semantic feature converge in representation towards a shared low-dimensional manifold or abstraction\cite{laurent2023feature}. In particular, we found abstractions of the sort for each element of the dataset's blueprint, suggesting that the model has created an \textit{in silico} abstract world model. Using gain-of-function manipulation experiments, where we predictably altered the outputs of the system by swapping one abstraction for another, we demonstrated that these abstractions play a critical role in the network's computations (section \ref{sec:abstraction_manipulation}). Our results suggest that the network progressively constructs higher-level representations from token-level information, which together form a compressed summary of the input. It then uses that information to infer the identity of any masked tokens. We also provide evidence substantiating the claim that transformers can organize their abstractions in a way that captures the compositional nature of their inputs (section \ref{sec:compositionality}). We showed that abstractions for semantic features that are independent tended to be factorized at the representational level. Additionally, we found that abstractions representing composite objects, constructed as the arrangement of smaller parts, were generated from the abstractions coding for the constituent objects rather than token information directly. Finally, we propose a new language-enhanced architecture (LEA) designed to more readily access the abstractions learned by the system (section \ref{sec:language_bottleneck}). The modified system features a language bottleneck that forces it to develop concise language-like representations of its inputs. We demonstrated that post-training, LEA develops a language geared towards talking about the abstractions we studied. In particular, we were able to obtain a one-to-one match between linguistic patterns used by LEA and elements of the dataset blueprint. We end the section by showing how this mapping can be leveraged to gain control over the network's decision-making process. 

Recent studies have drawn interesting parallels between activations in deep learning systems and neural activity in the mammalian brain\cite{bashivan2019neural,goldstein2022shared}. Likewise, we believe that further studying the abstraction process in artificial neural networks could help us better conceptualize the mechanisms underlying the formation, storage, and use of generalizable knowledge in the brain\cite{yang2019task,goudar2023schema}. However, the biological brain is not the only one that needs explaining. The recent push towards scaling up deep learning systems is yielding networks that are increasingly more powerful but also increasingly more opaque to human interpreters. In this context, we think that the systematic identification of abstractions could facilitate our understanding of the inner workings of larger transformer-based models (e.g. LLM). In particular, by characterizing how abstractions condition a network’s inference process, one might be able to describe its computations as a decision tree whose nodes correspond to these abstractions\cite{geiger2022inducing}. Furthermore, we also believe that the use of architectures like LEA, designed to force the network to describe its internal computations with discrete representations, might be beneficial for the creation of hybrid neuro-symbolic architectures that could enhance the reasoning capabilities of current deep learning systems\cite{wang2022towards}. Finally, we also speculate that systems like LEA, which essentially distill complex input data to their underlying blueprint, could help us make sense of highly dimensional datasets that have remained opaque to human interpreters.

%/////////////////////////////////////////////////////////////////////////////////////////
% Acknowledgments
%/////////////////////////////////////////////////////////////////////////////////////////

\newpage
\section{Author Contribution \& Acknowledgements}
QF. conceived, designed, and conducted all experiments in this study. JC. and TK. provided guidance and feedback throughout the project. QF. wrote the manuscript with input from all authors. The authors thank Susumu Tonegawa, Afif Aqrabawi, Timothy O'Connor, and David Bau for helpful discussions on the project. This work was supported by the Howard Hughes Medical Institute and the JPB Foundation.

%/////////////////////////////////////////////////////////////////////////////////////////
% REFERENCES
%/////////////////////////////////////////////////////////////////////////////////////////

\bibliographystyle{plain}
\bibliography{references}

%/////////////////////////////////////////////////////////////////////////////////////////
% SUPPLEMENTARY MATERIAL
%/////////////////////////////////////////////////////////////////////////////////////////

\captionsetup[figure]{name=Sup. Figure}
\setcounter{figure}{0}

\newpage
\section{Supplementary Material}

\renewcommand{\arraystretch}{1.}
\begin{table}[H]
	\centering
	\begin{tabular}{|c|c|c|}
		\hline
		Symbol & Value & Description \\
		\hline
		\hline
		$n$ & 8,10 & board width/height \\
		$n_b$ & 10 & number of background tokens \\
		$n_o$ & 10 & number of object tokens \\
		$N_{root}$ & 10 & number of root objects \\
		$m_{root}$ & 9 & number of tokens per root object \\
		$\textbf{g}_{root}$ & (3,3) & root object grid \\
		$N_{comp}$ & 5 & number of root objects \\
		$m_{comp}$ & 4 & number of root per composite object \\
		$\textbf{g}_{comp}$ & (2,2) & composite object grid \\
		\hline
	\end{tabular}
	\caption{Parameters used for HOD instances.}
	\label{tab:parameters_HOD}
\end{table}

\begin{table}[ht]
	\centering
	\begin{tabular}{|c|c|c|}
		\hline
		Symbol & Value & Description \\
		\hline
		\hline
		$k_b$ & 3 & Number of transformer blocks \\
		$k_h$ & 2 & Number of attention heads \\
		$d_e$ & 64 & Embedding dimension \\
		$d_p$ & 32 & Positional encoding dimension \\
		$d_{mlp}$ & 128 & MLP hidden dimension \\
		$V_t$ & 21 & Token vocabulary cardinality \\
		\hline
	\end{tabular}
	\caption{Network parameters for the vanilla transformer architecture.}
	\label{tab:parameters_vanilla}
\end{table}

\begin{table}[ht]
	\centering
	\begin{tabular}{|c|c|c|}
		\hline
		Symbol & Value & Description \\
		\hline
		\hline
		$k_b$ & 3 & Number of transformer blocks \\
		$k_h$ & 2 & Number of attention heads \\
		$d_e$ & 32 & Embedding dimension (token \& words) \\
		$d_p$ & 16 & Positional encoding dimension ($\mathbf{P}$, $\mathbf{P_l}$) \\
		$d_{mlp}$ & 64 & MLP hidden dimension \\
		\hline
		$V_t$ & 21 & Token vocabulary cardinality \\
		$V$ & 128 & VQ codebook size \\
		$l$ & 16 & words per sentence $s$ \\
		\hline
	\end{tabular}
	\caption{Network parameters for the language-enhanced transformer architecture. Parameters $k_b$ through $d_{mlp}$  are shared between the inference network (IN) and the auxiliary language network (ALN).}
	\label{tab:parameters_LEA}	
\end{table}

%-----------------------------------------------------------------------------------------

\begin{figure}[ht]
	\centering
	\includegraphics[width=0.9\textwidth]{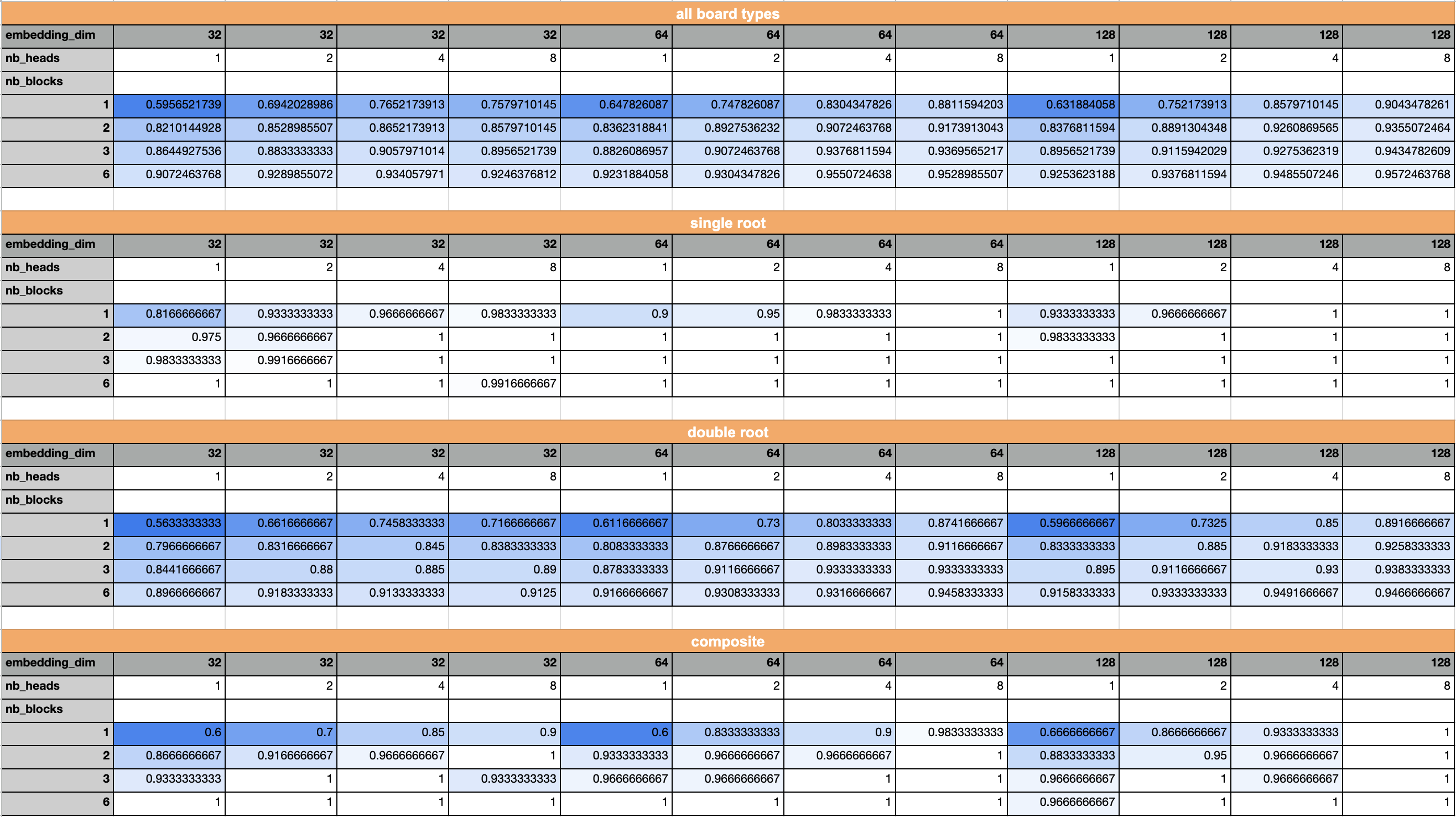}
	\caption{Network hyperparameter search: Median board accuracies for at least 3 runs (different network initialization and dataset instance). Accuracies are given for all board types (\textbf{top}), and individual types (single root object, double root object, and single composite object). We varied the embedding dimension (32, 64, 128), number of attention heads (1, 2, 4, and 8), and the number of transformer blocks.}
	\label{supfig:hyperparameter_search}
\end{figure}

\begin{figure}[ht]
	\centering
	\includegraphics[width=0.6\textwidth]{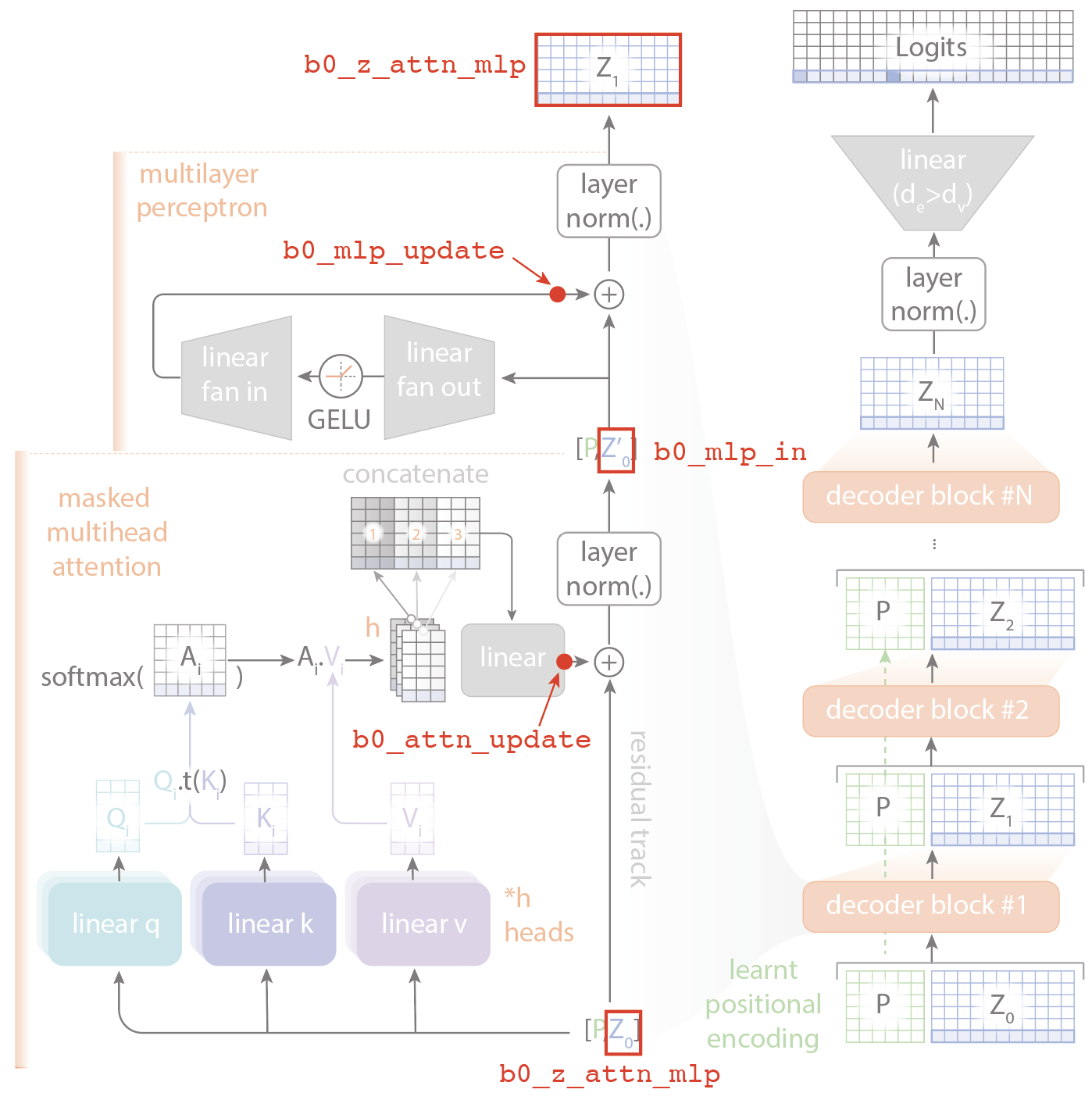}
	\caption{Delineation of computation stages studied in the paper. We focus on 4 stages per block: (i) attention update (\textnormal{\texttt{attn\_update}}), (ii) the input of the MLP subblock (\textnormal{\texttt{mlp\_in}}), (iii) MLP update (\textnormal{\texttt{mlp\_update}}), and (iv) output of the block (\textnormal{\texttt{z\_attn\_mlp}}). We prefix these stage names with \texttt{b0}, \texttt{b1}, or \texttt{b2} to indicate that they are part of transformer block 1, 2, and 3, respectively.}
	\label{supfig:computational_stages}
\end{figure}

\begin{figure}[ht]
	\centering
	\includegraphics[width=1.\textwidth]{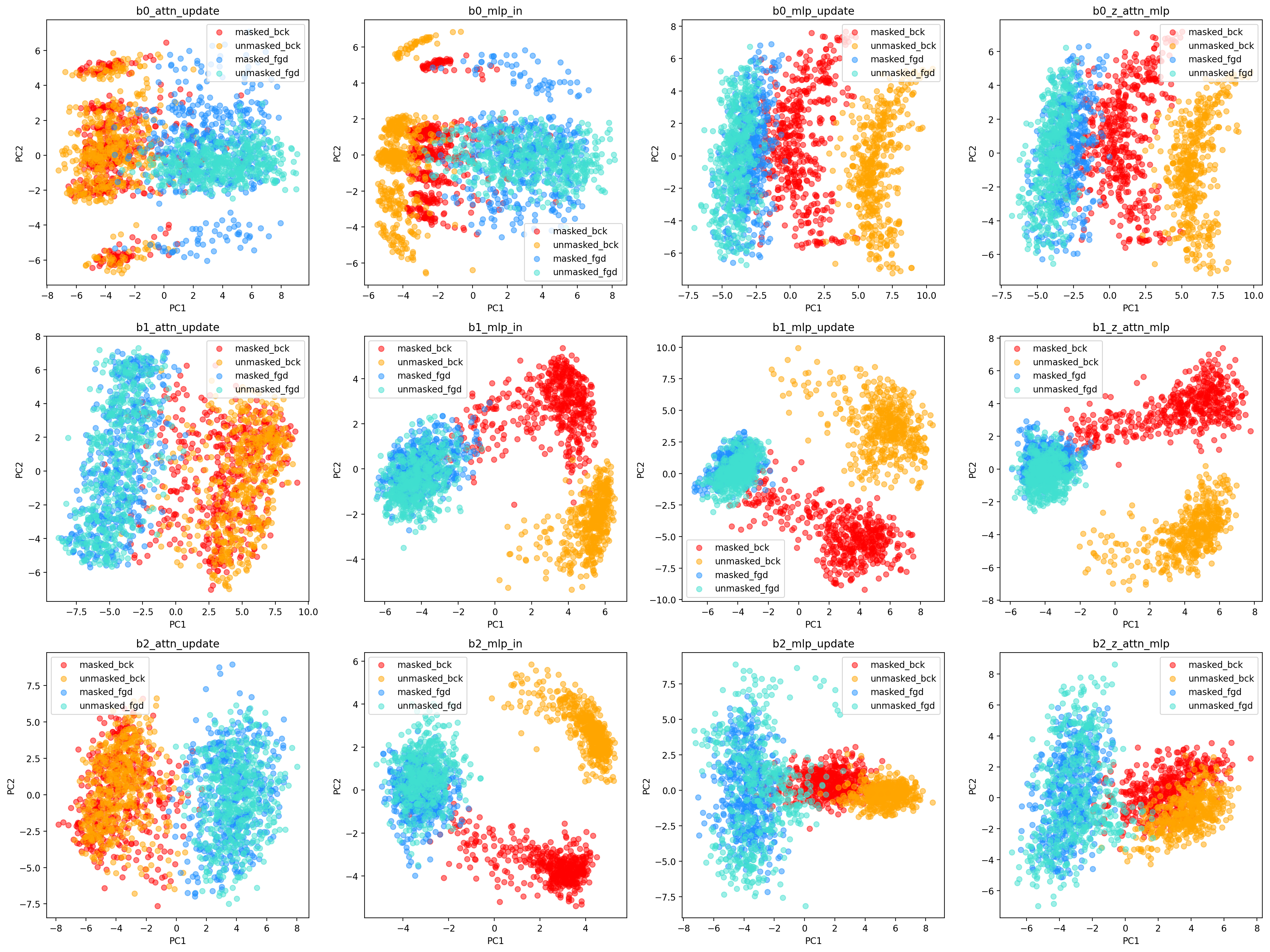}
	\caption{2d PCA projection of $\{z_i\}_i$ for randomly sample tokens at 12 computational stages throughout the network. Rows from top to bottom correspond to block 1 (\texttt{b0}) to 3 (\texttt{b2}), columns from left to right correspond to \textnormal{\texttt{attn\_update}}, \textnormal{\texttt{mlp\_in}}, \textnormal{\texttt{mlp\_update}}, and \textnormal{\texttt{z\_attn\_mlp}}). A total of 1000 random boards were generated (random background, root object, object position, masking). PCA on the sampled tokens was performed after feature normalization. Embeddings are color-coded to indicate their masking status (masked vs. unmasked) and type (background vs. object).}
	\label{fig:pca_bck_for_mask_unmask}
\end{figure}

\begin{figure}[ht]
	\centering
	\includegraphics[width=0.9\textwidth]{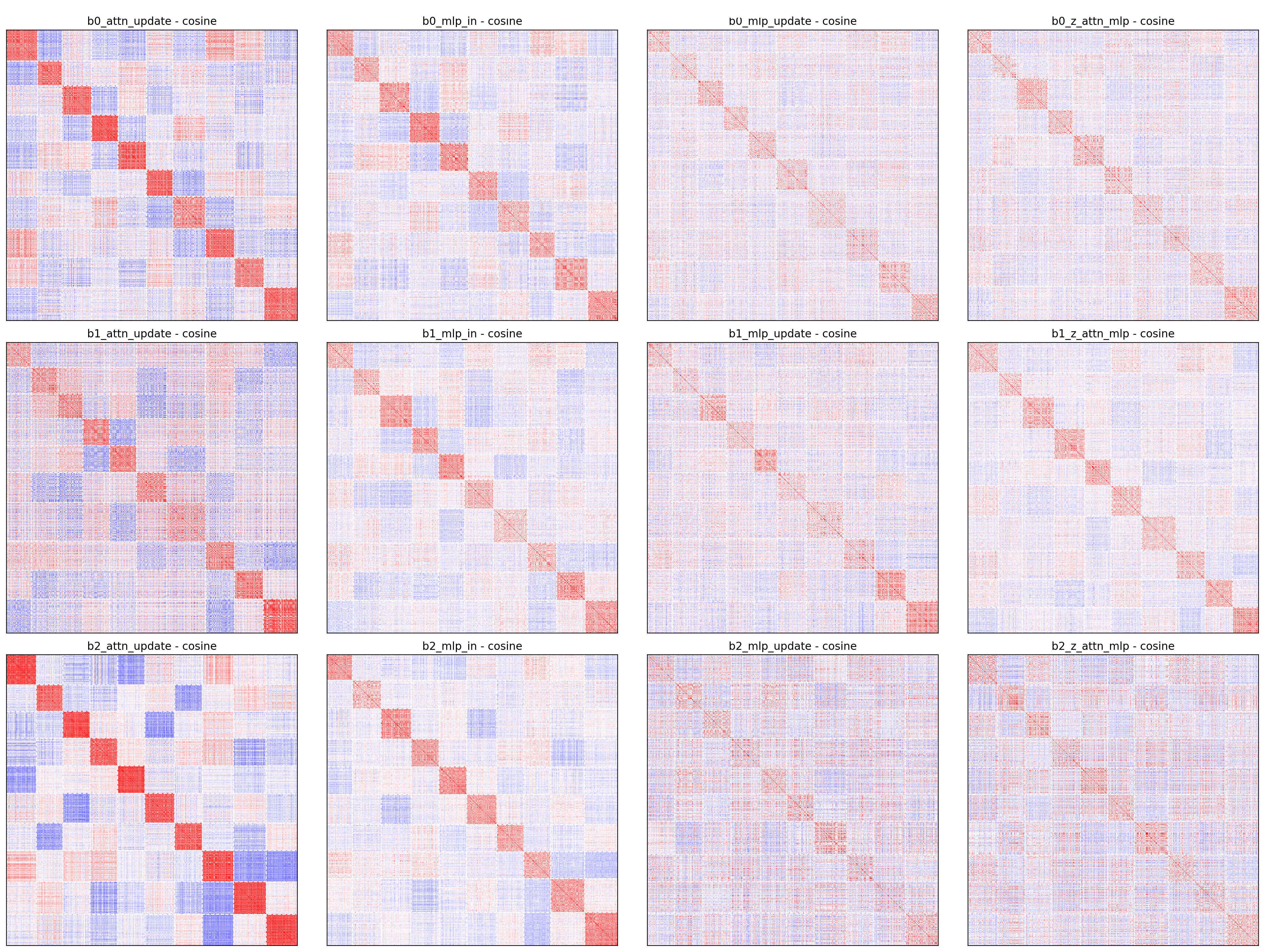}
	\caption{Pairwise cosine similarities between unmasked object tokens at various computational stages. Token representations are sorted by object membership to reveal any existing clustering based on this feature. The color scale ranges from -1 (blue) to +1 (red) and is the same across plots. Each row corresponds to a different transformer block. Each column corresponds to a different computational stage within the block.}
	\label{supfig:obj_abs_rsm_obj}
\end{figure}

\begin{figure}[ht]
	\centering
	\includegraphics[width=0.5\textwidth]{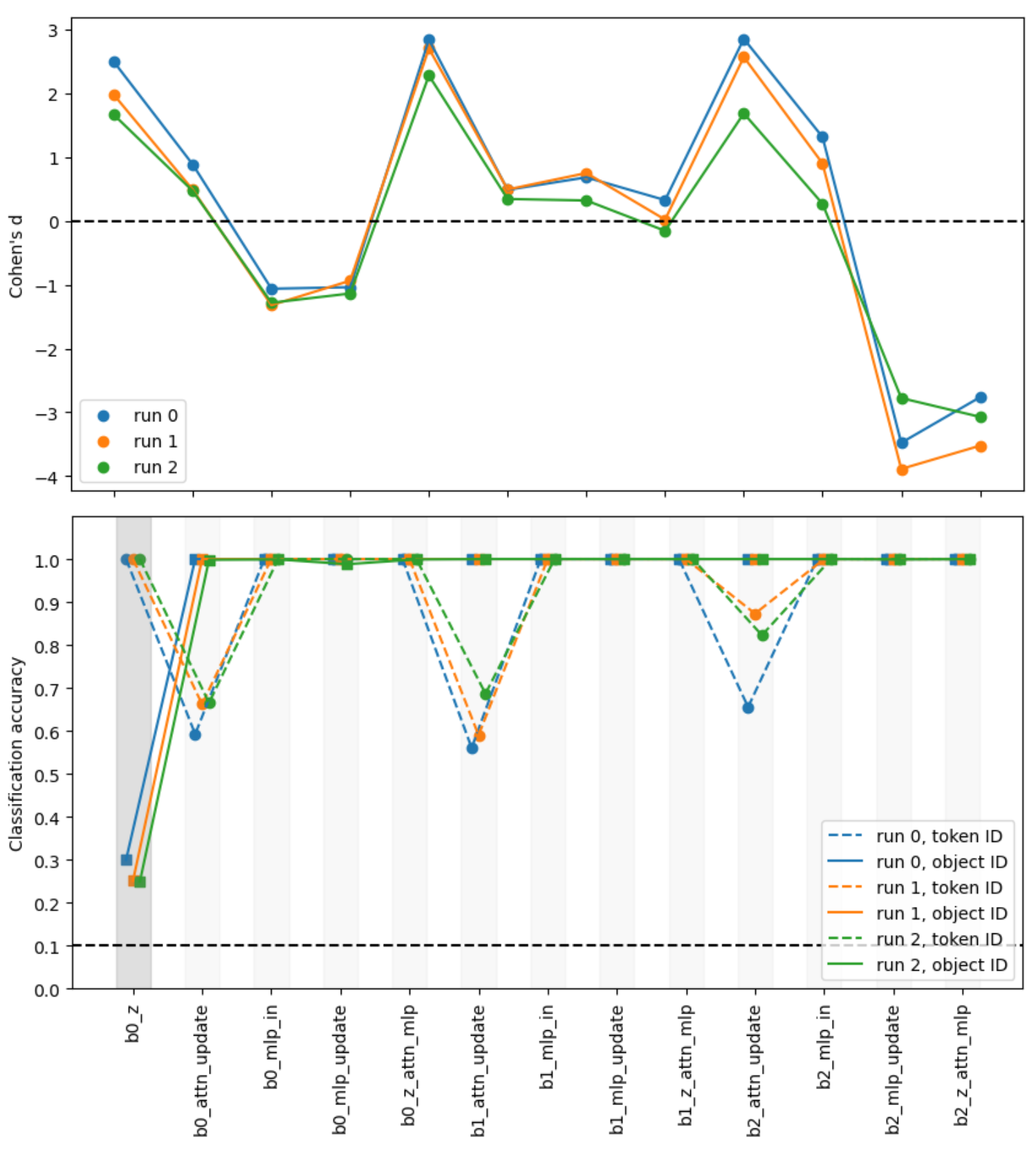}
	\caption{\textbf{Top}: Data aggregated across runs (different network initialization and dataset instance). Cohen’s d (difference between two means expressed in standard deviation units) across computational stages for groups “same object different token” versus “same token different object” (see Figure \ref{fig:obj_abs_violin_stringent} for details). \textbf{Bottom}: Accuracies of linear probes for token ID and object membership (object ID) across computational stages. Accuracies for three distinct runs are shown (blue, orange, green). Dashed lines give accuracies for the multi-class classification of token ID using a linear probe from unmasked token representations. Solid lines give accuracies for the multi-class classification of object membership. \textnormal{\texttt{b0\_z}} corresponds to initial token embeddings.}
	\label{supfig:obj_abs_cohenD_linearprobe}
\end{figure}

\begin{figure}[ht]
	\centering
	\includegraphics[width=0.5\textwidth]{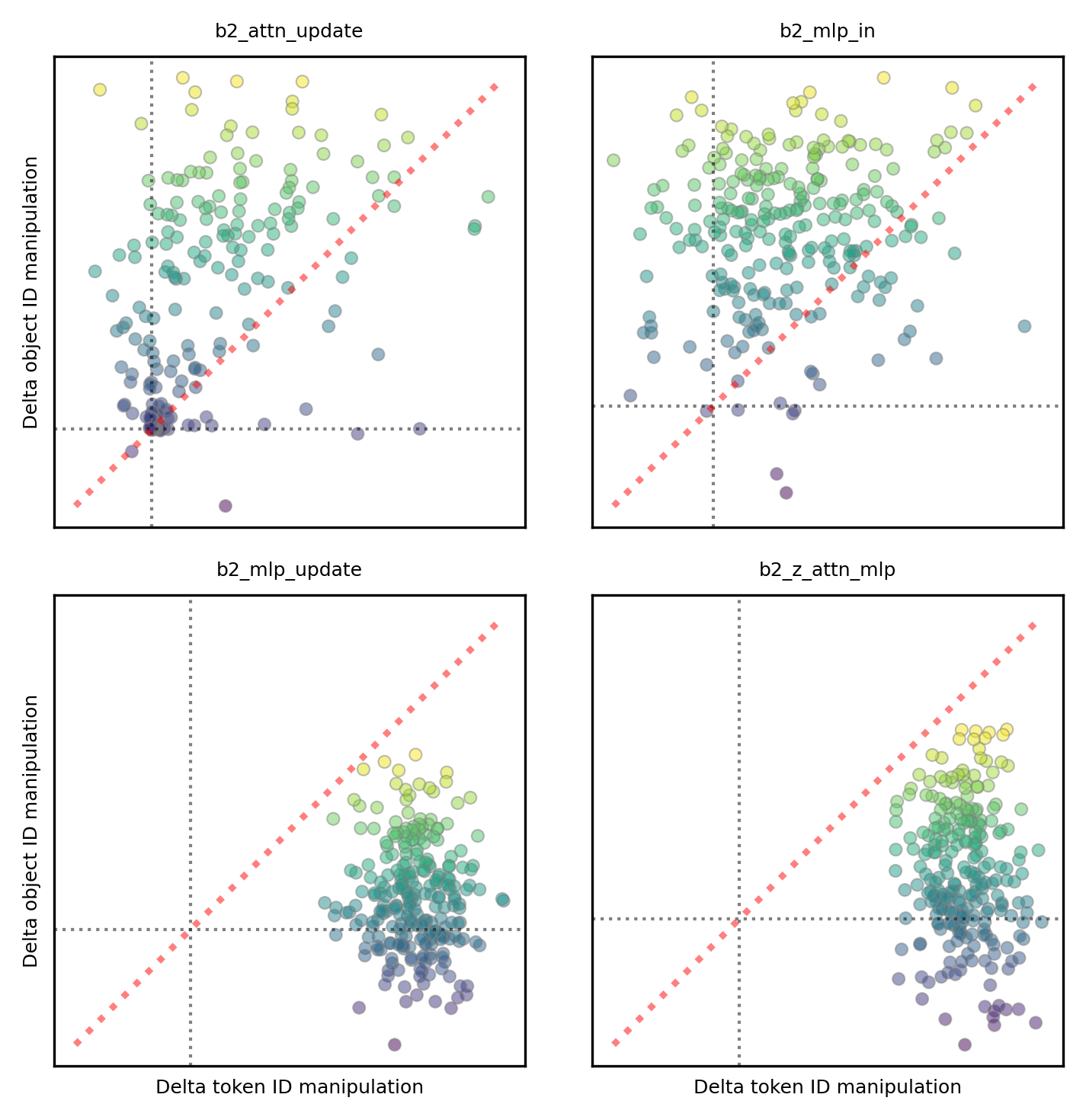}
	\caption{Head-to-head comparison of $\Delta$ for manipulations targeted at token ID (x-axis) vs. object membership (y-axis) at four critical computational stages. Each dot represents the manipulation of a given masked object token.}
	\label{supfig:many_manip_objvtokenID}
\end{figure}

\begin{figure}[ht]
	\centering
	\includegraphics[width=1.\textwidth]{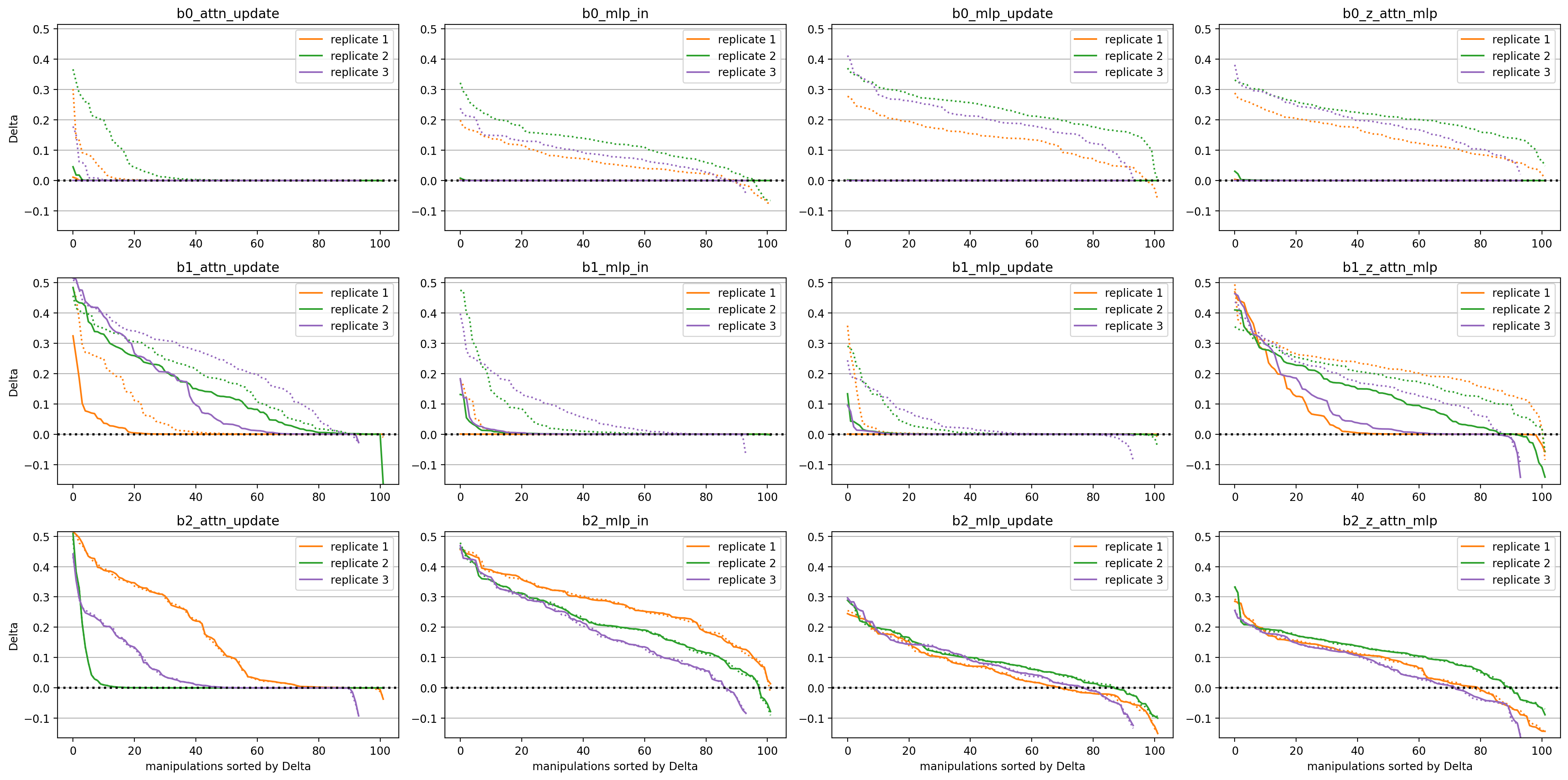}
	\caption{Effect of object membership abstraction manipulations across computational stages for three different runs (replicate 1-3). Solid and dotted lines show effects for single token edits and bulk object edits, respectively.}
	\label{supfig:many_manip_severalRuns}
\end{figure}

\begin{figure}[ht]
	\centering
	\includegraphics[width=0.4\textwidth]{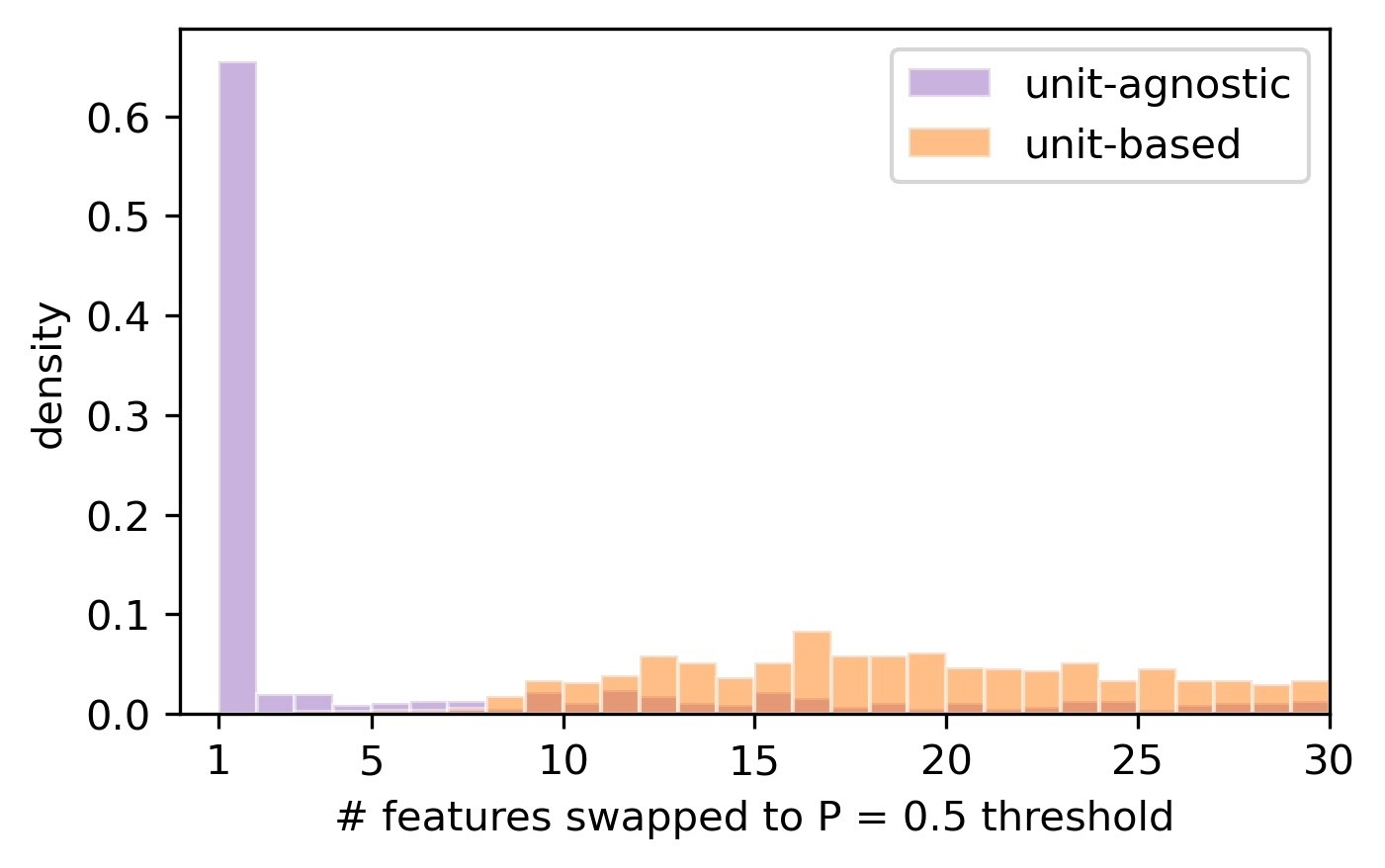}
	\caption{Number of features swapped to $P(t=t_B|z_A^*) \geq 0.5$ for unit-based and unit-agnostic methods. Only includes manipulations for which the 0.5 criterion was reached in under 30 feature edits for both approaches.}
	\label{supfig:unit_agnostic_bars}
\end{figure}

\begin{figure}[ht]
	\centering
	\includegraphics[width=1.\textwidth]{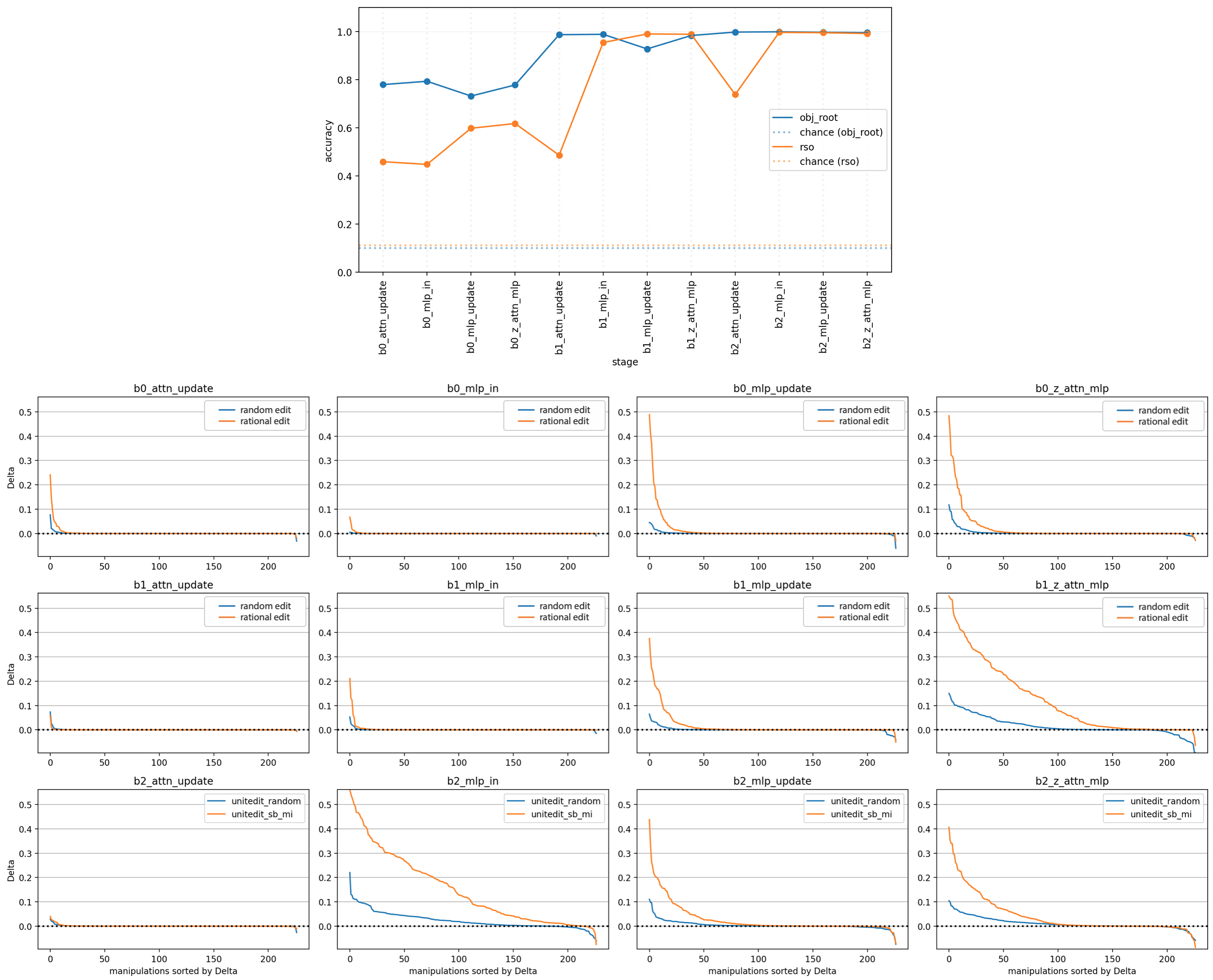}
	\caption{Delineation of causal relative spatial orientation (RSO) abstractions. \textbf{Top}: Multi-class linear classifiers for object membership (\textnormal{\texttt{obj\_root}}, blue) and RSO (\textnormal{\texttt{rso}}, orange) were trained and evaluated on embeddings from 12 distinct computational stages. The dotted lines correspond to chance level accuracies for each classification task. \textbf{Bottom}: Effect of RSO abstraction manipulations across cases and computational stages. Aggregated manipulations results for each computational stage. Solid orange and blue lines show the sorted $\Delta$ (y-axis) for different manipulations (x-axis) for rational edits (units ranked based on object information) and random edits (units ranked randomly), respectively. Replacement RSO abstractions were obtained from RSO prototype embeddings rather than target token representations.}
	\label{supfig:rep_ind_rso_probeAndManip}
\end{figure}

\begin{figure}[ht]
	\centering
	\includegraphics[width=0.7\textwidth]{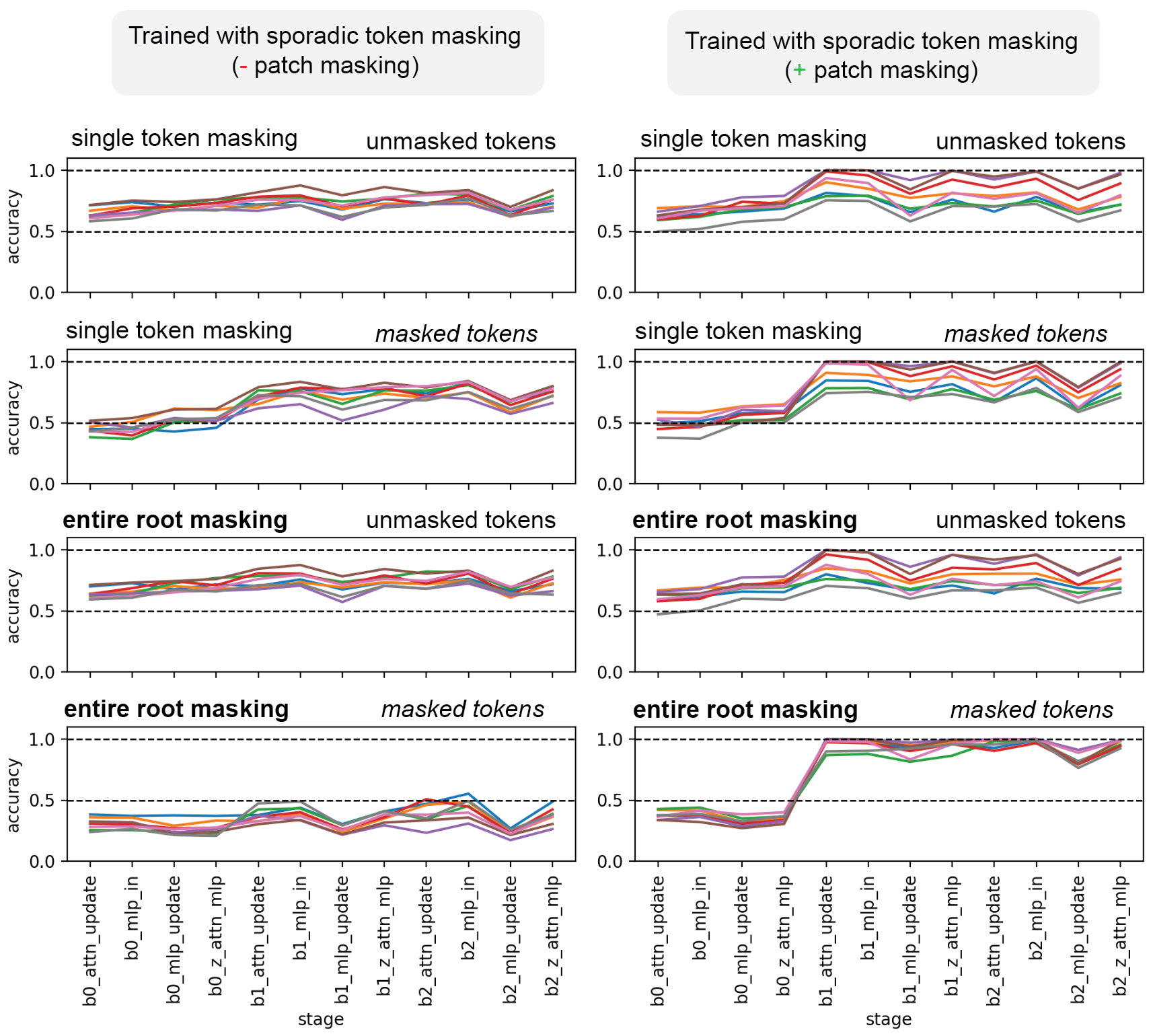}
	\caption{Effect of patch masking on the emergence of composite object abstractions. Comparison between networks trained with sporadic only (\textbf{left}) vs. sporadic + patch masking (\textbf{right}). After training, we subjected all networks to a collection of random boards, each featuring a single composite object on a random background. In half of the boards, a single token was masked (equivalent to sporadic masking), while in the other half, an entire constituent root object was masked (equivalent to patch masking). We extracted all object token embeddings at 12 computational stages, and divided them into four groups: (i) unmasked tokens part of sporadically masked boards (\textbf{row 1}); (ii) masked tokens from the same boards (\textbf{row 2}); (iii) unmasked tokens part of patch masked boards (\textbf{row 3}); (iv) masked tokens from the same boards (\textbf{row 4}). We then trained linear probes to classify composite object membership at each stage and for each group. We report the accuracy of each probe evaluated on a held-out evaluation set.}
	\label{supfig:comp_patch_sporadic_masking}
\end{figure}

\begin{figure}[ht]
	\centering
	\includegraphics[width=1.\textwidth]{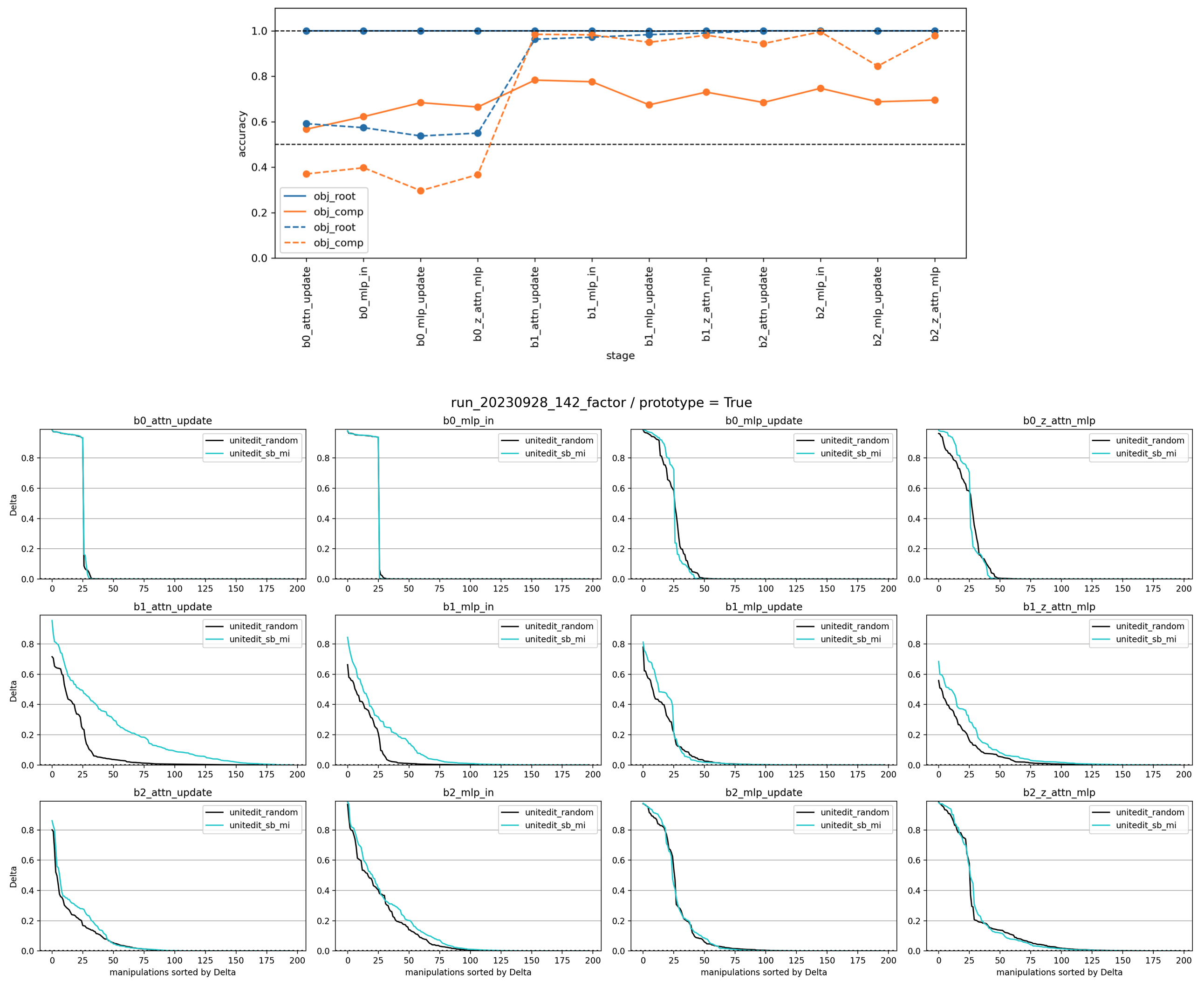}
	\caption{Delineation of causal composite abstractions: \textbf{Top}: Accuracies for linear probes against level 1 and 2 object abstraction across 12 computational stages. Probes were trained on unmasked (solid line) or masked (dashed line) token embeddings to classify level 1 object membership (\textnormal{\texttt{obj\_root}}, blue) or level 2 (\textnormal{\texttt{obj\_comp}}, orange). \textbf{Bottom}: Manipulation of level 2 composite abstractions in the embedding of patch-masked tokens at 12 different computational stages. Unit-based manipulations aimed at swapping composite abstraction with a prototype of the target abstraction. We performed bulk edits of all tokens in the patch-masked constituent root object. Solid blue and black lines show results for rational and random edits, respectively.}
	\label{supfig:comp_level2_delineation}
\end{figure}

\begin{figure}[ht]
	\centering
	\includegraphics[width=0.6\textwidth]{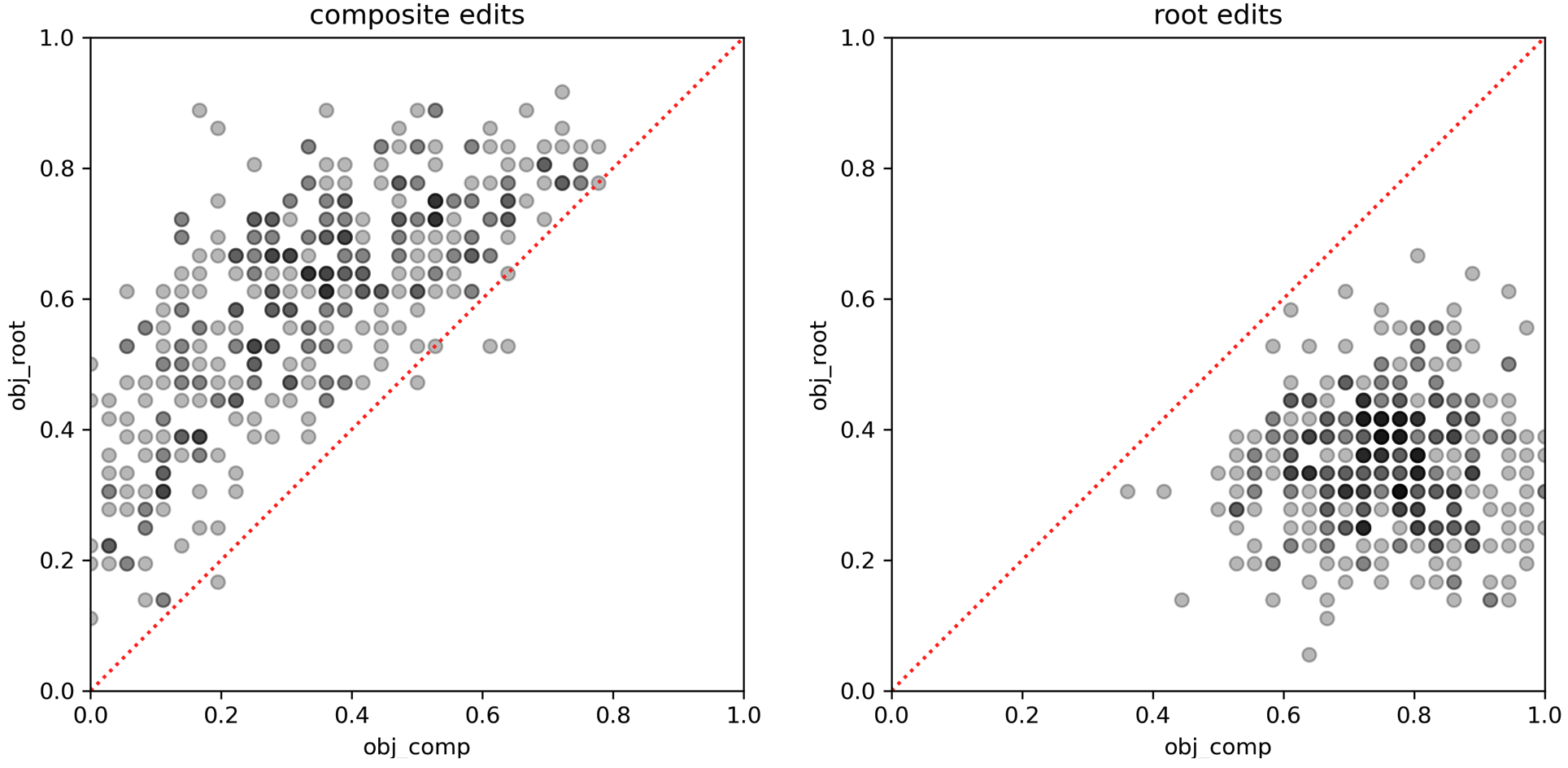}
	\caption{Efficacy and cross-talk of 1D edits against level 1 and level 2 object abstractions. Each point corresponds to the accuracies over the edited embedding across several boards (same composite, all possible single masking of the constituents). the x-axis shows the accuracy of the level 2 abstraction probe (\textnormal{\texttt{obj\_comp}}), and the y-axis the level 1 abstraction probe (\textnormal{\texttt{obj\_root}}). Edits were targeted at level 2 abstraction (\textbf{left}) or level 1 abstraction (\textbf{right}). Editing consisted of swapping the component along a single dimension of interest with the component of a randomly selected embedding that belongs to a different class.}
	\label{supfig:comp_decoding_edits}
\end{figure}

\begin{figure}[ht]
	\centering
	\includegraphics[width=0.7\textwidth]{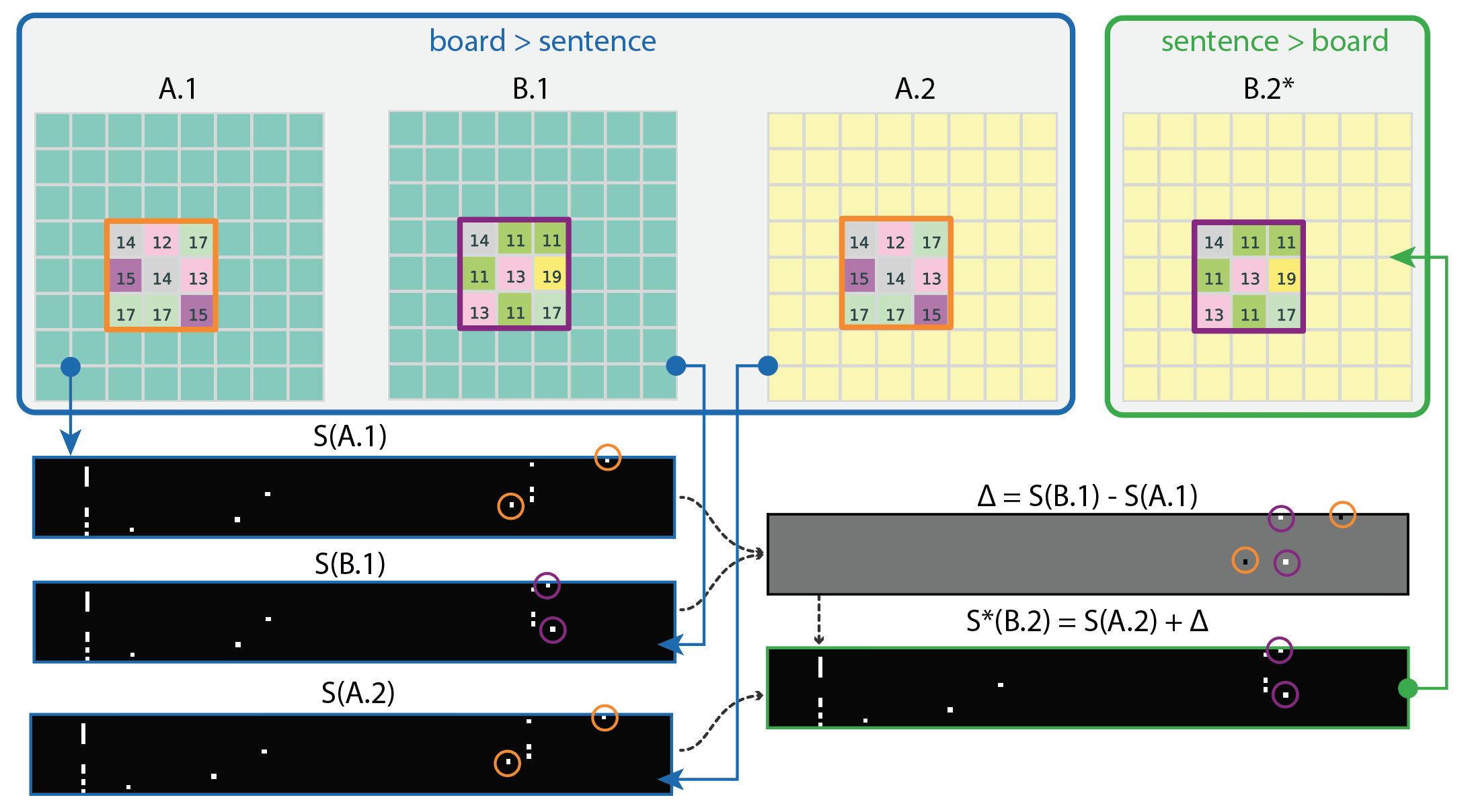}
	\caption{Example of contextual independence, swapping objects. See Figure \ref{fig:comp_context_independence} for details.}
	\label{supfig:comp_context_independence_obj}
\end{figure}

\end{document}